\documentclass{article}
\usepackage{graphicx} 
\usepackage{authblk}
\usepackage{multirow}
\usepackage{amssymb}
\usepackage[table]{xcolor}\usepackage{tikz}
\usetikzlibrary{arrows.meta, positioning}
\usepackage{pifont} 
\newcommand{\xmark}{\ding{55}} 
\usepackage{amsmath} 
\usepackage{makecell}
\usepackage{multirow}
\usepackage[colorlinks=true,
            linkcolor=blue,
            citecolor=blue,
            urlcolor=blue]{hyperref}        
\usepackage{longtable}  
\usepackage[font=small]{caption}

\usepackage{array} 
\usepackage{float} 
\usepackage{caption}
\usepackage{xcolor}
\usepackage{comment}
\usepackage{hyperref}
\usepackage[numbers,sort&compress]{natbib}
\usepackage{url}
\usepackage[a4paper, total={7in, 9in}]{geometry}
\usepackage{lipsum} 
\newcommand{\keywords}[1]{\textbf{Keywords:} #1}

\title{A Data-Centric Review of Plant Disease Datasets: Taxonomy, Critical Analysis, Environmental Variability, and Implications for Precision Agriculture}

\author{
\large
\textit{Aamir Hilal}$^{1*}$ \quad
\textit{Shabir Ahmad Sofi}$^{1}$ \quad
\textit{Neeraj Goel}$^{2}$ \\
\vspace{-10pt}
{\footnotesize
\textcolor{blue}{aamir\_hilal@nitsri.ac.in} \quad
\textcolor{blue}{shabir@nitsri.ac.in} \quad
\textcolor{blue}{neeraj@iitrpr.ac.in}
}
}
\vspace{-10pt}
\date{
{\footnotesize
$^{1}$Department of Information Technology, National Institute of Technology, Hazratbal Srinagar 190006, Jammu \& Kashmir, India \\
$^{2}$Department of Computer Science and Engineering, Indian Institute of Technology, Ropar Rupnagar 140001, Punjab, India
$^{*}$\textbf{Corresponding author:} \textcolor{blue}{aamir\_hilal@nitsri.ac.in}
}
}
\vspace{-10pt}
\begin{document}
\maketitle
\begin{abstract}
Despite rapid advances in artificial intelligence, achieving reliable real-world plant disease detection remains a persistent challenge. In recent years, visual and deep learning–based approaches have shown promising results for automated plant disease detection, however, their real-world deployment remains limited. A key bottleneck is the heavy reliance on laboratory-generated datasets that lack environmental diversity, realistic backgrounds, and balanced class distributions, leading to poor generalization under field conditions. In contrast, datasets collected directly from real agricultural environments capture natural variability and more accurately reflect the challenges faced by farmers across diverse geographical regions. This review presents a comprehensive and critical analysis of existing visual and deep learning approaches for plant disease detection, with a primary focus on the characteristics of plant disease datasets. It establishes a structured taxonomy of plant disease datasets by acquisition setting, accessibility, plant diversity, disease composition, class structure, and imbalance severity, accompanied by a critical analysis of their implications for model generalization and real-world deployment. A comparative examination of laboratory and real-field datasets reveals critical gaps that hinder the development of reliable and deployable disease detection systems. Furthermore, this review systematically analyzes how multi-level dataset imbalance (intra-class, inter-crop, cross-dataset) and limited environmental variability influence model performance and robustness, an area that remains insufficiently examined in existing surveys. Beyond image-based approaches, this review highlights the growing importance of integrating environmental parameters such as temperature, humidity, and leaf wetness with image data to improve predictive performance under dynamic field conditions. The review further outlines major challenges, unresolved research issues, and future directions concerning dataset construction, environmental variability, structural imbalance, standardization, and multimodal disease monitoring frameworks. This paper presents a review which will be beneficial for development of next-generation multimodal frameworks capable of supporting real-world precision agriculture.

\end{abstract}

\section{Introduction}
\label{sec:Introduction}
Plant diseases continue to pose a serious threat to global food security, leading to substantial reductions in crop yield and significant economic losses worldwide. The Food and Agriculture Organization (FAO) reports that up to 40\% of crops are lost annually due to plant diseases \cite{fao_plant_production_protection}, emphasizing the urgent need for timely and accurate detection strategies. As the global population grows and the demand for food, feed, and fiber increases, ensuring sustainable agricultural productivity has become more critical than ever~\cite{un2022worldpopulation,un2022wpp}. In this context, precision agriculture has emerged as a transformative approach that leverages advanced technologies to enhance crop monitoring, optimize resource utilization, and support data-driven farm management. However, the effectiveness of such intelligent agricultural systems depends heavily on the ability to detect plant diseases accurately and at early stages. This objective is increasingly challenged by climate change, environmental degradation, rapid urbanization, and shrinking agricultural land, all of which place additional strain on modern farming systems~\cite{un_population_2021,unfpa2024}.
\par
In recent years, advances in artificial intelligence (AI), particularly in computer vision and deep learning, have revolutionised sustainable agriculture, assisting farmers in making informed decisions and reducing potential crop losses \cite{chouhan2020applications,gajjar2022real,murecsan2020detecting,yao2024deep}. Most importantly, the identification of plant diseases through leaf imagery has undergone extensive research and is widely applied in practice, showcasing their substantial influence in the field \cite{ashok2020tomato,azlah2019review,li2021plant}. This emphasis stems from the fact that many diseases manifest visibly on leaves, making them a reliable indicator for early diagnosis~\cite{kanda2021deep,lee2021conditional,othman2022plant}. Deep learning models especially convolutional neural networks (CNNs) have shown remarkable capability in extracting discriminative features from leaf images, frequently outperforming traditional visual inspection methods \cite{yao2023machine}. Despite these promising developments, the successful deployment of such systems in real-world agricultural environments remains limited.
\par
One of the primary reasons for this gap is the nature of the datasets used to train and evaluate these models. The effectiveness of deep learning models is closely tied to the quality, scale, and diversity of the training data. This relationship is well documented in both academic research and practical applications, which emphasize the critical role that datasets play in the detection of plant disease. Numerous studies highlight a significant challenge in this field: the scarcity of readily available, high-quality datasets \cite{chouhan2020applications,li2021plant}. Addressing this scarcity is crucial for advancing research and improving outcomes in plant disease diagnosis. There are many datasets available on the different leaf diseases over a wide variety of plants and crops, but still, there is a lack of large-scale and more diverse, publicly available datasets, which is cited as one of the primary reasons for resolving plant disease diagnosis problems \cite{kumar2022retracted,agarwal2021plant}. Although numerous plant disease datasets have been introduced over the past decade, a substantial portion has been generated under controlled laboratory conditions characterized by uniform backgrounds, stable illumination, and minimal environmental variability \cite{seetharaman2022leaf,syed2022citrus}. These datasets often fail to capture the variability and complexity of real-world agricultural environments where heterogeneous backgrounds, occlusions, fluctuating lighting conditions, and diverse disease expressions are common. This limitation can lead to models that excel in the controlled lab setup but encounter difficulties when exposed to field conditions. For instance, the widely used publicly available PlantVillage dataset has reported high classification accuracy; however, the controlled acquisition settings do not adequately reflect the unpredictable nature of real agricultural environments \cite{mohanty2016using,fenu2022evaluating,hughes2015open}.

\par Although numerous review studies have explored deep learning techniques for plant disease detection, most have primarily focused on model architectures and performance improvements, often giving less attention to the datasets that underpin these methods. Important dataset attributes including acquisition settings, availability (public versus proprietary), plant diversity, disease composition, class structure, and imbalance severity have not been comprehensively synthesized. Moreover, the extent to which these characteristics influence model robustness and real-world applicability remains insufficiently examined. This lack of a dataset-centered perspective represents a notable gap in the current literature. Thus, there is a lack of sufficient large-scale non-lab datasets, which remains a major challenge for enabling deep learning-based plant disease detection. Therefore, there is a need to focus on creating datasets that include images taken in real-world environments to better reflect the varied and dynamic conditions of natural settings \citep{liu2021plant,xu2024plant}.
\par
Beyond dataset considerations, environmental conditions play a crucial role in the emergence and progression of plant diseases. Factors such as temperature, soil moisture, humidity, wind, leaf wetness, etc, contribute significantly to the development and severity of plant diseases \cite{stevens1960cultural}. These Factors notably impact both the growth of pathogens and the susceptibility of plants to various diseases~\cite{colhoun1973effects}. Climatic fluctuations may also shift agroclimatic zones, potentially exposing crops to new pathogens and increasing disease severity \cite{hunjan2020climate}. Despite these influences, most existing detection frameworks rely predominantly on visual data, often overlooking the contextual insights offered by environmental parameters. Integrating environmental information with leaf image data represents is an emerging research direction with strong potential to enhance predictive reliability, particularly under dynamic field conditions. Such multimodal approaches may support the development of next-generation disease prediction systems capable of operating effectively in complex real-world scenarios.

\par In view of these challenges, the present review takes a dataset-centric approach to plant disease identification research. It establishes a systematic taxonomy of plant disease datasets according to the setting in which data are acquired (laboratory vs. field), accessibility, plant diversity (single vs. multi-species), disease and class composition, and class imbalance characteristics. The review then discusses how each of these attributes influences aspects of model performance, robustness, and readiness for deployment, followed by a critical review of laboratory versus real-field datasets to identify critical gaps in current data resources. Furthermore, it outlines important considerations for future dataset development in light of the growing recognition of the need to combine environmental data with visual information. This review will assist in advancing reliable, scalable, and real-world-ready plant disease detection systems for precision agriculture. 

\subsection{Motivation}
\label{sec:Motivation}
Despite notable progress in deep learning–based plant disease detection, there are only limited transitions from experimental success to dependable real-world deployment. Increasingly, this challenge is being attributed not solely to model design but to the characteristics of the datasets used for training and evaluation. As research moves toward operational agricultural intelligence systems, the need for datasets that accurately represent field conditions has become more critical than ever.

\par Many existing plant disease datasets are constrained by factors such as controlled acquisition environments, limited accessibility, narrow crop coverage, inconsistent annotation practices, and skewed class distributions. In particular, class imbalance and the scarcity of large-scale real-field datasets continue to hinder the development of models capable of generalizing beyond experimental settings. Without a clearer understanding of these dataset properties, further improvements in model architectures alone may offer limited practical benefit.

\par At the same time, plant disease development is closely influenced by environmental conditions, including temperature, humidity, and leaf wetness. However, most current detection frameworks rely predominantly on visual data, often overlooking the contextual insights that environmental parameters can provide. Incorporating such information alongside image data presents a promising opportunity to improve predictive reliability, especially under dynamic field conditions.

\par Motivated by these observations, this review adopts a dataset-centric perspective to systematically examine the structural characteristics of plant disease datasets, assess their implications for model robustness and deployment readiness, and identify critical gaps that must be addressed to support real-world precision agriculture.

\subsection{Contributions}
\label{sec:Contributions}
This review provides a comprehensive, dataset focused synthesis of plant disease detection research, emphasizing the central role of data in enabling robust and deployable intelligent agricultural systems. The primary contributions of this study are summarized as follows:
\begin{itemize}
    \item \textbf{Structured taxonomy of plant disease datasets:} \\
    The review organizes existing datasets according to key attributes such as acquisition setting (laboratory versus field), accessibility (public versus proprietary), plant diversity (single versus multi-species), disease and class composition, and the presence of class imbalance.
    
    \item \textbf{Dataset-centric perspective on model performance:} \\ 
    It examines how fundamental dataset characteristics influence model generalization, robustness, and readiness for real-world deployment, highlighting dataset quality as a critical factor in practical disease detection.
    
    \item \textbf{Comparative analysis of laboratory and real-field datasets:}\\  
    A detailed comparison is presented to clarify the strengths and limitations of controlled and field-acquired data, along with their implications for agricultural applications.
    
    \item \textbf{Focused discussion on class imbalance:}\\  
    The review explores the prevalence of imbalance (intra-class, inter-crop, cross-datase in plant disease datasets and discusses its effects on model bias, predictive stability, and evaluation outcomes—an area that remains insufficiently synthesized in prior literature.
    
    \item \textbf{Consideration of environmental variability:}  \\
    It further examines the role of environmental factors in disease progression and discusses the potential benefits of integrating environmental data with visual inputs to enhance predictive performance.
    
    \item \textbf{Guidance for future dataset development:}  \\
    Finally, the review outlines key considerations for constructing next-generation datasets that better capture real-world variability and support the development of scalable, field-deployable plant disease detection systems for precision agriculture.
\end{itemize}

\paragraph{Organization of the Paper}
The remainder of this paper is organized as follows. Section \ref{sec:Background} provides background on deep learning paradigms and architectural strategies employed in plant disease detection. Section \ref{sec:Existing Surveys on Plant Disease Detection} reviews existing survey studies and identifies gaps in dataset-centric analysis. Section \ref{sec:Systematic Literature Search and Selection Strategy} outlines the systematic literature search and selection methodology adopted in this review. Section \ref{sec:Taxonomy of Plant Disease Datasets} presents a structured taxonomy of plant disease datasets, categorizing them based on scope, acquisition setting, accessibility, and class composition. Section \ref{sec:Critical Analysis of Plant Disease Datasets} critically analyzes the structural characteristics of these datasets, including quantitative imbalance assessment using the normalized imbalance ratio (NIR). Section \ref{sec:Class Imbalance and Distribution Bias} discusses class imbalance and distribution bias and examines their implications for model learning behavior. Section \ref{sec:Deep Learning Models Across Plant Disease Datasets} reviews deep learning models across plant disease datasets and analyzes the interaction between model complexity and dataset characteristics. Section \ref{sec:Environmental Variability and Multimodal Consideration} explores environmental variability and the integration of multimodal data for context-aware disease prediction. Section \ref{sec:Challenges and Open issues} summarizes open challenges and emerging research directions. Section \ref{sec:Discussion} provides a broader discussion of the findings, and Section \ref{sec:Conclusion} concludes the paper with key insights and future directions toward field-ready precision agriculture systems.

\section{Background}
\label{sec:Background}
Computer vision and deep learning are advanced technologies that play a crucial role in the early and precise detection and diagnosis of plant diseases. These technologies analyze visual signs to detect and diagnose plant diseases. By learning hierarchical feature representations from plant imagery, these models can autonomously undertake various tasks crucial for disease detection \cite{sunil2023systematic,lee2023deep}. These tasks include identifying various objects within images, recognizing specific patterns indicative of diseases, categorizing plant diseases based on visual cues, and monitoring crop conditions \cite{barbedo2016review,patricio2018computer,mohanty2016using}. These models are highly effective for detecting plant diseases because they can learn patterns in images and can identify the specific disease affecting the crop, which aids in making informed decisions about treatment and control measures~\cite{samajpati2016hybrid,rozario2016segmentation}. Deep learning models especially, convolutional neural networks (CNNs) are highly effective deep learning models revolutionizing plant identification and disease detection \cite{kumar2020leaf,bharate2020classification}. Their capability consistently outperforms traditional machine learning methods, solidifying their status as the ideal solution for these critical tasks \cite{dang2021citrus,tulshan2019plant,rumy2021iot,rajesh2020leaf}.

\par It's important to highlight that while deep learning models are formidable in classifying and detecting diseases, they thrive on large datasets of images and relevant metadata for training. This dependency on extensive data is a strength that allows these models to navigate complexity and achieve broad applicability in real-world scenarios. Factors such as dataset scale, annotation quality, class distribution, and environmental variability often influence model selection, training strategies, and generalization capability. Therefore, understanding the landscape of deep learning models provides essential context for examining how dataset properties shape research outcomes in plant disease detection.
\par
\begin{figure}[h]
    \centering
    \includegraphics[width=0.7\textwidth]{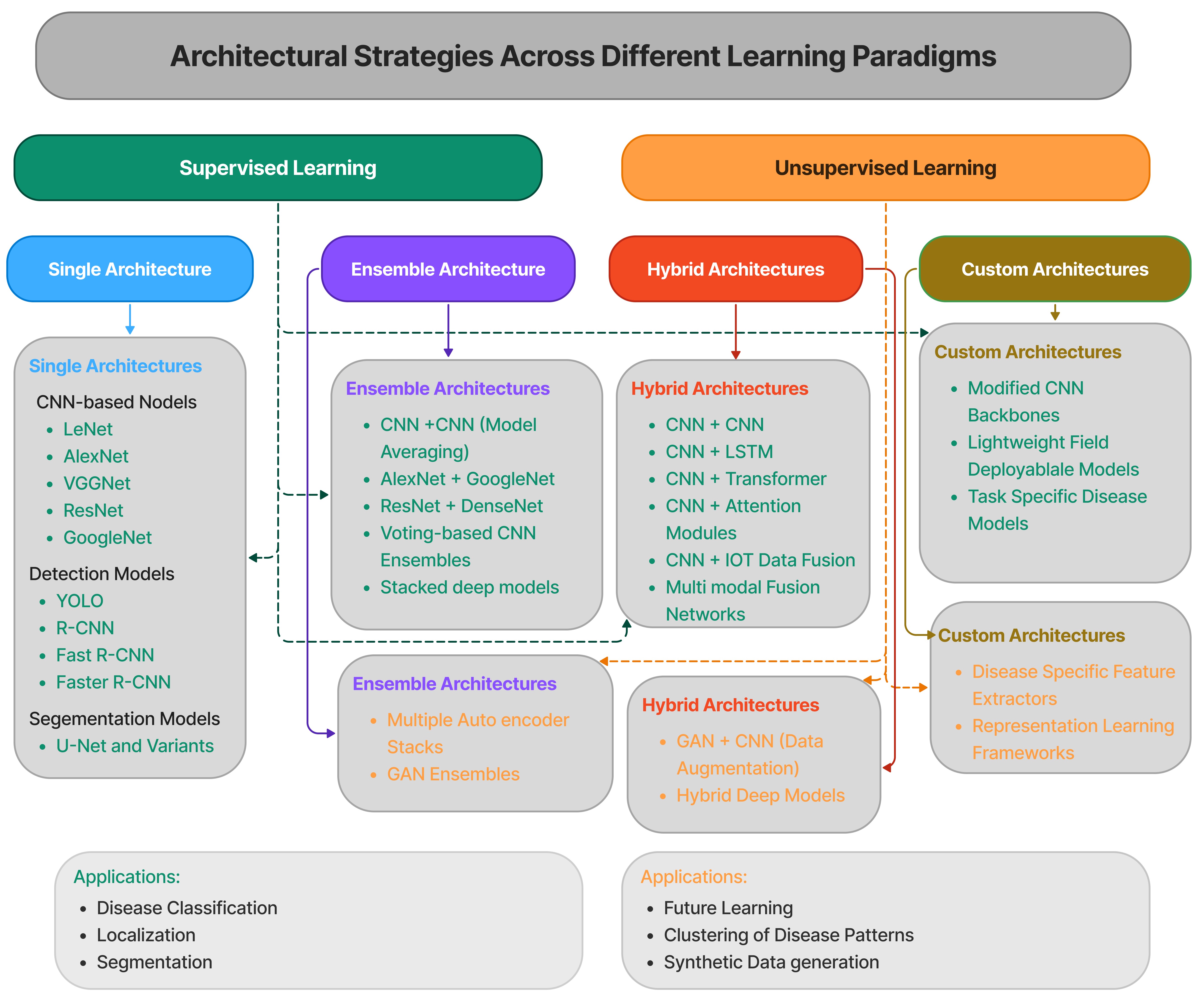}
    \vspace*{-2mm}
    \caption{Taxonomy of deep learning models for plant disease detection structured along two dimensions: learning paradigm and architectural strategy.}
    \label{fig: Taxonomy of deep learning models.}
\end{figure}

In plant disease detection research, methodological diversity arises from two orthogonal dimensions: learning paradigm and architectural strategy. While learning paradigms define how supervision signals are utilized, architectural strategies determine how spatial, temporal, and multimodal features are extracted and fused. Figure \ref{fig: Taxonomy of deep learning models.} presents a streamlined taxonomy of deep learning models for plant disease detection structured hierarchically according to learning paradigm and architectural strategy.

\subsection{Learning Paradigms in Plant Disease Detection}
\label{sec:Learning Paradigms in Plant Disease Detection} 
Existing deep learning approaches for plant disease detection can be broadly categorized based on their learning paradigms, including supervised, unsupervised, and hybrid strategies. Among these, supervised learning remains the most widely adopted due to the availability of annotated disease datasets. However, the growing scarcity of labeled data has encouraged exploration of unsupervised and hybrid approaches capable of leveraging unlabeled images. Supervised learning dominates current research due to the availability of annotated datasets, whereas unsupervised and hybrid strategies are increasingly explored to mitigate labeling constraints.

\subsection{Architectural Strategies for Plant Disease Detection}
Deep learning architectures employed in plant disease detection can be broadly categorized based on their structural design and modeling strategies. While numerous variants have been proposed in the literature, most approaches can be grouped into four primary categories: single, ensemble, hybrid, and custom convolutional neural network (CNN) models. This categorization reflects the evolving efforts of researchers to balance predictive performance, computational complexity, and adaptability to diverse dataset characteristics.

\subsubsection{Single Deep Learning Architectures}
Single CNN models consist of a single architecture trained on a specific dataset for plant disease detection. Single CNN models are used to extract features from input images and make predictions about the presence or absence of diseases. This approach offers simplicity and efficiency, especially when working with well-defined datasets and straightforward classification tasks. Researchers often fine-tune the architecture's parameters to achieve optimal performance for specific tasks. A prime example of this is the use of the VGG convolutional neural network (CNN) model, which is specifically designed to analyze rice leaf images. By extracting complex features from these images, the model's ability to identify a wide range of plant diseases is enhanced significantly \cite{jiang2021recognition}. 
\par Single deep learning architectures have been extensively utilized for plant disease classification due to their structural simplicity and strong feature extraction capabilities. 
Despite their effectiveness, the performance of single architectures is often contingent upon the availability of sufficiently large and well-balanced datasets. In data-constrained scenarios, researchers frequently rely on transfer learning and augmentation techniques to enhance generalization.

\subsubsection{Ensemble Architectures}
Ensemble CNN models combine predictions from multiple individual CNN models to make a final decision. An ensemble of several CNN models can significantly improve the accuracy of predictions compared to the accuracy of any single model included in the ensemble. For example, an ensemble comprising DenseNet161, ResNet152, and VGG19 can be applied to image classification. In this approach, combining predictions from different CNNs improves overall performance \cite{turkoglu2022plantdiseasenet}.

\par 
Ensemble approaches integrate predictions from multiple deep learning models to improve classification reliability and reduce variance. By aggregating complementary feature representations, these models often achieve higher predictive performance than individual architectures.
However, ensemble models typically require substantial training data to avoid overfitting and to fully leverage architectural diversity, further emphasizing the importance of dataset scale and variability.

\subsubsection{Hybrid Architectures}
Hybrid CNN models combine the strengths of different types of models or different approaches. These models often leverage pre-trained CNNs as feature extractors and combine them with additional layers or modules tailored to the requirements of plant disease detection tasks. Hybrid models aim to capitalize on the strengths of different architectures while addressing specific challenges or constraints present in plant disease detection scenarios. For example, a Hybrid CNN combination of CNNs and GAN can detect apple diseases from leaf images \cite{chen2023apple}.

\par
Hybrid architectures combine deep learning models with complementary techniques to address specific challenges such as limited data availability, feature enhancement, or synthetic data generation.
These approaches are frequently motivated by dataset limitations, particularly when real-world training samples are scarce.

\subsubsection{Custom Architectures}
Custom CNN models are built from scratch to address specific tasks, utilizing custom curated datasets. These models are particularly beneficial in scenarios where pre-trained models do not provide adequate performance or when the specific features needed for effective analysis are not captured by existing architectures. Additionally, building a custom CNN allows for targeted feature extraction, enabling the model to learn characteristics that existing models may overlook. For instance, a custom CNN model is made with two convolutional, two pooling, and three fully connected layers with an output layer for disease classification \cite{rakib2024lightweight}.
\par Custom-designed architectures are often developed to address task-specific requirements or to adapt to the constraints of specialized datasets.
Their design frequently reflects the characteristics of the datasets for which they are intended, highlighting the reciprocal relationship between model development and data availability.

\subsection{Data Dependency of Deep Learning Models}
Regardless of architectural sophistication, these deep learning models thrive on large, diverse, and well-annotated datasets to learn intricate patterns and features, enabling them to make precise predictions. However, the datasets available for training deep learning often fall short in terms of the number of images, prompting researchers to employ data augmentation, transfer learning, and synthetic data generation techniques to compensate for dataset deficiencies \cite{elfatimi2022beans,seetharaman2022leaf,saleem2021mango,jiang2021recognition}. This involves artificially expanding the dataset by applying transformations, thereby enhancing the model's ability to generalize \cite{jin2022grapegan}. While these strategies can improve performance, they do not fully substitute for realistic field-acquired datasets. Additionally, many available datasets are collected in labs under controlled conditions, often not capturing real-world variability and unpredictability \cite{seetharaman2022leaf}. There are very few real-world field datasets that offer the diverse and unpredictable variables that models need to manage in real-world applications, such as varying lighting, weather conditions, and different backgrounds \cite{metre2022reviewing,kumar2022retracted,syed2022citrus}. This lack of real field datasets poses a significant challenge for developing robust models that perform well outside the lab \cite{kumar2022retracted,thakur2022trends,syed2022citrus}. The few real field datasets that do exist are often not publicly available, limiting their accessibility to researchers and hindering the advancement of field-ready models. This scarcity of real-world data underscores the need for more open and accessible real-field datasets to improve model generalization and applicability, ultimately leading to more reliable and effective deep learning models in real-world scenarios.
\par The diversity of deep learning architectures reflects the rapid advancement of automated plant disease detection research. Nevertheless, model success remains intrinsically linked to dataset quality, diversity, and environmental realism. This underscores the need for a systematic examination of plant disease datasets.

\section{Dataset-Centric Review of Existing Surveys on Plant Disease Detection}
\label{sec:Existing Surveys on Plant Disease Detection}
Numerous survey studies have reviewed the literature on plant disease detection, primarily emphasizing the development and performance of deep learning models. While these reviews provide valuable insights into architectural advances and classification performance, most of them treat datasets merely as experimental resources rather than as fundamental analytical components of the research pipeline. A closer examination of the literature reveals several dataset-related gaps that remain insufficiently explored. First, many surveys provide limited discussion on the structural properties of datasets, including class distribution, dataset scale, acquisition conditions, and representational diversity. In particular, the imbalance between disease categories, variations in dataset composition across crops, and inconsistencies in the inclusion of healthy classes are rarely analyzed in a systematic manner. Second, a large proportion of existing studies rely heavily on laboratory-collected datasets with controlled backgrounds and simplified imaging conditions. The limited availability of field-acquired datasets restricts the ability to evaluate model robustness under realistic agricultural environments. Furthermore, current surveys often overlook how dataset characteristics influence the behavior and evaluation of deep learning models. Environmental factors such as lighting variation, background complexity, and climate-related disease expression are also seldom incorporated into dataset analyses.

These limitations highlight the need for a dataset-centric perspective when reviewing plant disease detection research. Instead of focusing solely on model architectures, this review examines the structural composition of widely used datasets, analyzes imbalance patterns and acquisition environments, and investigates how dataset characteristics influence model learning behavior and reported performance.

\par
The field of plant disease detection through deep learning has been extensively reviewed in numerous review articles. A notable example is the recent survey by Upadhyay et al. \cite{upadhyay2025deep}, published in 2025 provided a comprehensive review of deep learning and computer vision models for plant disease detection, analyzing 278 research articles. While the study extensively discussed imaging modalities, CNN architectures, object detection frameworks, and advanced vision models, it did not perform a dataset-centric structural evaluation. Another review article by Yue et al. \cite{yue2025deep} provided a comprehensive review of deep learning techniques for plant disease and pest recognition, including CNN-based classification, two-stage and single-stage object detection, hyperspectral asymptomatic detection, and lightweight model design. Pacal et al. \cite{pacal2024systematic} presented a comprehensive systematic review of deep learning techniques for plant disease classification, detection, and segmentation. While the study summarizes commonly used datasets and reports dataset sizes and accessibility, it does not conduct structural analysis of dataset composition, class imbalance severity, or distribution bias. Yilmaz et al. \cite{yilmaz2025advancements}, published in 2025, present a comprehensive systematic review of computer vision, IoT, and deep learning approaches in smart agriculture, evaluating 198 studies and discussing dataset limitations qualitatively, but without performing quantitative structural analysis of data diversity or class imbalance. Bhargava et al. \cite{bhargava2024plant} provide a comprehensive review of machine learning, deep learning, hyperspectral imaging, few-shot learning, and molecular diagnostic techniques for plant disease detection, discussing model architectures and datasets qualitatively but without conducting structural dataset analysis Another review paper Ngugi et al. \cite{ngugi2024revolutionizing}, emphasizes the important role of deep learning algorithms, especially convolutional neural networks (CNN), in in enhancing the precision and efficiency of identifying and diagnosing crop diseases by analysing 155 relevant articles, including publications, conference proceedings, review articles etc. Another review article by Yao et al. \cite{yao2024deep}, published in 2024, presents a comprehensive study on the application of deep learning techniques for plant identification and disease classification using leaf images and also highlights the importance of data availability and the need for diverse datasets by analysing 75 related works and 17 different plant disease datasets. Similarly, Jianping et al. \cite{yao2023machine}, in their 2023 survey, examined over 90 review articles spanning machine learning, deep learning, and augmented learning and summarized 13 publicly available datasets specifically for smart agriculture applications in leaf disease detection. Shafik et al. \cite{shafik2023systematic}, published in 2023, provide a comprehensive overview of the current state of research in the field of plant disease detection by evaluating various classification techniques and highlighting the importance of datasets in training detection models by analysing 176 research works published between 2012 and 2022. In another 2023 survey, Sunil et al. \cite{sunil2023systematic} analyzed 160 research studies focused on employing deep learning techniques for the detection and classification of plant diseases, providing an analysis of various contemporary deep learning techniques and the datasets available for plant leaf diseases. Bhagat et al. \cite{bhagat2022comprehensive}, also from 2023, reviewed 179 papers focusing on leaf disease and plant species classification. Thakur et al. \cite{thakur2022trends}, published in 2022, analysed 147 related articles highlighting the scarcity of publicly available datasets and robust methodologies to enhance the applicability of these techniques in field conditions. Bangari et al. \cite{bangari2022survey}, in their 2022 publication, discuss different classification models concentrated more on potato leaf diseases. Tugrul et al. \cite{tugrul2022convolutional}, published in 2022, analysed 100 relevant research works focusing on the effectiveness of CNNs in agriculture and emphasised the importance of suitable datasets and methodologies for enhancing disease identification. Similarly, Metre et al. \cite{metre2022reviewing} and Kumar et al. \cite{kumar2022retracted}, published in 2022, reviewed over 30 articles on machine learning and deep learning for plant leaf disease diagnosis. Additionally, comprehensive reviews by Applalanaidu et al. \cite{applalanaidu2021review} and Agarwal et al. \cite{agarwal2021plant} further explored the plant leaf diseases diagnosis and classification by deep learning techniques. Table \ref{tab: Comparison of existing survey papers.} below presents a critical comparison of existing survey papers, highlighting the limited attention given to dataset characteristics, acquisition settings, class imbalance, and environmental variability.
\par
Existing surveys in plant disease detection have largely concentrated on deep learning architectures and reported accuracy metrics, with comparatively limited critical examination of the datasets underpinning these models. This review diverges from prior work by positioning dataset structure as the central analytical axis. We introduce a structured taxonomy of plant disease datasets and conduct quantitative imbalance analysis using a normalized imbalance ratio (NIR) to systematically evaluate class distribution skewness across single-plant and multi-plant datasets. Beyond descriptive comparison, we analyze how structural dataset characteristics influence model selection, augmentation strategies, and reported performance outcomes. By explicitly linking laboratory bias, environmental variability, and minority-class suppression to model generalization challenges, this study advances a deployment-oriented, data realism framework for plant disease detection in precision agriculture.

\begin{table}[h]
\centering
\caption{Comparison of existing survey papers on plant disease detection highlighting dataset-centric research gaps.}
\vspace*{-3mm}
\label{tab: Comparison of existing survey papers.}
\resizebox{\textwidth}{!}{
\begin{tabular}{c c c c c c c c c c c c}
\hline

\textbf{\makecell{Ref}} & 
\textbf{\makecell{Year}} & 
\textbf{\makecell{Model \\ Coverage}} & 
\textbf{\makecell{Architecture \\ Analysis}} & 
\textbf{\makecell{Dataset \\ Inclusion}} & 
\textbf{\makecell{Dataset \\ Analysis}} & 
\textbf{\makecell{Data \\ Diversity}} & 
\textbf{\makecell{Class \\ Imbalance}} & 
\textbf{\makecell{Acquisition \\ Setting}} & 
\textbf{\makecell{Dataset \\ Availability}} & 
\textbf{\makecell{Env \\ Consideration}} & 
\textbf{\makecell{Dataset \\ Limitations}} \\

\hline
\textbf{\cite{upadhyay2025deep}} & 2025 & \checkmark & Moderate & \checkmark & Limited & \xmark & \xmark & \xmark & \xmark & \xmark & \xmark \\ \hline
\textbf{\cite{yilmaz2025advancements}} & 2025 & \checkmark & Moderate & \checkmark & Limited & \xmark & \xmark & \xmark & \xmark & \xmark & \xmark \\ \hline
\textbf{\cite{yue2025deep}} & 2025 & \checkmark & Comprehensive & \checkmark & Limited & \xmark & \xmark & \xmark & \xmark & \xmark & \xmark \\ \hline
\textbf{\cite{pacal2024systematic}} & 2024 & \checkmark & Comprehensive & \checkmark & Limited & \xmark & \xmark & \xmark & \xmark & \xmark & \xmark \\ \hline
\textbf{\cite{yao2024deep}} & 2024 & \checkmark & Moderate & \checkmark & Limited & \xmark & \xmark & \xmark & \xmark & \xmark & \xmark \\ \hline
\textbf{\cite{bhargava2024plant}} & 2024 & \checkmark & Moderate & \checkmark & Limited & \checkmark & \xmark & \xmark & \checkmark & \xmark & \xmark \\ \hline
\textbf{\cite{ngugi2024revolutionizing}} & 2024 & \checkmark & Moderate & \checkmark & Limited & \xmark & \xmark & \xmark & \xmark & \xmark & \xmark \\ \hline
\textbf{\cite{yao2023machine}} & 2023 & \checkmark & Limited & \checkmark & Limited & \checkmark & \xmark & \xmark & \checkmark & \xmark & \xmark \\ \hline
\textbf{\cite{shafik2023systematic}} & 2023 & \checkmark & Moderate & \checkmark & Limited & \xmark & \xmark & \checkmark & \xmark & \xmark & \xmark \\ \hline
\textbf{\cite{sunil2023systematic}} & 2023 & \checkmark & Comprehensive & \checkmark & Moderate & \checkmark & \xmark & \xmark & \checkmark & \xmark & \xmark \\ \hline
\textbf{\cite{bhagat2022comprehensive}} & 2023 & \checkmark & Moderate & \checkmark & Moderate & \xmark & \xmark & \xmark & \xmark & \xmark & \xmark \\ \hline
\textbf{\cite{thakur2022trends}} & 2022 & \checkmark & Moderate & \checkmark & Limited & \xmark & \xmark & \checkmark & \xmark & \xmark & \xmark \\ \hline
\textbf{\cite{metre2022reviewing}} & 2022 & \checkmark & Limited & \checkmark & Limited & \xmark & \xmark & \xmark & \xmark & \xmark & \xmark \\ \hline
\textbf{\cite{tugrul2022convolutional}} & 2022 & \checkmark & Limited & \checkmark & Limited & \xmark & \xmark & \xmark & \checkmark & \xmark & \xmark \\ \hline
\textbf{\cite{bangari2022survey}} & 2022 & \checkmark & Moderate & \checkmark & Limited & \xmark & \xmark & \xmark & \xmark & \xmark & \xmark \\ \hline
\textbf{\cite{applalanaidu2021review}} & 2021 & \checkmark & Limited & \checkmark & Limited & \xmark & \xmark & \xmark & \xmark & \xmark & \xmark \\ \hline
\textbf{\cite{agarwal2021plant}} & 2021 & \checkmark & Limited & \checkmark & Limited & \checkmark & \xmark & \xmark & \xmark & \xmark & \xmark \\ \hline

\rowcolor{gray!15}
\textbf{This Survey} & -- & \checkmark & Moderate & \checkmark & \textbf{Comprehensive} & \checkmark & \checkmark & \checkmark & \checkmark & \checkmark & \checkmark \\

\hline
\end{tabular}}

\begin{minipage}{17cm}
\small\itshape{In this table, "\checkmark" indicates that a topic is addressed, "\xmark" denotes its absence. The depth of analysis is categorized as Limited, Moderate, or Comprehensive, reflecting the breadth of coverage and level of critical evaluation provided in each survey.
\par The table presents a structured comparison of existing review studies based on key dimensions, including model coverage, architectural analysis, dataset inclusion and analysis, data diversity, class imbalance, acquisition settings, dataset availability, environmental considerations, and reported dataset limitations. This comparison highlights the comparatively limited attention given to dataset-centric factors in prior surveys and underscores the need for a more comprehensive analysis to support the development of robust and real-world-ready plant disease detection systems.}
\end{minipage}
\end{table}

\par
As shown in Table \ref{tab: Comparison of existing survey papers.}, most existing surveys primarily emphasize model architectures while offering only limited analysis of dataset characteristics. Critical aspects such as structured dataset taxonomy, acquisition settings, class imbalance, and environmental variability remain comparatively underexplored. These observations highlight the need for a comprehensive dataset-centric review to better support the development of robust and deployable plant disease detection systems.

\section{Systematic Literature Search and Selection Strategy}
\label{sec:Systematic Literature Search and Selection Strategy}
To ensure comprehensive coverage of the literature, a systematic search strategy was adopted to identify relevant studies on deep learning–based plant disease detection and associated datasets. An insightful analysis of data from multiple sources over the past decade reveals significant phases of advancement, growth, and maturity within this dynamic field. The literature search was conducted across major scientific databases, including the ACM Digital Library, IEEE Xplore, Springer, Scopus, Google Scholar and others such as Wiley, Web of Science, and arXiv. These resources provide a comprehensive foundation for identifying significant research contributions in the realm of plant disease diagnosis. A systematic approach was employed to retrieve relevant research papers, ensuring a thorough exploration of the field of plant disease detection. Initially, the search strategy involved the use of a diverse set of search terms, including but not limited to “crop disease detection,” “crop disease prediction,” “plant disease identification,” “crop disease classification,” “plant disease datasets,” “single plant disease datasets,” “multi-plant disease datasets,”. The initial search strings yielded a broad range of results, many of which were not directly relevant to the specific focus on plant disease detection using deep learning techniques. The generic nature of these results highlighted the need for a more refined search query. A structured Boolean query was developed to capture studies related to plant disease detection, deep learning methodologies, and dataset characteristics. The refined query was:
\par
\vspace{10pt}
\par
\begin{center}
(“plant disease” OR “crop disease” OR “leaf disease” OR “fruit disease”) 
AND (“deep learning” OR “computer vision” OR “convolutional neural network” OR “CNN”) AND (“prediction” OR “detection” OR “classification” OR “identification”) AND 
(“dataset” OR “data set” OR “image dataset” OR “benchmark” OR “database”)
\end{center}
\par
\vspace{10pt}

\par
The query was applied to titles, abstracts, and keywords wherever supported, with minor adaptations made to accommodate database-specific indexing requirements. The initial search returned a total of 6784 publications across all databases, as illustrated in \ref{tab: Number of articles}. Duplicate studies appearing across multiple databases were removed, resulting in 3664 unique articles. These studies were subsequently screened based on their titles to remove clearly irrelevant works, reducing the set to 820 papers. After eliminating duplicates, a set of predefined research selection criteria was applied to determine which articles to include or exclude in the review, as shown in Table \ref{tab: Inclusion and Exclusion Criteria for Study Selection..}. The review prioritizes studies that provide explicit discussion of plant disease datasets, as data characteristics form the central focus of this work.

\begin{table}[h!]
\footnotesize
\centering
\caption{Number of articles retrieved from each database.}
\label{tab: Number of articles}
\vspace*{-3mm}
\begin{tabular}{lc}
\hline
\textbf{Database} & \textbf{Retrieved Articles} \\
\hline
ACM Digital Library & 50 \\
Scopus & 1654 \\
SpringerLink & 1328 \\
IEEE Xplore & 1742 \\
Google Scholar & 1934 \\
Other sources & 76 \\
\hline
Total & 6784 \\
\hline
\end{tabular}
\end{table}

\begin{table}[h!]
\footnotesize
\centering
\caption{Inclusion and Exclusion Criteria for Study Selection.}
\vspace*{-3mm}
\label{tab: Inclusion and Exclusion Criteria for Study Selection..}
\begin{tabular}{p{7.0cm}p{7.0cm}}
\hline
    \textbf{Considered articles} & \textbf{Unconsidered articles} \\ \hline
    Studies on plant disease detection using deep learning techniques. & Studies on general agriculture without focus on disease detection or deep learning applications \\ \hline
    Articles that describe or analyze plant disease datasets relevant to deep learning models. & Articles focused on traditional plant pathology methods without using datasets or DL models \\ \hline
    Research that includes single-plant or \\ multi-plant datasets for identifying diseases in crops or plants & Studies that do not provide or analyze datasets or only mention deep learning models in theoretical terms. \\ \hline
    Articles discussing image datasets specifically collected for plant disease classification & Research on crop yield prediction or growth monitoring without a focus on disease detection. \\ \hline
    Articles discussing the comparison of DL and datasets in plant disease detection across different plant species. & Papers using non-disease-related datasets (e.g., plant growth data, soil composition data) without disease-specific context. \\ \hline
    Articles published between 2019 to 2025. & The original language was not English \\ \hline
\end{tabular}
\end{table}

\begin{figure}[H]
\centering
\centering
    \includegraphics[width=1.0\textwidth]{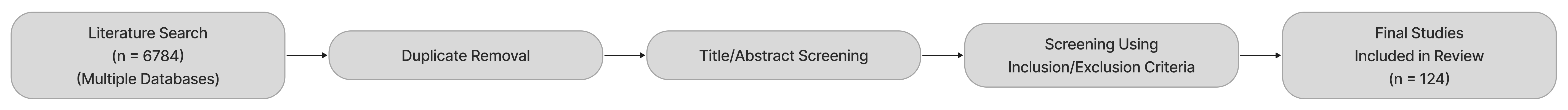}
    \vspace*{-2mm}
    \caption{Overview of the systematic literature search and study selection process.}
    \label{fig: Overview of the systematic literature search and study selection process.}
\end{figure}

After applying the selection criteria, several articles were excluded from the study. These articles were removed because they did not focus on deep learning models or relevant disease datasets. Specifically, they either lacked the use of convolutional neural networks or did not utilize appropriate plant or crop disease datasets. Additionally, some research works were excluded due to containing similar information, which did not contribute new insights to the study. Finally, full text evaluation based on predefined inclusion and exclusion criteria resulted in 124 research articles being selected for detailed analysis in this review. The study selection process follows a structured screening procedure consisting of identification, duplicate removal, title screening, abstract screening, and full-text eligibility assessment. Although the study does not strictly follow the PRISMA reporting framework, the adopted process reflects similar systematic principles commonly used in systematic mapping studies to ensure transparency and methodological rigor. The workflow used to filter and select relevant studies is illustrated in Figure~\ref{fig: Overview of the systematic literature search and study selection process.}. While the review adopts a structured search methodology, the emphasis remains on providing a critical dataset-centric synthesis rather than conducting a statistical meta-analysis.

\section{Taxonomy of Plant Disease Datasets}
\label{sec:Taxonomy of Plant Disease Datasets}
The effectiveness and generalizability of deep learning–based plant disease detection systems are fundamentally shaped by the quality and diversity of datasets. These models rely heavily on large volumes of diverse and accurately labeled data to learn and generalize effectively. Although plenty of plant disease datasets have been developed over the last decade, they are hugely different regarding the plants included, the acquisition environment, accessibility, scale, and composition of the diseases. Such differences bring about an effect not only on what is claimed as model performance but also on how robust and truly applicable the real-world frameworks are for diseased detection. Thus, the need for a structured taxonomy of plant disease datasets arises to help organize the research landscape, clarify structural distinctions, and provide an outline of critical analyses. Therefore, the categorization of available datasets regarding single/multi-species coverage, laboratory/field acquisition, and public/private availability is here undertaken in order to provide an ordered overview of already existing data resources and to set the stage for a subsequent discussion of their limitations and implications in model development.

\subsection{Importance of Dataset Structure in Deep Learning–Based Disease Detection}
In deep learning-based plant disease detection, the dataset structure gives considerable influence on the behavior, performance, and generalization capability of the model. Deep neural networks learn representations from data, which is contrary to traditional rule-based systems; hence, factors such as dataset distribution, diversity, and composition strongly influence what the model learns from the data and how it performs in previously unseen environments. The overall complexity of the learning task and the robustness of the produced model, therefore, depend on the number of plant species represented, the diversity of disease classes, balance of class distribution, acquisition condition (lab vs. field), and scale of the dataset. 

\par Dataset structure not only influences predictive accuracy but also model bias, sensitivity to minority disease classes, and robustness against environmental variations. For example, datasets collected under controlled lab conditions are likely to lead excellent classification accuracy; however, this does not translate well to real-field conditions with occlusions, changing illumination, and a heterogeneous background. This also applies if class distribution within a dataset is highly imbalanced: a model might then be biased towards classifying instances in categories forming the majority of the dataset. Overall accuracy may be high, but detection will fail on minority classes.  Knowing the dataset structure becomes important for setting reported results in context and further understanding the applicability of results in plant disease detection systems. 

\par A systematic examination of the attributes of the datasets allows one to learn general recurring patterns across studies, revealing structural gaps that may limit progress toward reliable real-world applications. Hence, there is a need to analyze datasets via well-defined taxonomic dimensions so that researchers understand better how data characteristics drive modeling strategies and identify areas requiring further dataset design refinement to enable the next generation of precision agriculture systems. Table \ref{tab:master_dataset} summarizes a structurally consolidated overview of datasets commonly used in deep learning-based plant disease detection research, aiming to systemically organize the diverse landscape of plant disease datasets. Rather than focusing on experimental performance, the table summarizes fundamental dataset attributes, including plant scope (single versus multi-species), acquisition setting (laboratory versus field), accessibility (public versus proprietary), and core structural characteristics such as the number of crops, disease classes, and total images. This structural representation enables a comparative understanding of dataset design choices and highlights variations that may influence model development and evaluation. By consolidating these attributes into a unified framework, the table serves as a reference point for the subsequent taxonomy and critical analysis presented in the following sections.

\begin{table}[H]
\centering
\caption{Structural overview of plant disease datasets utilized in deep learning--based detection research.}
\vspace*{-3mm}
\label{tab:master_dataset}
\resizebox{\textwidth}{!}{
\begin{tabular}{c l c c c c c c}
\hline
\textbf{Ref} &
\textbf{Dataset} &
\textbf{Year} &
\textbf{Acquisition} &
\textbf{Availability} &
\textbf{No. of Crops} &
\textbf{No. of Classes} &
\textbf{No. of Images} \\
\hline

\multicolumn{8}{c}{\textbf{Single-Plant Datasets}} \\
\hline

\cite{jiang2019real} & ALDD Apple Dataset & 2019 & Lab & Private & 1 & 5 & 26,377 \\
\cite{wu2019identification} & Soybean Dataset & 2019 & Lab & Private & 1 & 7 & 1,470 \\
\cite{dananjayan2022assessment} & Citrus Dataset & 2022 & Lab & Private & 1 & 3 & 2,684 \\
\cite{joshi2022ricebios} & Rice Dataset & 2022 & Lab & Private & 1 & 3 & 442 \\
\cite{khan2022deep} & Apple Plant Dataset & 2022 & Lab & Private & 1 & 10 & 5,201 \\
\cite{chug2023novel} & Tomato Dataset & 2023 & Field & Private & 1 & 2 & 155 \\
\cite{khatoon2021image} & Tomato Dataset & 2021 & Field & Private & 1 & 13 & 15,335 \\
\cite{ashwinkumar2022automated} & Tomato Dataset & 2022 & Lab & Private & 1 & 5 & 5,512 \\
\cite{prajapati2017detection} & Rice Dataset & 2017 & Lab & Public & 1 & 3 & 120 \\
\cite{bevers2022pictures} & Soybean Dataset & 2022 & Lab & Public & 1 & 8 & 9,648 \\
\cite{syed2022citrus} & Citrus Dataset & 2022 & Lab & Public & 1 & 4 & 597 \\
\cite{tassis2021deep} & Coffee Dataset & 2021 & Field & Private & 1 & 4 & 390 \\
\cite{adem2023sugar} & Sugar Dataset & 2023 & Field & Private & 1 & 4 & 155 \\
\cite{elfatimi2022beans} & Beans Dataset & 2022 & Lab & Public & 1 & 3 & 1,296 \\
\cite{liu2022ds} & Citrus Dataset & 2022 & Lab & Private & 1 & 5 & 3,086 \\
\cite{seetharaman2022leaf} & Banana Dataset & 2022 & Field & Private & 1 & 4 & 1,875 \\
\cite{thet2020grape} & Grape Dataset & 2020 & Lab & Private & 1 & 6 & 6,000 \\
\cite{rangarajan2022detection} & Wheat Dataset & 2022 & Lab & Private & 1 & 2 & 2,721 \\
\cite{perez2022exploring} & Grape Dataset & 2022 & Lab & Private & 1 & 3 & 72 \\

\hline
\multicolumn{8}{c}{\textbf{Multi-Plant Datasets}} \\
\hline

\cite{hughes2015open} & PlantVillage Dataset & 2015 & Lab & Public & 14 & 38 & 54,309 \\
\cite{singh2020plantdoc} & PlantDoc Dataset & 2020 & Mixed & Public & 13 & 30 & 2,569 \\
\cite{saleem2022performance} & NZDL Fruit Dataset & 2022 & Field & Public & 5 & 20 & 3,545 \\
\cite{liang2019pd2se} & AI Challenger Dataset & 2019 & Lab & Private & 10 & 45 & 26,747 \\
\cite{tugrul2022convolutional} & Turkey Plant Dataset & 2022 & Field & Public & 7 & 15 & 4,447 \\
\cite{chouhan2019data} & Multi Plant Dataset & 2019 & Lab & Public & 11 & 22 & 4,079 \\
\cite{moupojou2023fieldplant} & FieldPlant Dataset & 2023 & Field & Public & 3 & 27 & 8,629 \\
\cite{liu2021plant} & Plant Disease Dataset-271 (PDD271) & 2021 & Field & Private & 38 & 271 & 220,592 \\
\cite{jiang2021recognition} & UCI-Rice and Wheat Dataset & 2021 & Lab & Public & 2 & 5 & 200 \\
\cite{shovon2023plantdet} & PlantDet Dataset (Rice and Betel) & 2023 & Mixed & Private & 2 & 8 & 3,710 \\
\cite{brown30plant} & Multispectral Image Dataset & 2023 & Field & Private & 5 & 10 & 3,016 \\

\hline
\end{tabular}
}
\begin{minipage}{17cm}
\small\itshape{This table presents a structural overview of plant disease datasets utilized in deep learning–based detection research. 
\par The table summarizes acquisition setting, availability, crop diversity, class composition, and dataset scale. Some datasets include multispectral \cite{shovon2023plantdet,brown30plant} or hyperspectral imagery \cite{rangarajan2022detection,perez2022exploring},where the precise number of images or the picture modality may vary.}
\end{minipage}
\end{table}

\par 
While Table~\ref{tab:master_dataset} provides a consolidated structural overview of plant disease datasets, a more systematic classification is necessary to better understand their inherent variations and research implications. The datasets summarized above differ not only in scale and composition but also in plant coverage, acquisition environment, and accessibility. To facilitate structured comparison and highlight these distinctions, the following subsections categorize existing datasets according to key taxonomic dimensions, beginning with plant scope, followed by acquisition setting and availability. This taxonomy-oriented organization enables a clearer understanding of dataset design patterns and prepares the foundation for subsequent critical analysis.

\subsection{Taxonomy by Plant Scope: Single vs Multi-Species Datasets}
\begin{figure}[H]
    \centering
    \includegraphics[width=0.8\textwidth]{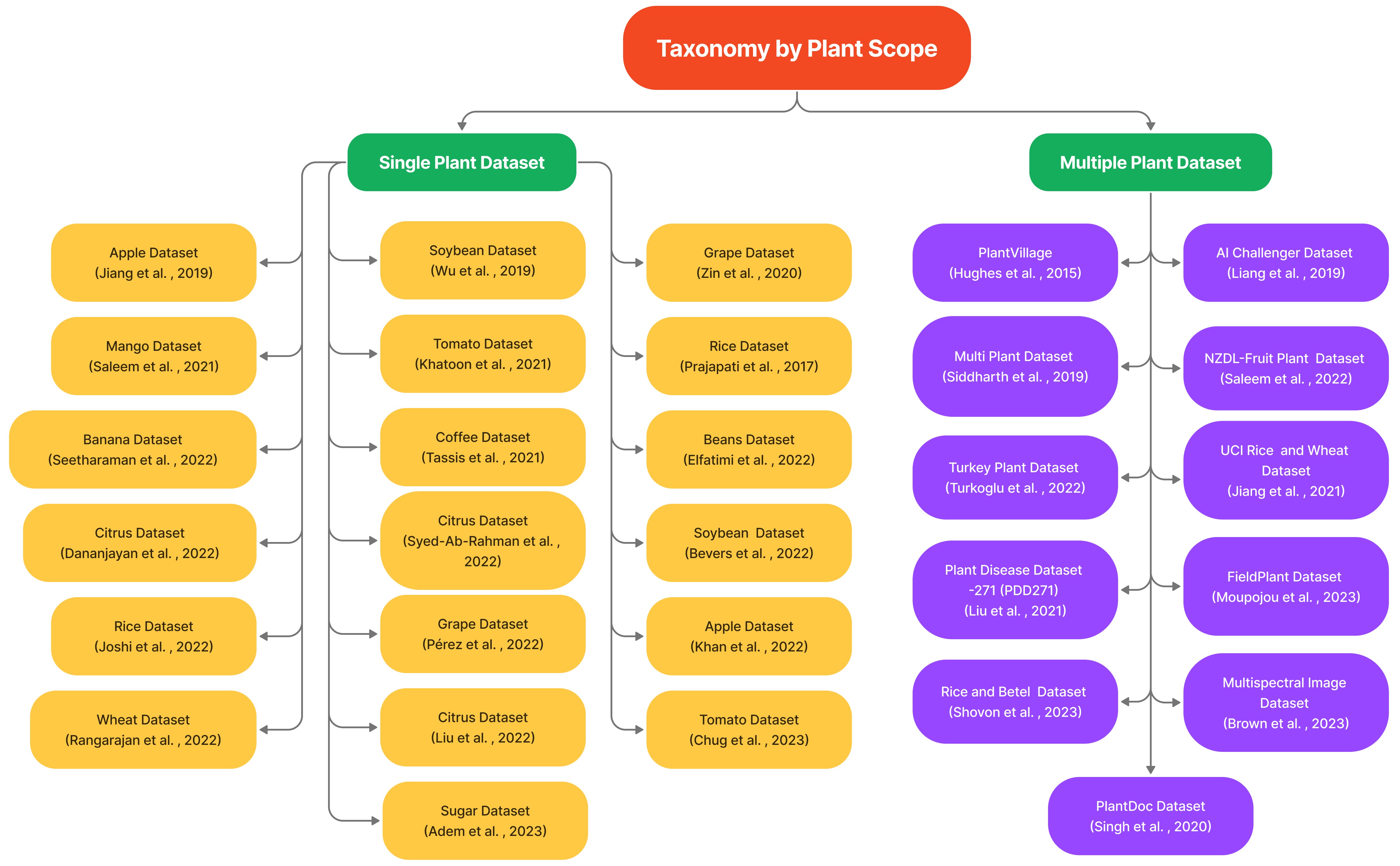}
    \vspace{-3mm}
    \caption{Taxonomy of plant disease datasets based on plant scope.}
    \label{fig: Taxonomy of plant disease datasets based on plant scope.}
\end{figure}

Plant disease datasets differ considerably in terms of the number of crop species they include. This distinction is important because the scope of plant coverage directly affects the complexity of the learning problem and the potential generalizability of the resulting models. To visually summarize the plant-scope categorization derived from Table~\ref{tab:master_dataset}, Figure~\ref{fig: Taxonomy of plant disease datasets based on plant scope.} presents the taxonomy of plant disease datasets based on crop coverage.

\subsubsection{Single-Plant Datasets}
Single-plant datasets focus on disease detection within a specific crop species. Typically, these datasets contain images of one plant type with several disease categories, along with healthy samples. Because the plant morphology remains consistent across the dataset, the variation is primarily disease-specific rather than species-specific. This often makes the classification task more structured and easier for deep learning models to optimize.

\par
Such datasets are commonly used in crop-focused research, where the objective is to design and fine-tune a model for a particular agricultural application. Their relatively controlled variability can lead to high reported accuracy, especially when the images are collected in laboratory conditions. However, this narrow focus may limit the model’s ability to generalize across different crops or mixed farming environments.

\par
As shown in the Table~\ref{tab:master_dataset}, many single-plant datasets are privately collected and often lab-based, although some field-acquired and publicly available datasets also exist. While these datasets are valuable for targeted disease diagnosis, their limited plant diversity highlights the need for broader representations when aiming for scalable, real-world deployment.

\subsubsection{Multi-Plant Datasets}
In contrast, multi-plant datasets include multiple crop species within the same dataset. These datasets are inherently more diverse, as they incorporate variations in leaf structure, texture, disease symptoms, and environmental conditions across different crops.

From a modeling perspective, multi-plant datasets introduce a higher level of complexity. The model must not only differentiate between diseases but also implicitly learn plant-specific visual characteristics. Although this increases the difficulty of the classification task, it also enhances the potential for building more generalized disease detection systems suitable for practical agricultural scenarios.

Multi-plant datasets are particularly relevant for precision agriculture systems, where a single diagnostic model may need to operate across different crops in the same farming ecosystem. However, increased diversity often results in uneven class distributions, especially when some crop disease combinations are underrepresented. These structural variations will be examined in more detail in the subsequent analysis section.

\subsection{Taxonomy by Acquisition Setting: Laboratory vs Field Datasets}
\begin{figure}[H]
    \centering
    \includegraphics[width=0.79\textwidth]{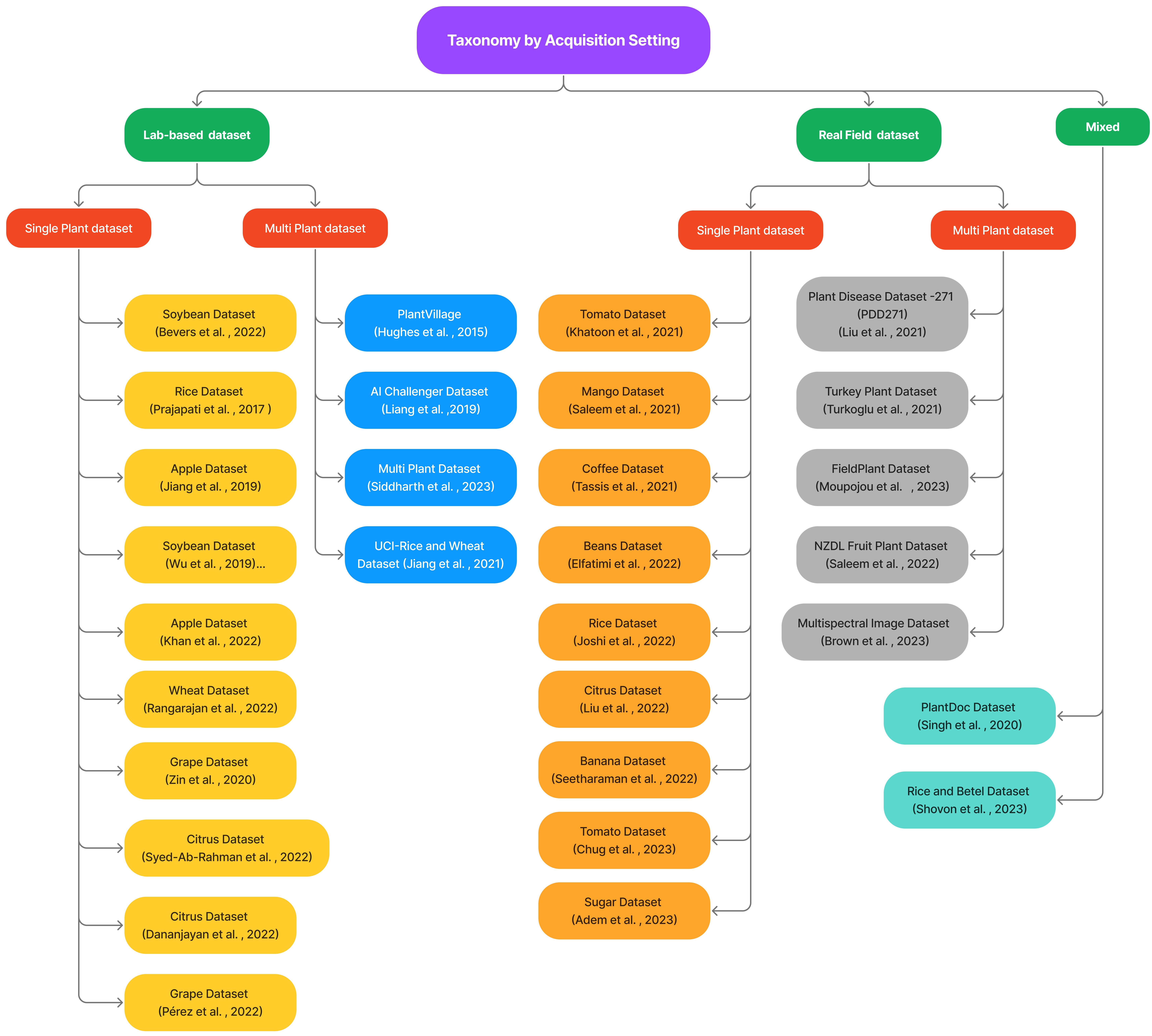}
    \vspace*{-3mm}
    \caption{Taxonomy of plant disease datasets based on acquisition environment.}
    \label{fig: Taxonomy of plant disease datasets based on acquisition environment.}
\end{figure}

Beyond plant scope, datasets can also be distinguished based on the environment in which images were acquired. The acquisition setting significantly influences image variability, background complexity, and environmental realism. Similarly, based on the acquisition attributes summarized in Table~\ref{tab:master_dataset}, Figure~\ref{fig: Taxonomy of plant disease datasets based on acquisition environment.} illustrates the taxonomy of datasets according to their image collection environment.

\subsubsection{Laboratory-Based Datasets}
Laboratory-based datasets are typically collected under controlled imaging conditions characterized by uniform backgrounds, stable lighting, and minimal occlusions. These controlled environments reduce noise and variability, enabling models to achieve high classification accuracy during experimental evaluation.

However, such controlled conditions may not accurately reflect real agricultural environments. As a result, models trained exclusively on laboratory datasets may experience performance degradation when deployed in field settings where lighting, weather, and background variability are unpredictable.

\subsubsection{Field-Based Datasets}
Field-based datasets are collected directly from agricultural environments under natural lighting and environmental conditions. These datasets capture real-world variability, including occlusions, heterogeneous backgrounds, shadow effects, and disease expression at different growth stages.

Although field datasets provide greater ecological validity, they often introduce higher complexity and potential noise. Consequently, training models on field data may require larger dataset sizes and more robust learning strategies to achieve stable performance.

\subsubsection{Mixed Acquisition Datasets}
Some datasets incorporate both laboratory and field-acquired images, creating mixed acquisition environments. These datasets attempt to balance controlled clarity with environmental realism. While they may improve diversity, inconsistencies in acquisition conditions can introduce domain variability that influences model training behavior.

\subsection{Taxonomy by Accessibility: Public vs Proprietary Datasets}
\begin{figure}[H]
    \centering
    \includegraphics[width=0.8\textwidth]{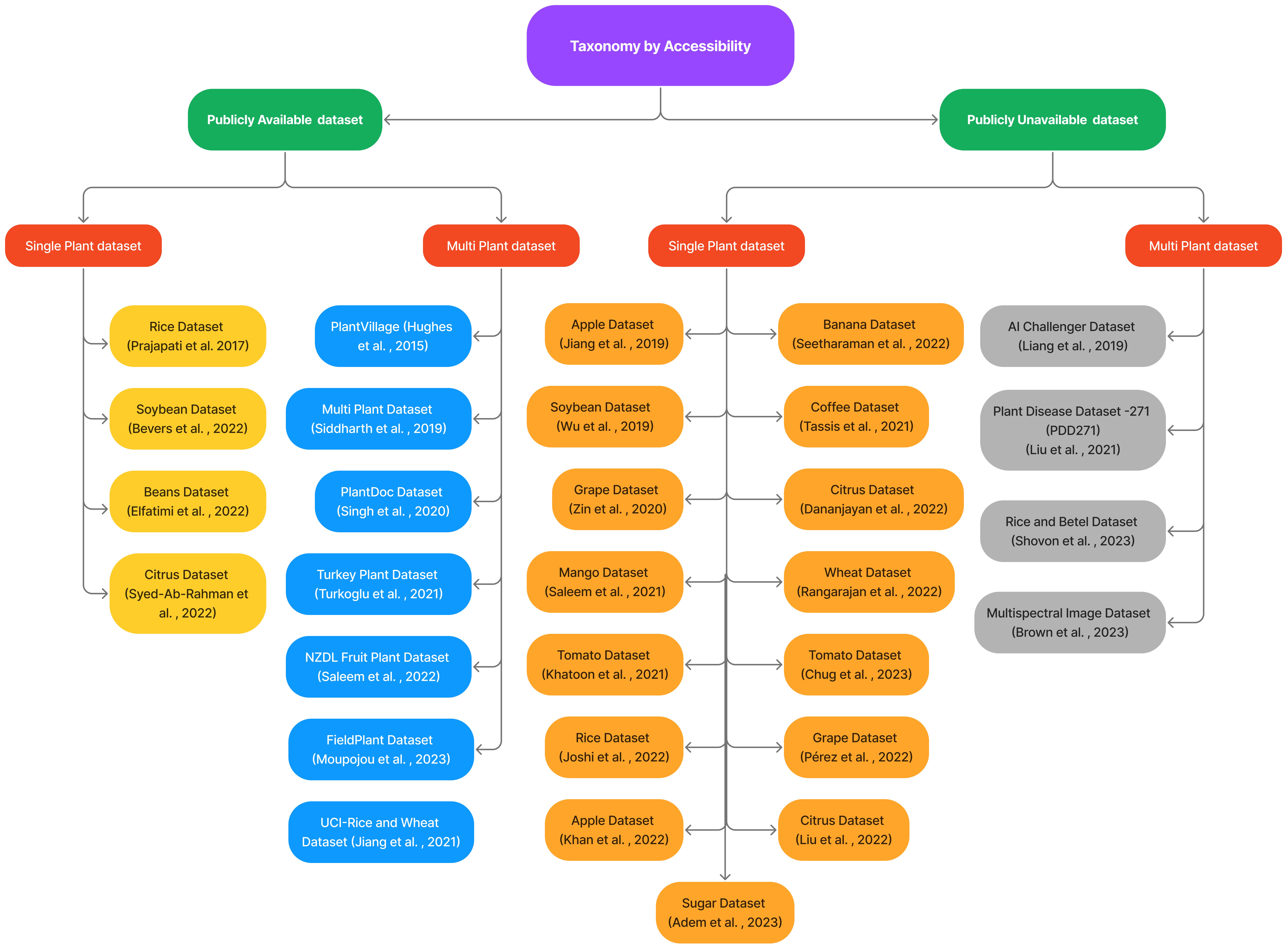}
    \vspace*{-3mm}
    \caption{Taxonomy of plant disease datasets based on  dataset accessibility.}
    \label{fig: Taxonomy of plant disease datasets based on  dataset accessibility.}
\end{figure}

Accessibility represents another critical structural dimension influencing research reproducibility and comparative benchmarking.
Similarly, based on the availability attributes summarised in Table~\ref{tab:master_dataset}, Figure~\ref{fig: Taxonomy of plant disease datasets based on  dataset accessibility.} illustrates the taxonomy of datasets according to their accessibility.

\subsubsection{Publicly Available Datasets}
Public datasets are openly accessible to the research community and play a fundamental role in benchmarking, model comparison, and methodological standardization. They enable reproducible research and facilitate fair evaluation across different modeling approaches. To facilitate reproducibility and comparative evaluation, Table~\ref{tab:public_datasets} lists publicly available plant disease datasets along with their corresponding access links. While the structural characteristics of these datasets have been summarized in Table~\ref{tab:master_dataset}, the present table specifically emphasizes accessibility, enabling researchers to readily locate and utilize open data resources for benchmarking and model development.

\begin{table}[H]
\centering
\footnotesize
\caption{Publicly available plant disease datasets with access information.}
\vspace*{-3mm}
\label{tab:public_datasets}
\begin{tabular}{l l l}
\hline
\textbf{Ref} & \textbf{Dataset Name} & \textbf{Access Link} \\
\hline

\multicolumn{3}{l}{\textbf{Publicly Available Single-Plant Datasets}} \\ \hline

\cite{prajapati2017detection} & Rice Dataset & \url{https://archive.ics.uci.edu/ml/datasets/Rice+Leaf+Diseases} \\

\cite{bevers2022pictures} & Soybean Dataset & \url{https://datadryad.org/stash/dataset/doi:10.5061/dryad.41ns1rnj3} \\

\cite{elfatimi2022beans} & Beans Dataset & \url{https://github.com/AI-Lab-Makerere/ibean/tree/master} \\

\cite{syed2022citrus} & Citrus Plant Dataset & \url{https://www.kaggle.com/datasets/drlisbeek/citrus-leaves-prepared} \\

\hline
\multicolumn{3}{l}{\textbf{Publicly Available Multi-Plant Datasets}} \\ \hline

\cite{hughes2015open} & PlantVillage Dataset & \url{https://data.mendeley.com/datasets/tywbtsjrjv/1} \\

\cite{chouhan2019data} & Multi Plant Dataset & \url{https://data.mendeley.com/datasets/hb74ynkjcn/1} \\

\cite{singh2020plantdoc} & PlantDoc Dataset & \url{https://github.com/pratikkayal/PlantDoc-Dataset} \\

\cite{jiang2021recognition} & UCI-Rice and Wheat Dataset & \url{https://archive.ics.uci.edu/ml/datasets/Rice+Leaf+Diseases} \\

\cite{tugrul2022convolutional} & Turkey Plant Dataset & \url{https://github.com/mturkoglu23/PlantDiseaseNet} \\

\cite{saleem2022performance} & NZDL Fruit Dataset & \url{https://github.com/kmarnfri/NZDLPlantDisease-v1} \\

\cite{moupojou2023fieldplant} & FieldPlant Dataset & \url{https://universe.roboflow.com/plant-disease-detection/fieldplant} \\

\hline
\end{tabular}
\end{table}

However, many public datasets are laboratory-based, which may limit their representativeness for real-field applications.

\subsubsection{Proprietary or Private Datasets}
Private datasets are typically developed by individual research groups and are not publicly released. While these datasets may include valuable real-field images or specialized disease categories, their restricted accessibility limits reproducibility and cross-study comparison.

The limited availability of large-scale field datasets remains a major barrier to developing and validating robust plant disease detection systems.

\section{Critical Analysis of Plant Disease Datasets}
\label{sec:Critical Analysis of Plant Disease Datasets}
The taxonomy presented in the previous section helps organize plant disease datasets according to their scope, acquisition setting, and accessibility. While this structured classification provides clarity, it does not fully capture the deeper characteristics that influence how these datasets perform in real modeling scenarios. In practice, the true impact of a dataset lies not only in whether it is single or multi-plant, public or private, or lab or field-based, but also in how its internal structure shapes model learning.

Key aspects such as the number of disease categories, the distribution of images across classes, and the degree of balance or skewness within the dataset play a critical role in determining model behavior. For example, a dataset may include a large number of classes, yet still suffer from severe imbalance if certain diseases are underrepresented. Similarly, datasets collected in controlled environments may simplify classification tasks while masking challenges that arise in real-field deployment.

Although class imbalance is widely acknowledged in plant disease datasets, it is often described qualitatively or reported through raw per-class frequencies without a standardized comparative framework. Simple imbalance ratios, such as majority-to-minority class counts, are highly dataset-dependent and cannot be meaningfully compared across datasets with different scales or numbers of classes. To enable consistent structural comparison across heterogeneous plant disease datasets, a normalized metric is required that remains scale-invariant and interpretable across both small single-plant datasets and large multi-plant benchmarks.

To move beyond categorical description, a deeper examination of dataset composition is necessary. In this section, we analyze plant disease datasets in terms of structural attributes including the number of disease classes, total image count, and distribution of samples across categories and class imbalance, as uneven class distributions are frequently observed in both single-plant and multi-plant datasets. Understanding this
aspect is essential for evaluating the dataset’s balance and detecting potential biases, as these factors directly affect
the performance and generalization capabilities of deep learning models that rely on this data. 

A notable observation in our analysis is that certain datasets both single-plant and multi-plant contain only a single class category, either exclusively diseased samples or exclusively healthy samples. In these cases, no inherent inter-class distribution exists, making conventional imbalance computation infeasible. To enable consistent comparative analysis across all datasets, we introduce an auxiliary reference class during imbalance computation. This additional class is used solely for analytical purposes to allow the calculation of a normalized imbalance ratio (NIR), while preserving bounded interpretability. This adjustment enables fair structural comparison across datasets where certain disease categories or healthy classes are underrepresented or absent. This formulation ensures that the imbalance metric remains bounded between 0 and 1. Values close to 0 indicate relatively balanced distributions, whereas values approaching 1 reflect increasing skewness and dominance of certain classes.

\vspace{1em}
{\small
\begin{equation}
\text{Imbalance Ratio} = \frac{\text{Number of Images in the Largest Class} - \text{Number of Images in the Smallest Class}}{\text{Number of Images in the Largest Class}}
\label{eq:imbalance_ratio}
\end{equation}
}
\vspace{1em}
\par
Importantly, the NIR metric is not introduced as a replacement for performance-based evaluation metrics, but rather as a structural descriptor of dataset composition. By quantifying distribution skewness independently of model accuracy, NIR enables analysis of how dataset imbalance may predispose models toward majority-class dominance, augmentation dependency, or architectural compensation strategies.

\subsection{Structural Characteristics of Single-Plant Datasets}
Single-plant datasets are widely used in deep learning–based plant disease detection research. By focusing on a single crop species, these datasets offer a relatively controlled environment in which models learn to distinguish among disease symptoms within consistent plant morphology. This focused structure often simplifies the classification task and contributes to the high accuracy frequently reported in experimental studies.

\par
In a comprehensive lab condition, Jiang et al. \cite{jiang2019real} successfully developed an extensive apple dataset known as ALDD, which initially contained 2,029 distinct images. Through a series of data augmentation techniques, they expanded this dataset to an impressive total of 26,377 images. Wu et al. \cite{wu2019identification} curated a Soybean Leaf Disease Dataset, meticulously capturing images directly from the fields of Xiangyan Northeast Agricultural University in China. All images were acquired within a controlled environment. Dananjayan et al. \cite{dananjayan2022assessment} established a specialized dataset of Conghue citrus leaf consisting of 2,684 images, all sourced from carefully controlled settings in Guangzhou, China. Joshi et al. \cite{joshi2022ricebios} created a dataset of rice plant leaf featuring 442 images encompassing three distinct classes. These images were captured using a combination of smartphones and digital cameras taken in a natural farm field setting, thus reflecting real-world conditions. Khan et al. \cite{khan2022deep} made significant contributions with their apple plant dataset, which comprises images from nine different classes. This dataset captures the diversity of apple plants in a real-time farm environment. Khatoon et al. \cite{khatoon2021image} also focused on disease classification by creating a tomato disease dataset, sourcing images directly from farms under natural conditions. Ashwinkumar et al. \cite{ashwinkumar2022automated} developed a dataset of tomato plant leaves that features five distinct classes, incorporating a combination of images with and without backgrounds. Additionally, Chug et al. \cite{chug2023novel} generated another plant of tomato plant, which includes two disease classes characterized by a mixture of backdrop and non-backdrop images. The Rice Plant Leaf Dataset was generated by Prajapati et al. \cite{prajapati2017detection}, comprising images captured directly from rice farmland. Bevers et al. \cite{bevers2022pictures} developed an extensive soybean plant dataset, which contains 9,648 images that are categorized into eight different classes. These images have been collected over a two-year span during various stages of the plant's growth in Alabama, USA. The dataset is characterized by a variety of backgrounds. Furthermore, Jiang et al. \cite{jiang2021recognition} assembled a collection of wheat leaf disease samples for their Image Database for Agricultural Diseases and Pests Research at the Chinese Academy of Sciences. Elfatimi et al. \cite{elfatimi2022beans} created a beans dataset, with images captured in fields by the National Crops Resources Research Institute in Uganda. Tassis et al. \cite{tassis2021deep} generated the BRACOT dataset, which focuses on Arabica coffee trees. This dataset includes images collected from the mountainous regions of Marechal Floriano in the state of Espírito Santo, Brazil. A dataset of sugarbeet war curated by Adem et al. \cite{adem2023sugar}, with images showcasing the different growing stages of the sugar beet plants in a thriving farm field, all captured in daylight.

\par
However, a closer look at their internal composition reveals notable structural variations. The number of disease classes differs across datasets, and image counts are often unevenly distributed among categories. In several cases, certain disease classes dominate the dataset, while others contain significantly fewer samples. Additionally, some datasets exclude healthy images entirely, which further influences class distribution patterns.

\par
Such imbalance can affect training dynamics and bias model predictions toward majority classes. Table~\ref{tab:single_plant_frequency} summarizes the structural properties of representative single-plant datasets, including class diversity, image distribution, and normalized imbalance ratio. These characteristics provide important context for interpreting reported performance and assessing the extent to which results may generalize beyond controlled, crop-specific scenarios.

{\footnotesize
\begin{longtable}{p{3.5cm}p{4.0cm}p{2.5cm}p{2.5cm}p{2.0cm}}
\caption{Disease Distribution in Major Single-Plant Datasets}
\vspace*{-3mm}
\label{tab:single_plant_frequency} \\ \hline
\textbf{Dataset} & \textbf{Class of diseases} & \textbf{No. of images} & \textbf{Total Images} & \textbf{Skewness measure} \\ \hline
\endfirsthead
\hline
\textbf{Dataset} & \textbf{Class of diseases} & \textbf{No. of images} & \textbf{Total Images} & \textbf{Skewness measure} \\ \hline
\endhead
\hline
\endfoot
\hline
\endlastfoot

\multirow{5}{=}{Apple dataset ALDD \cite{jiang2019real}}
& Alternaria Leaf Spot & 5343 & \multirow{5}{=}{26,377} & \multirow{5}{=}{0.155} \\
\cline{2-3}
& Apple-Rust & 5694 &  &  \\
\cline{2-3}
& Mosaic-Virus & 4875 &  &  \\
\cline{2-3}
& Gray-Spot & 4810 &  &  \\
\cline{2-3}
& Brown-Spot & 5655 &  &  \\ \hline

\multirow{7}{=}{Soybean Dataset \cite{wu2019identification}}
& Bacterial & 200 & \multirow{7}{=}{1,470} & \multirow{7}{=}{0.294} \\
\cline{2-3}
& Downy-Mildew & 255 &  &  \\
\cline{2-3}
& Spider-Mite & 235 &  &  \\
\cline{2-3}
& Pest & 200 &  &  \\
\cline{2-3}
& Virus & 200 &  &  \\
\cline{2-3}
& Pesticide & 180 &  &  \\
\cline{2-3}
& Healthy & 200 &  &  \\ \hline

\multirow{3}{=}{Citrus Dataset \cite{dananjayan2022assessment}}
& Anthracnose & 993 & \multirow{3}{=}{2,684} & \multirow{3}{=}{0.243} \\
\cline{2-3}
& Bacterial Brown Spot & 752 &  &  \\
\cline{2-3}
& Melanose & 939 &  &  \\ \hline

\multirow{3}{=}{Rice Dataset \cite{joshi2022ricebios}}
& Fungal & 103 & \multirow{3}{=}{442} & \multirow{3}{=}{0.874} \\
\cline{2-3}
& Bacterial & 301 &  &  \\
\cline{2-3}
& Healthy & 38 &  &  \\ \hline

\multirow{10}{=}{Apple plant dataset \cite{khan2022deep}}
& Apple Scab & 639 & \multirow{10}{=}{5,201} & \multirow{10}{=}{0.954} \\
\cline{2-3}
& Powdery Mildew & 320 &  &  \\
\cline{2-3}
& Alternaria & 418 &  &  \\
\cline{2-3}
& Necrosis & 125 &  &  \\
\cline{2-3}
& MLB & 371 &  &  \\
\cline{2-3}
& Mosaic & 312 &  &  \\
\cline{2-3}
& Insect & 221 &  &  \\
\cline{2-3}
& Multiple Symptoms & 2462 &  &  \\
\cline{2-3}
& Damaged & 112 &  &  \\
\cline{2-3}
& Healthy & 221 &  &  \\ \hline

\multirow{2}{=}{Tomato Dataset \cite{chug2023novel}}
& Healthy & 79 & \multirow{2}{=}{155} & \multirow{2}{=}{0.038} \\
\cline{2-3}
& Early Blight & 76 &  &  \\ \hline

\multirow{13}{=}{Tomato Dataset \cite{khatoon2021image}}
& Mosaic Virus & 830 & \multirow{13}{=}{15,335} & \multirow{13}{=}{0.896} \\
\cline{2-3}
& Powdery Mildew & 241 &  &  \\
\cline{2-3}
& Bacterial Spots & 2321 &  &  \\
\cline{2-3}
& Late Blight & 1909 &  &  \\
\cline{2-3}
& Early Blight & 1000 &  &  \\
\cline{2-3}
& Leaf Miner & 1230 &  &  \\
\cline{2-3}
& Septoria & 1771 &  &  \\
\cline{2-3}
& Taro Caterpillar & 672 &  &  \\
\cline{2-3}
& Spider Mite & 1754 &  &  \\
\cline{2-3}
& Whitefly & 420 &  &  \\
\cline{2-3}
& Melon Fly & 294 &  &  \\
\cline{2-3}
& Mold & 952 &  &  \\
\cline{2-3}
& Tomato Healthy & 1941 &  &  \\ \hline

\multirow{5}{=}{Tomato Dataset \cite{ashwinkumar2022automated}}
& Early Blight & 1000 & \multirow{5}{=}{5,512} & \multirow{5}{=}{0.968} \\
\cline{2-3}
& Late Blight & 1909 &  &  \\
\cline{2-3}
& Target Spot & 60 &  &  \\
\cline{2-3}
& Mold & 952 &  &  \\
\cline{2-3}
& Healthy & 1591 &  &  \\ \hline

\multirow{3}{=}{Rice Dataset \cite{prajapati2017detection}}
& Bacterial Blight & 40 & \multirow{3}{=}{120} & \multirow{3}{=}{0.00} \\
\cline{2-3}
& Leaf Smut & 40 &  &  \\
\cline{2-3}
& Brown Spot & 40 &  &  \\ \hline

\multirow{8}{=}{Soybean Dataset \cite{bevers2022pictures}}
& Bacterial Blight & 484 & \multirow{8}{=}{9,648} & \multirow{8}{=}{0.703} \\
\cline{2-3}
& Rust & 1627 &  &  \\
\cline{2-3}
& Frogeye Leaf Spot & 1540 &  &  \\
\cline{2-3}
& Downy Mildew & 652 &  &  \\
\cline{2-3}
& Cercospora Leaf Blight & 1598 &  &  \\
\cline{2-3}
& Potassium Deficiency & 1034 &  &  \\
\cline{2-3}
& Target Spot & 1081 &  &  \\
\cline{2-3}
& Healthy & 1632 &  &  \\ \hline

\multirow{4}{=}{Citrus Dataset \cite{syed2022citrus}}
& Blackspot & 172 & \multirow{4}{=}{597} & \multirow{4}{=}{0.716} \\
\cline{2-3}
& Canker & 163 &  &  \\
\cline{2-3}
& Huanglongbing & 204 &  &  \\
\cline{2-3}
& Healthy & 58 &  &  \\ \hline

\multirow{4}{=}{Coffee Dataset \cite{tassis2021deep}}
& Rust & 119 & \multirow{4}{=}{390} & \multirow{4}{=}{0.727} \\
\cline{2-3}
& Cercospora leaf spot & 33 &  &  \\
\cline{2-3}
& Brown leaf spot & 121 &  &  \\
\cline{2-3}
& Healthy & 117 &  &  \\ \hline

\multirow{3}{=}{Beans Dataset \cite{elfatimi2022beans}}
& Angular leaf spot & 432 & \multirow{3}{=}{1,296} & \multirow{3}{=}{0.018} \\
\cline{2-3}
& Bean Rust & 436 &  &  \\
\cline{2-3}
& Healthy & 428 &  &  \\ \hline

\multirow{4}{=}{Sugar Dataset \cite{adem2023sugar}}
& Mild Disease & 20 & \multirow{4}{=}{155} & \multirow{4}{=}{0.677} \\
\cline{2-3}
& Severe Disease & 35 &  &  \\
\cline{2-3}
& Mild and severe Disease & 62 &  &  \\
\cline{2-3}
& Healthy & 38 &  &  \\ \hline

\multirow{5}{=}{Citrus Dataset \cite{liu2022ds}}
& Healthy & 564 & \multirow{5}{=}{3,086} & \multirow{5}{=}{0.256} \\
\cline{2-3}
& Canker & 634 &  &  \\
\cline{2-3}
& Scab & 592 &  &  \\
\cline{2-3}
& Black Spot & 743 &  &  \\
\cline{2-3}
& Anthracnose & 553 &  &  \\ \hline

\end{longtable}
}

\subsection{Structural Characteristics of Multi-Plant Datasets}
Multi-plant datasets reflect a more realistic scenario of plant disease detection and involve more complexity than single-plant datasets. Thus, besides including more crop species, the number of classes of diseases in a single dataset is considerably large. Due to such intrinsic heterogeneity, they are considered more important for assessing the robustness and capability of models to generalize under diverse agricultural conditions.

The analyzed multi-plant datasets exhibit substantial variation in crop coverage, ranging from limited two-species collections to large-scale datasets encompassing more than ten crop categories. While increased crop diversity enhances ecological realism, it simultaneously introduces structural heterogeneity in feature distributions. In several datasets, certain crops contribute disproportionately to the total sample count, potentially biasing learned representations toward dominant species.
\par
Across datasets, the number of disease classes varies significantly, with some collections focusing on a restricted set of common diseases while others attempt broad coverage. However, increased class count does not necessarily imply structural balance. The frequency distribution tables reveal strong disparities between majority and minority disease classes, particularly in datasets collected under field conditions.
\par
Unlike single-plant datasets, imbalance in multi-plant datasets occurs at two levels: intra-crop (within a species) and inter-crop (across species). This layered skewness complicates model training, as networks may inadvertently prioritize dominant crop-disease combinations while underrepresenting rare interactions.
\par
Field-collected multi-plant datasets often demonstrate greater variability in class distribution compared to lab-controlled datasets. The presence of natural environmental noise, variable disease incidence rates, and uneven crop representation contributes to increased structural skewness, as reflected in their normalized imbalance ratios.
These structural characteristics suggest that performance metrics reported on multi-plant datasets should be interpreted cautiously. High overall accuracy may mask poor minority-class performance, particularly in datasets exhibiting strong skewness. Consequently, evaluating cross-crop generalization and minority disease detection remains critical for assessing real-world deployment readiness.

\subsubsection{Inter-crop structural imbalance (crop-level distribution differences)}
Multi-plant datasets introduce an additional layer of structural complexity that is absent in single-plant collections: inter-crop imbalance. In these datasets, the distribution of samples is not only uneven across disease classes within a crop but also across different crop species represented in the dataset. This uneven crop representation can significantly influence learning dynamics, particularly when certain crops contribute disproportionately to the total sample pool.

To systematically examine these structural properties, we analyse representative multi-plant datasets by detailing their plant species, disease classes per species, per-class image counts, total image volume, and normalised imbalance ratio (NIR) across plant species. The following tables~\ref{tab:multiplant_distribution}--\ref{tab:rice_betel_crop_distribution} present the frequency distribution of disease categories within each dataset. This structured representation enables direct observation of dominant classes, minority categories, and overall distribution skewness across plant species in the multiplant dataset.

\begin{table}[H]
\centering
\footnotesize
\caption{Crop distribution statistics for the Multi-Plant Dataset \cite{chouhan2019data}.}
\vspace*{-3mm}
\label{tab:multiplant_distribution}
\begin{tabular}{p{3.0cm}p{1.2cm}p{1.2cm}p{1.2cm}p{1.5cm}p{2.0cm}p{1.5cm}}
\hline
\textbf{Plant Name} & \textbf{Healthy} & \textbf{Diseased} &
\textbf{Total Images (Crop)} & \textbf{\% of Dataset} &
\textbf{Dominant Class} & \textbf{Skewness Measure} \\ 
\hline

Arjun & 220 & 232 & 452 & 11.08\% & Diseased & 0.052 \\ \hline
Mango & 170 & 265 & 435 & 10.66\% & Diseased & 0.358 \\ \hline
Bael & 0 & 118 & 118 & 2.89\% & Diseased & 1.00 \\ \hline
Guava & 277 & 142 & 419 & 10.27\% & Healthy & 0.487 \\ \hline
Chinar & 103 & 120 & 223 & 5.46\% & Diseased & 0.142 \\ \hline
Lemon & 159 & 77 & 236 & 5.78\% & Healthy & 0.516 \\ \hline
Jamun & 279 & 345 & 624 & 15.29\% & Diseased & 0.191 \\ \hline
Pongamia Pinnata & 332 & 276 & 608 & 14.90\% & Healthy & 0.169 \\ \hline
Jatropha & 133 & 124 & 257 & 6.30\% & Healthy & 0.068 \\ \hline
Basil & 148 & 0 & 148 & 3.2\% & Healthy & 1.00 \\ \hline
Pomegranate & 287 & 272 & 559 & 13.7\% & Healthy & 0.052 \\ \hline

\end{tabular}
\end{table}

\par
Siddharth et al. \cite{chouhan2019data} created a Multi Plant Database with 4,079 images of 11 plant species. However, the dataset has limitations: it categorizes leaves only as healthy or diseased without specifying types of plant diseases, which may hinder detailed diagnosis. Additionally, the images were captured in a controlled environment, and there is a significant imbalance in the representation of certain crops and diseases within the dataset. The structural distribution of disease classes across different crops in the Multiplant dataset is summarized in Table~\ref{tab:multiplant_distribution}.

\begin{table}[H]
\centering
\footnotesize
\caption{Crop distribution statistics for the PlantVillage dataset \cite{hughes2015open}.}
\vspace*{-3mm}
\label{tab:plantvillage_distribution}
\resizebox{\textwidth}{!}{
\begin{tabular}{p{1.5cm}p{3.8cm}p{1.5cm}p{1.5cm}p{1.4cm}p{2.9cm}p{1.6cm}}
\hline
\textbf{Plant Name} & \textbf{Class of Diseases} & \textbf{No. of Images} & \textbf{Total Images (Crop)} & \textbf{\% of Dataset} & \textbf{Dominant Disease Class} & \textbf{Skewness Measure} \\
\hline

\multirow{4}{*}{Apple}
& Apple Cider Rust & 276 
& \multirow{4}{*}{3,172}
& \multirow{4}{*}{5.84\%}
& \multirow{4}{*}{Healthy}
& \multirow{4}{*}{0.833} \\ \cline{2-3}
& Apple Black Rot & 621 & & & & \\ \cline{2-3}
& Apple Scab & 630 & & & & \\ \cline{2-3}
& Apple Healthy Leaf & 1645 & & & & \\ \hline

\multirow{2}{*}{Peach}
& Peach Bacterial Spot & 2297
& \multirow{2}{*}{2,657}
& \multirow{2}{*}{4.89\%}
& \multirow{2}{*}{Bacterial Spot}
& \multirow{2}{*}{0.843} \\ \cline{2-3}
& Peach Healthy Leaf & 360 & & & & \\ \hline

\multirow{2}{*}{Blueberry}
& Blueberry Healthy Leaf & 1502
& \multirow{2}{*}{1,502}
& \multirow{2}{*}{2.77\%}
& \multirow{2}{*}{Healthy}
& \multirow{2}{*}{1.00} \\ \cline{2-3}
& Blueberry Diseased Leaf & 0 & & & & \\ \hline

\multirow{2}{*}{Cherry}
& Cherry Healthy Leaf & 854
& \multirow{2}{*}{1,906}
& \multirow{2}{*}{3.51\%}
& \multirow{2}{*}{Powdery Mildew}
& \multirow{2}{*}{0.188} \\ \cline{2-3}
& Cherry Powdery Mildew & 1052 & & & & \\ \hline

\multirow{4}{*}{Grape}
& Grape Black Measles & 1384
& \multirow{4}{*}{4,063}
& \multirow{4}{*}{7.48\%}
& \multirow{4}{*}{Black Measles}
& \multirow{4}{*}{0.694} \\ \cline{2-3}
& Grape Leaf Blight (Isariopsis Leaf Spot) & 1076 & & & & \\ \cline{2-3}
& Grape Black Rot & 1180 & & & & \\ \cline{2-3}
& Grape Healthy Leaf & 423 & & & & \\ \hline

\multirow{2}{*}{Raspberry}
& Raspberry Healthy Leaf & 371
& \multirow{2}{*}{371}
& \multirow{2}{*}{0.68\%}
& \multirow{2}{*}{Healthy}
& \multirow{2}{*}{1.00} \\ \cline{2-3}
& Raspberry Diseased Leaf & 0 & & & & \\ \hline

\multirow{2}{*}{Orange}
& Orange Huanglongbing & 5507
& \multirow{2}{*}{5,507}
& \multirow{2}{*}{10.14\%}
& \multirow{2}{*}{Huanglongbing}
& \multirow{2}{*}{1.00} \\ \cline{2-3}
& Orange Healthy Leaf & 0 & & & & \\ \hline

\multirow{4}{*}{Corn}
& Corn Common Rust & 1192
& \multirow{4}{*}{3,852}
& \multirow{4}{*}{7.09\%}
& \multirow{4}{*}{Common Rust}
& \multirow{4}{*}{0.570} \\ \cline{2-3}
& Corn Gray Leaf Spot & 513 & & & & \\ \cline{2-3}
& Corn Northern Leaf Blight & 985 & & & & \\ \cline{2-3}
& Corn Healthy & 1162 & & & & \\ \hline

\multirow{2}{*}{Pepper}
& Pepper Bacterial Spot & 997
& \multirow{2}{*}{2,475}
& \multirow{2}{*}{4.55\%}
& \multirow{2}{*}{Healthy}
& \multirow{2}{*}{0.325} \\ \cline{2-3}
& Pepper Healthy Leaf & 1478 & & & & \\ \hline

\multirow{3}{*}{Potato}
& Potato Early Blight & 1000
& \multirow{3}{*}{2,152}
& \multirow{3}{*}{3.96\%}
& \multirow{3}{*}{Early Blight}
& \multirow{3}{*}{0.848} \\ \cline{2-3}
& Potato Late Blight & 1000 & & & & \\ \cline{2-3}
& Potato Healthy Leaf & 152 & & & & \\ \hline

\multirow{2}{*}{Strawberry}
& Strawberry Healthy Leaf & 456
& \multirow{2}{*}{1,565}
& \multirow{2}{*}{2.88\%}
& \multirow{2}{*}{Leaf Scorch}
& \multirow{2}{*}{0.581} \\ \cline{2-3}
& Strawberry Leaf Scorch & 1109 & & & & \\ \hline

\multirow{2}{*}{Soybean}
& Soybean Healthy Leaf & 5090
& \multirow{2}{*}{5,090}
& \multirow{2}{*}{9.37\%}
& \multirow{2}{*}{Healthy}
& \multirow{2}{*}{1.00} \\ \cline{2-3}
& Soybean Diseased Leaf & 0 & & & & \\ \hline

\multirow{2}{*}{Squash}
& Squash Powdery Mildew & 1835
& \multirow{2}{*}{1,835}
& \multirow{2}{*}{3.38\%}
& \multirow{2}{*}{Powdery Mildew}
& \multirow{2}{*}{1.00} \\ \cline{2-3}
& Squash Healthy Leaf & 0 & & & & \\ \hline

\multirow{10}{*}{Tomato}
& Tomato Early Blight & 1000
& \multirow{10}{*}{18,162}
& \multirow{10}{*}{33.44\%}
& \multirow{10}{*}{Yellow Leaf Curl Virus}
& \multirow{10}{*}{0.930} \\ \cline{2-3}
& Tomato Yellow Leaf Curl Virus & 5357 & & & & \\ \cline{2-3}
& Tomato Septoria Leaf Spot & 1771 & & & & \\ \cline{2-3}
& Tomato Bacterial Leaf Spot & 2127 & & & & \\ \cline{2-3}
& Tomato Late Blight & 1910 & & & & \\ \cline{2-3}
& Tomato Leaf Mold & 952 & & & & \\ \cline{2-3}
& Tomato Target Spot & 1404 & & & & \\ \cline{2-3}
& Tomato Two Spotted Spider Mite & 1676 & & & & \\ \cline{2-3}
& Tomato Mosaic Virus & 373 & & & & \\ \cline{2-3}
& Tomato Healthy Leaf & 1592 & & & & \\ \hline

\end{tabular}
}
\end{table}

\par
The PlantVillage \cite{hughes2015open}, curated by researchers Hughes and Salathé, encompasses a comprehensive collection of 54,309 images representing 14 distinct crop species. The images were captured under controlled laboratory conditions with a digital camera.The dataset is offered in three versions: original RGB images, grayscale versions of the raw images, and segmented RGB images that feature only the leaf, color-corrected and isolated. The structural distribution of disease classes across different crops in the PlantVillage dataset is summarized in Table~\ref{tab:plantvillage_distribution}. 
\par
The PlantDoc dataset, created by Singh et al. \cite{singh2020plantdoc}, was compiled from diverse internet sources. consists of 2,569 images covering 13 plant species. Most of the images were sourced from the internet, which resulted in generally poor quality. Additionally, some images were taken in a laboratory setting, and many images also contain watermarks. The dataset includes not only images of leaves but also fruits and other parts of plants. The structural distribution of dataset is summarized in Table~\ref{tab:plantdoc_distribution}. Because different versions of the dataset have been cleaned and processed (for detection vs. classification), the exact image count varies across repositories, but class distribution remains the same.

\begin{table}[H]
\centering
\footnotesize
\caption{Crop-wise structural distribution of disease categories in the PlantDoc dataset \cite{singh2020plantdoc}. }
\vspace*{-3mm}
\label{tab:plantdoc_distribution}
\begin{tabular}{p{1.5cm}p{3.8cm}p{1.2cm}p{1.5cm}p{1.4cm}p{2.9cm}p{1.6cm}}
\hline
\textbf{Plant Name} & \textbf{Class of diseases} & \textbf{No of images} & \textbf{Total Images (Crop)} & \textbf{\% of Dataset} & \textbf{Dominant Disease Class} & \textbf{Skewness measure} 
\\ \hline

\multirow{3}{*}{Apple}
& Apple Scab Leaf & 92 & \multirow{3}{*}{272} & \multirow{3}{*}{10.59\%} & \multirow{3}{*}{Apple Scab} & \multirow{3}{*}{0.33} \\ \cline{2-3}
& Apple Rust Leaf & 89 & & & & \\ \cline{2-3}
& Apple Healthy Leaf & 91 & & & & \\ \hline

\multirow{2}{*}{Blueberry}
& Blueberry Healthy Leaf & 117 & \multirow{2}{*}{117} & \multirow{2}{*}{4.55\%} & \multirow{2}{*}{Healthy Leaf} & \multirow{2}{*}{1.00} \\ \cline{2-3}
& Blueberry Diseased Leaf & 0 & & & & \\ \hline

\multirow{2}{*}{Cherry}
& Cherry Healthy Leaf & 57 & \multirow{2}{*}{57} & \multirow{2}{*}{2.21\%} & \multirow{2}{*}{Healthy Leaf} & \multirow{2}{*}{1.00} \\ \cline{2-3}
& Cherry Diseased Leaf & 0 & & & & \\ \hline

\multirow{2}{*}{Grape}
& Grape Leaf Black Rot & 64 & \multirow{2}{*}{133} & \multirow{2}{*}{5.17\%} & \multirow{2}{*}{Healthy Leaf} & \multirow{2}{*}{0.072} \\ \cline{2-3}
& Grape Healthy Leaf & 69 & & & & \\ \hline

\multirow{2}{*}{Bell Pepper}
& Pepper Leaf Spot & 62 & \multirow{2}{*}{123} & \multirow{2}{*}{4.78\%} & \multirow{2}{*}{Leaf Spot} & \multirow{2}{*}{0.016} \\ \cline{2-3}
& Pepper Healthy Leaf & 61 & & & & \\ \hline

\multirow{2}{*}{Soybean}
& Soybean Healthy Leaf & 65 & \multirow{2}{*}{65} & \multirow{2}{*}{2.53\%} & \multirow{2}{*}{Healthy Leaf} & \multirow{2}{*}{1.00} \\ \cline{2-3}
& Soybean Diseased Leaf & 0 & & & & \\ \hline

\multirow{3}{*}{Corn}
& Corn Leaf Blight & 192 & \multirow{3}{*}{376} & \multirow{3}{*}{14.63\%} & \multirow{3}{*}{Leaf Blight} & \multirow{3}{*}{0.646} \\ \cline{2-3}
& Corn Gray Leaf Spot & 68 & & & & \\ \cline{2-3}
& Corn Rust Leaf & 116 & & & & \\ \hline

\multirow{2}{*}{Peach}
& Peach Healthy Leaf & 112 & \multirow{2}{*}{112} & \multirow{2}{*}{4.35\%} & \multirow{2}{*}{Healthy Leaf} & \multirow{2}{*}{1.00} \\ \cline{2-3}
& Peach Diseased Leaf & 0 & & & & \\ \hline

\multirow{2}{*}{Potato}
& Potato Early Blight & 117 & \multirow{2}{*}{222} & \multirow{2}{*}{8.70\%} & \multirow{2}{*}{Early Blight} & \multirow{2}{*}{0.102} \\ \cline{2-3}
& Potato Leaf Late Blight & 105 & & & & \\ \hline

\multirow{2}{*}{Raspberry}
& Raspberry Healthy Leaf & 119 & \multirow{2}{*}{119} & \multirow{2}{*}{4.63\%} & \multirow{2}{*}{Healthy Leaf} & \multirow{2}{*}{1.00} \\ \cline{2-3}
& Raspberry Diseased Leaf & 0 & & & & \\ \hline

\multirow{2}{*}{Strawberry}
& Strawberry Healthy Leaf & 96 & \multirow{2}{*}{96} & \multirow{2}{*}{3.73\%} & \multirow{2}{*}{Healthy Leaf} & \multirow{2}{*}{1.00} \\ \cline{2-3}
& Strawberry Diseased Leaf & 0 & & & & \\ \hline

\multirow{2}{*}{Squash}
& Squash Powdery Mildew & 130 & \multirow{2}{*}{130} & \multirow{2}{*}{5.10\%} & \multirow{2}{*}{Powdery Mildew} & \multirow{2}{*}{1.00} \\ \cline{2-3}
& Squash Healthy Leaf & 0 & & & & \\ \hline

\multirow{8}{*}{Tomato}
& Tomato Early Blight & 88 & \multirow{8}{*}{744} & \multirow{8}{*}{28.96\%} & \multirow{8}{*}{Bacterial Spot} & \multirow{8}{*}{0.642} \\ \cline{2-3}
& Tomato Late Blight & 111 & & & & \\ \cline{2-3}
& Tomato Bacterial Spot & 151 & & & & \\ \cline{2-3}
& Tomato Septoria Spot & 110 & & & & \\ \cline{2-3}
& Tomato Yellow Virus Leaf & 76 & & & & \\ \cline{2-3}
& Tomato Mosaic Virus & 54 & & & & \\ \cline{2-3}
& Tomato Mold Leaf & 91 & & & & \\ \cline{2-3}
& Tomato Healthy & 63 & & & & \\ \hline

\end{tabular}
\end{table}

\begin{table}[H]
\centering
\footnotesize
\caption{Crop-wise structural distribution of disease categories in the Turkey Plant Dataset \cite{tugrul2022convolutional}.}
\vspace*{-3mm}
\label{tab:turkey_distribution}
\begin{tabular}{p{1.0cm}p{4.0cm}p{1.5cm}p{1.5cm}p{1.4cm}p{3.0cm}p{1.6cm}}
\hline
\textbf{Plant Name} & \textbf{Class of Disease} & \textbf{No. of Images} & 
\textbf{Total Images (Crop)} & \textbf{\% of Dataset} & 
\textbf{Dominant Disease Class} & \textbf{Skewness Measure} \\ 
\hline

\multirow{4}{*}{Apple} 
    & Apple Aphis spp & 162 
    & \multirow{4}{*}{1416} 
    & \multirow{4}{*}{31.84\%} 
    & \multirow{4}{*}{Venturia inaequalis} 
    & \multirow{4}{*}{0.744} \\ \cline{2-3}
    & Apple Venturia inaequalis & 633 & & & & \\ \cline{2-3}
    & Apple Monilia lava & 255 & & & & \\ \cline{2-3}
    & Apple Eriosoma lanigerum & 366 & & & & \\ 
\hline

\multirow{2}{*}{Peach} 
    & Peach Monilia lava & 314 
    & \multirow{2}{*}{746} 
    & \multirow{2}{*}{16.77\%} 
    & \multirow{2}{*}{Parthenolecanium corni} 
    & \multirow{2}{*}{0.264} \\ \cline{2-3}
    & Peach Parthenolecanium corni & 427 & & & & \\ 
\hline

\multirow{2}{*}{Pear} 
    & Pear Erwinia amylovora & 215 
    & \multirow{2}{*}{215} 
    & \multirow{2}{*}{4.83\%} 
    & \multirow{2}{*}{Pear Erwinia amylovora} 
    & \multirow{2}{*}{1.00} \\ \cline{2-3}
    & Pear Healthy & 0 & & & & \\
\hline

\multirow{4}{*}{Apricot} 
    & Apricot Monilia lava & 85 
    & \multirow{4}{*}{1400} 
    & \multirow{4}{*}{31.48\%} 
    & \multirow{4}{*}{Coryneum beijerinckii} 
    & \multirow{4}{*}{0.93} \\ \cline{2-3}
    & Apricot Coryneum beijerinckii & 1100 & & & & \\ \cline{2-3}
    & Apricot Drying Symptom & 139 & & & & \\ \cline{2-3}
    & Apricot Cancer Symptom & 76 & & & & \\ 
\hline

\multirow{2}{*}{Walnut} 
    & Walnut Eriophyes erineus & 69 
    & \multirow{2}{*}{249} 
    & \multirow{2}{*}{5.59\%} 
    & \multirow{2}{*}{Gnomonia leptostyla} 
    & \multirow{2}{*}{0.617} \\ \cline{2-3}
    & Walnut Gnomonia leptostyla & 180 & & & & \\ 
\hline

\multirow{2}{*}{Cherry} 
    & Cherry Aphis spp & 356 
    & \multirow{2}{*}{356} 
    & \multirow{2}{*}{8.00\%} 
    & \multirow{2}{*}{Aphis spp} 
    & \multirow{2}{*}{1.00} \\ \cline{2-3}
    & Cherry Healthy & 0 & & & & \\ 
\hline

\multirow{2}{*}{Plum} 
    & Plum Aphis spp & 70 
    & \multirow{2}{*}{70} 
    & \multirow{2}{*}{1.57\%} 
    & \multirow{2}{*}{Aphis spp} 
    & \multirow{2}{*}{1.00} \\ \cline{2-3}
    & Plum Healthy & 0 & & & & \\ 
\hline

\end{tabular}
\end{table}

\par
The Turkey Plant Dataset, created by Turkoglu et al. \cite{tugrul2022convolutional} is a comprehensive collection of 4,447 images of plants showing different diseases and pests found in Turkey. The images are either with or without background scenery and are organized into 15 classes. The images are collected from field studies done at the Faculty of Agriculture at Inonu and Bingol Universities. The dataset includes natural images encompassing various scenarios such as soil, trees, leaves, and sky. However, there is a significant imbalance between certain crops and diseases in the dataset, as some crops lack images of healthy plants. The structural distribution of disease classes across different crops in the Turkey Plant dataset is summarized in Table~\ref{tab:turkey_distribution}.

\begin{table}[H]
\centering
\footnotesize
\caption{Crop distribution statistics for the FieldPlant Dataset \cite{moupojou2023fieldplant}.}
\vspace*{-3mm}
\label{tab:fieldplant_distribution}
\begin{tabular}{p{1.5cm}p{3.0cm}p{1.5cm}p{1.5cm}p{1.4cm}p{2.8cm}p{1.6cm}}
\hline
\textbf{Plant Name} & \textbf{Class of Disease} & \textbf{No. of Images} &
\textbf{Total Images (Crop)} & \textbf{\% of Dataset} &
\textbf{Dominant Disease Class} & \textbf{Skewness Measure} \\
\hline

\multirow{16}{*}{Corn}
 & Leaf Blight & 1655
 & \multirow{16}{*}{3,194}
 & \multirow{16}{*}{37.0\%}
 & \multirow{16}{*}{Leaf Blight}
 & \multirow{16}{*}{0.996} \\ \cline{2-3}
 & Brown Spots & 220 & & & & \\ \cline{2-3}
 & Corn Stripe & 164 & & & & \\ \cline{2-3}
 & Corn Yellowing & 493 & & & & \\ \cline{2-3}
 & Corn Streak & 251 & & & & \\ \cline{2-3}
 & Chlorotic Leaf Spot & 26 & & & & \\ \cline{2-3}
 & Grey Leaf Spot & 54 & & & & \\ \cline{2-3}
 & Insect Damage & 17 & & & & \\ \cline{2-3}
 & Corn Healthy & 175 & & & & \\ \cline{2-3}
 & Corn Mildew & 39 & & & & \\ \cline{2-3}
 & Purple Discoloration & 8 & & & & \\ \cline{2-3}
 & Rust Leaf & 49 & & & & \\ \cline{2-3}
 & Corn Smut & 7 & & & & \\ \cline{2-3}
 & Corn Charcoal & 7 & & & & \\ \cline{2-3}
 & Yellow Spots & 22 & & & & \\ \cline{2-3}
 & Violet Discoloration & 7 & & & & \\ \hline

\multirow{5}{*}{Cassava}
 & Cassava Mosaic & 1898
 & \multirow{5}{*}{2638}
 & \multirow{5}{*}{30.57\%}
 & \multirow{5}{*}{Cassava Mosaic}
 & \multirow{5}{*}{0.985} \\ \cline{2-3}
 & Bacterial Blight & 29 & & & & \\ \cline{2-3}
 & Brown Leaf Spot & 213 & & & & \\ \cline{2-3}
 & Cassava Healthy & 421 & & & & \\ \cline{2-3}
 & Cassava Root Rot & 77 & & & & \\ \hline

\multirow{6}{*}{Tomato}
 & Brown Spots & 1978
 & \multirow{6}{*}{2797}
 & \multirow{6}{*}{32.41\%}
 & \multirow{6}{*}{Brown Spots}
 & \multirow{6}{*}{0.999} \\ \cline{2-3}
 & Tomato Healthy & 279 & & & & \\ \cline{2-3}
 & Blight Leaf & 404 & & & & \\ \cline{2-3}
 & Yellow Virus & 115 & & & & \\ \cline{2-3}
 & Leaf Mosaic Virus & 19 & & & & \\ \cline{2-3}
 & Bacterial Wilt & 2 & & & & \\ \hline

\end{tabular}
\end{table}

\par
FieldPlant disease dataset created by Moupojou et al. \cite{moupojou2023fieldplant}, comprises a substantial collection of  5,170 images depicting various plant diseases observed in three significant crops: tomato, corn, and cassava sourced from the vibrant plantations of Cameroon prevalent diseases . However, the datsaset is annotated with 8,629  leaves categorized into 27 distinct disease classes. While the dataset offers invaluable insights for research and practical applications in agronomy and plant pathology, it does reveal a notable class imbalance. Certain crops and their associated diseases are less represented, with fewer images available for analysis. Additionally, the range of crop diseases included in the dataset is somewhat restricted. The structural distribution of disease classes across different crops in the FieldPlant dataset is summarized in Table~\ref{tab:fieldplant_distribution}.
\par 
The Fruit Plant Dataset, commonly known as the NZDL-Plant Disease dataset, was curated by Saleem et al. \cite{saleem2022performance}. This comprehensive dataset includes a total of 3,545 images across five key horticultural crops from New Zealand that represent 20 distinct classes of crop diseases. The images were acquired by using a smartphone across several local horticultural fields. Despite its extensive collection of images, the dataset has limitations in terms of the number of samples available for different crops, resulting in an imbalance that may significantly influence the effectiveness of models developed using this data. The structural distribution of disease classes across different crops in the NZDL Fruit Plant dataset is summarized in Table~\ref{tab:NDZL_distribution}
\par
The AI Challenger Dataset \cite{liang2019pd2se} spans across 10 plant species. While this dataset provides a substantial quantity of 26,747 images, its diversity and comprehensiveness may be limited due to the controlled conditions under which the data was collected. As with many image-based datasets, it may not fully represent the complexity and variability of real-world agricultural environments. Additionally, the dataset exhibits an imbalance between certain crops and diseases, with some crops having fewer images across different disease classes. The structural distribution of dataset is summarized in Table~\ref{tab:aichallenger_distribution}.

\par
\begin{table}[H]
\centering
\footnotesize
\caption{Crop-wise structural distribution of disease categories in the NZDL Fruit Plant Dataset \cite{saleem2022performance}.}
\vspace*{-3mm}
\label{tab:NDZL_distribution}
\begin{tabular}{p{1.2cm}p{3.8cm}p{1.5cm}p{1.5cm}p{1.4cm}p{3.0cm}p{1.6cm}}
\hline
\textbf{Plant Name} & \textbf{Class of Disease} & \textbf{No. of Images} & \textbf{Total Images (Crop)} & \textbf{\% of Dataset} & \textbf{Dominant Disease Class} & \textbf{Skewness Measure} \\
\hline

\multirow{7}{*}{Apple}
& Apple Leaf Black Spot & 81 
& \multirow{7}{*}{1202}
& \multirow{7}{*}{33.90\%}
& \multirow{7}{*}{Apple Leaf – Spot}
& \multirow{7}{*}{0.760} \\ \cline{2-3}
& Apple Leaf – Mosaic Virus & 224 \\ \cline{2-3}
& Apple Leaf – Spot & 271 \\ \cline{2-3}
& Apple Stem – Canker & 169 \\ \cline{2-3}
& Apple Fruit – Black Rot & 65 \\ \cline{2-3}
& Apple Leaf – Healthy & 215 \\ \cline{2-3}
& Apple Fruit – Healthy & 177 \\ \hline

\multirow{2}{*}{Grape}
& Grape Cane – Black Spot & 80
& \multirow{2}{*}{250}
& \multirow{2}{*}{7.05\%}
& \multirow{2}{*}{Grape Cane – Healthy}
& \multirow{2}{*}{0.529} \\ \cline{2-3}
& Grape Cane – Healthy & 170 \\ \hline

\multirow{3}{*}{Avocado}
& Avocado Leaf Spot & 104
& \multirow{3}{*}{580}
& \multirow{3}{*}{16.36\%}
& \multirow{3}{*}{Avocado Leaf – Canker}
& \multirow{3}{*}{0.584} \\ \cline{2-3}
& Avocado Leaf – Canker & 250 \\ \cline{2-3}
& Avocado Leaf – Healthy & 226 \\ \hline

\multirow{2}{*}{Kiwi}
& Kiwi Leaf – Canker & 318
& \multirow{2}{*}{589}
& \multirow{2}{*}{16.62\%}
& \multirow{2}{*}{Kiwi Leaf – Canker}
& \multirow{2}{*}{0.149} \\ \cline{2-3}
& Kiwi Leaf – Healthy & 271 \\ \hline

\multirow{6}{*}{Pear}
& Pear Leaf – Scab & 208
& \multirow{6}{*}{924}
& \multirow{6}{*}{26.06\%}
& \multirow{6}{*}{Pear Leaf – Healthy}
& \multirow{6}{*}{0.660} \\ \cline{2-3}
& Pear Leaf – Blight & 123 \\ \cline{2-3}
& Pear Stem – Canker & 96 \\ \cline{2-3}
& Pear Fruit – Stony Pit & 73 \\ \cline{2-3}
& Pear Leaf – Healthy & 215 \\ \cline{2-3}
& Pear Fruit – Healthy & 209 \\ \hline

\end{tabular}
\end{table}

{\footnotesize
\begin{longtable}{p{1.0cm}p{4.0cm}p{1.5cm}p{1.5cm}p{1.4cm}p{3.0cm}p{1.6cm}}
\caption{Crop distribution statistics for the AI Challenger Dataset \cite{liang2019pd2se}.}
\vspace*{-3mm}
\label{tab:aichallenger_distribution}\\
\hline
\textbf{Plant Name} & \textbf{Class of Diseases} & \textbf{No. of Images} & \textbf{Total Images (Crop)} & \textbf{\% of Dataset} & \textbf{Dominant Disease Class} & \textbf{Skewness Measure} \\
\hline

\multirow{3}{*}{Apple}
& Scab Apple G & 241
& \multirow{3}{*}{1,769}
& \multirow{3}{*}{6.61\%}
& \multirow{3}{*}{Healthy Apple}
& \multirow{3}{*}{0.871} \\ \cline{2-3}
& Scab Apple S & 174 \\ \cline{2-3}
& Healthy Apple & 1354 \\ \hline

\multirow{7}{*}{Grape}
& Leaf-Blight Fungus Grape G & 70
& \multirow{7}{*}{3,167}
& \multirow{7}{*}{11.80\%}
& \multirow{7}{*}{Leaf-Blight Fungus S}
& \multirow{7}{*}{0.903} \\ \cline{2-3}
& Black Rot Fungus Grape G & 435 \\ \cline{2-3}
& Black Measles Fungus Grape G & 587 \\ \cline{2-3}
& Leaf-Blight Fungus Grape S & 720 \\ \cline{2-3}
& Black Rot Fungus Grape S & 528 \\ \cline{2-3}
& Black Measles Fungus Grape S & 491 \\ \cline{2-3}
& Healthy Grape & 336 \\ \hline

\multirow{2}{*}{Cedar}
& Rust Apple Cedar G & 162
& \multirow{2}{*}{208}
& \multirow{2}{*}{0.77\%}
& \multirow{2}{*}{Rust Apple Cedar G}
& \multirow{2}{*}{0.716} \\ \cline{2-3}
& Rust Apple Cedar S & 46 \\ \hline

\multirow{3}{*}{Strawberry}
& Scorch Strawberry G & 219
& \multirow{3}{*}{1,162}
& \multirow{3}{*}{4.34\%}
& \multirow{3}{*}{Scorch Strawberry S}
& \multirow{3}{*}{0.671} \\ \cline{2-3}
& Scorch Strawberry S & 666 \\ \cline{2-3}
& Healthy Strawberry & 277 \\ \hline

\multirow{3}{*}{Cherry}
& Powdery Mildew Cherry G & 197
& \multirow{3}{*}{1066}
& \multirow{3}{*}{3.98\%}
& \multirow{3}{*}{Healthy Cherry}
& \multirow{3}{*}{0.728} \\ \cline{2-3}
& Powdery Mildew Cherry S & 186 \\ \cline{2-3}
& Healthy Cherry & 683 \\ \hline

\multirow{7}{*}{Corn}
& Cercospora Zeae MT & 216
& \multirow{7}{*}{2,603}
& \multirow{7}{*}{9.73\%}
& \multirow{7}{*}{Curvularia Leaf Spot S}
& \multirow{7}{*}{0.665} \\ \cline{2-3}
& Puccinia Polysora Corn G & 552 \\ \cline{2-3}
& Curvularia Leaf Spot Fungus Corn G & 238 \\ \cline{2-3}
& Cercospora Zeae MT Corn S & 191 \\ \cline{2-3}
& Puccinia Polysora Corn S & 406 \\ \cline{2-3}
& Curvularia Leaf Spot Fungus Corn S & 570 \\ \cline{2-3}
& Healthy Corn & 430 \\ \hline

\multirow{5}{*}{Potato}
& Early Blight Fungus Potato G & 232
& \multirow{5}{*}{3,246}
& \multirow{5}{*}{12.13\%}
& \multirow{5}{*}{Healthy Potato}
& \multirow{5}{*}{0.858} \\ \cline{2-3}
& Late Blight Fungus Potato G & 287 \\ \cline{2-3}
& Early Blight Fungus Potato S & 583 \\ \cline{2-3}
& Late Blight Fungus Potato S & 510 \\ \cline{2-3}
& Healthy Potato & 1634 \\ \hline

\multirow{3}{*}{Peach}
& Spot Peach G & 979
& \multirow{3}{*}{2,146}
& \multirow{3}{*}{8.02\%}
& \multirow{3}{*}{Spot Peach G}
& \multirow{3}{*}{0.707} \\ \cline{2-3}
& Bacterial Spot Peach S & 880 \\ \cline{2-3}
& Healthy Peach & 287 \\ \hline

\multirow{3}{*}{Pepper}
& Scab Pepper G & 327
& \multirow{3}{*}{1,930}
& \multirow{3}{*}{7.21\%}
& \multirow{3}{*}{Healthy Pepper}
& \multirow{3}{*}{0.721} \\ \cline{2-3}
& Pepper Scab S & 431 \\ \cline{2-3}
& Healthy Pepper & 1172 \\ \hline

\multirow{9}{*}{Tomato}
& Powdery Mildew Tomato G & 365
& \multirow{9}{*}{9,450}
& \multirow{9}{*}{35.33\%}
& \multirow{9}{*}{YLCV Virus Tomato S}
& \multirow{9}{*}{0.870} \\ \cline{2-3}
& Leaf Mold Fungus Tomato G & 371 \\ \cline{2-3}
& Septoria Leaf Spot Fungus Tomato G & 481 \\ \cline{2-3}
& YLCV Virus Tomato G & 1616 \\ \cline{2-3}
& Powdery Mildew Tomato S & 1104 \\ \cline{2-3}
& Leaf Mold Fungus Tomato S & 384 \\ \cline{2-3}
& Septoria Leaf Spot Fungus Tomato S & 922 \\ \cline{2-3}
& YLCV Virus Tomato S & 2826 \\ \cline{2-3}
& Healthy Tomato & 1381 \\ \hline
\begin{minipage}{17cm}

\small\itshape{In this table, "G" denotes General, while "S" indicates Serious.}
\end{minipage}   
\end{longtable}
}

\begin{table}[H]
\centering
\footnotesize
\caption{Crop-level distribution statistics for the UCI Rice and Wheat dataset \cite{jiang2021recognition}.}
\vspace*{-3mm}
\label{tab:uci_crop_distribution}
\begin{tabular}{p{1.1cm}p{3.0cm}p{1.5cm}p{1.5cm}p{1.4cm}p{2.8cm}p{1.6cm}}
\hline
\textbf{Plant} & \textbf{Class of disease} & \textbf{No. of images} & \textbf{Total Images (Crop)} & \textbf{\% of Dataset} & \textbf{Dominant Disease Class} & \textbf{Skewness Measure} \\
\hline

\multirow{3}{*}{Rice}
& Bacterial Leaf Blight & 40 & \multirow{3}{*}{120} & \multirow{3}{*}{60\%} & \multirow{3}{*}{Balanced} & \multirow{3}{*}{0.00} \\
\cline{2-3}
& Brown Spot & 40 & & & & \\
\cline{2-3}
& Leaf Blast & 40 & & & & \\
\hline

\multirow{2}{*}{Wheat}
& Leaf Rust & 40 & \multirow{2}{*}{80} & \multirow{2}{*}{40\%} & \multirow{2}{*}{Balanced} & \multirow{2}{*}{0.00} \\
\cline{2-3}
& Powdery Mildew & 40 & & & & \\
\hline

\end{tabular}
\end{table}

\par Jiang et al. \cite{jiang2021recognition} created a dataset comprising two crops (Rice and wheat) and five leaf diseases. The rice crop, consisting of 120 images, was obtained from the public UCI Machine Learning Repository, while wheat crop having 80 images were collected from public internet sources and verified with an agricultural disease database, resulting in 200 total images. To compensate for the limited sample size, offline data augmentation was applied, increasing the dataset to approximately 6000 images. However, the dataset relied on a very small number of original images, making the dataset heavily dependent on artificial augmentation. The wheat dataset was not drawn from a standardized benchmark source, potentially affecting reproducibility. The structural distribution of disease classes across different crops in the dataset is summarized in Table~\ref{tab:uci_crop_distribution}.

\begin{table}[h!]
\centering
\footnotesize
\caption{Crop-level distribution statistics for PlantDet dataset (Rice and Betel Leaf dataset) \cite{shovon2023plantdet}.}
\vspace*{-3mm}
\label{tab:rice_betel_crop_distribution}
\begin{tabular}{p{1.2cm}p{3.0cm}p{1.5cm}p{1.8cm}p{1.6cm}p{2.8cm}p{1.5cm}}
\hline
\textbf{Plant} & \textbf{Class of disease} & \textbf{No. of images} & \textbf{Total Images (Crop)} & \textbf{\% of Dataset} & \textbf{Dominant Disease Class} & \textbf{Skewness Measure} \\
\hline

\multirow{6}{*}{Rice}
& Bacterial Leaf Blight & 438 & \multirow{6}{*}{2710} & \multirow{6}{*}{73.05\%} & \multirow{6}{*}{Brown Spot} & \multirow{6}{*}{0.041} \\
\cline{2-3}
& Brown Spot & 466 & & & & \\
\cline{2-3}
& Healthy & 464 & & & & \\
\cline{2-3}
& Leaf Blast & 454 & & & & \\
\cline{2-3}
& Leaf Scald & 448 & & & & \\
\cline{2-3}
& Narrow Brown Spot & 440 & & & & \\
\hline

\multirow{2}{*}{Betel}
& Healthy & 500 & \multirow{2}{*}{1000} & \multirow{2}{*}{26.95\%} & \multirow{2}{*}{Balanced} & \multirow{2}{*}{0.00} \\
\cline{2-3}
& Unhealthy & 500 & & & & \\
\hline

\end{tabular}
\end{table}
\par
Shovon et al. \cite{shovon2023plantdet} created a PlantDet dataset for rice and betel leaf crops. The first is a public Rice Leaf dataset (from Kaggle) containing 2710 images across six classes. The second is a self-collected Betel Leaf dataset comprising 1000 field images, divided into two classes. However, the betel leaf dataset labels samples broadly as “unhealthy” rather than disease-specific categories, limiting fine-grained disease analysis. The dataset size is relatively small and covers only one crop with non-specific disease labeling, reducing diversity. Additionally, part of the data (Betel Leaf dataset) is private and not publicly accessible. The structural distribution of disease classes across different crops in the dataset is summarized in Table~\ref{tab:rice_betel_crop_distribution}.

\par
Brown et al \cite{brown30plant} introduced a multispectral plant disease dataset comprising 3016 field images collected using a camera with visible (RGB) and near-infrared (NIR) filters. The dataset includes five fruit crops with both healthy and diseased samples, captured under natural environmental conditions. However, the dataset is relatively small, exhibits class imbalance, and has limited publicly benchmarked validation, which may affect reproducibility and generalization.

\par
Plant Disease Dataset-271 (PDD271) \cite{liu2021plant} consists of 220,592 annotated images across 38 plant species and 271 disease categories collected in real-world scenarios with complex backgrounds. It covers various plant species, including fruit trees, vegetables, and field crops. The dataset includes both diseased and healthy plant leaf images. The dataset encompasses images of various plant species affected by multiple diseases, providing a diverse and extensive collection for training and testing recognition algorithms. However, the exact distribution of images across different crops is unclear, as the dataset is not publicly accessible due to policy restrictions imposed by the author. Only a limited subset of images is available through online resources.

\subsubsection{Cross-Dataset Imbalance Comparison}
While intra-crop imbalance reveals how unevenly disease classes are distributed within individual crops, a broader structural understanding requires comparison across entire multi-plant datasets. In this subsection, we extend the analysis by examining cross-dataset imbalance characteristics using two complementary measures: (i) the average crop-level normalized imbalance ratio (NIR), which reflects intra-crop skewness aggregated across species, and (ii) inter-crop variance, which captures dominance differences between crops within the same dataset. 

{\small
\begin{equation}
\text{NIR}_i = \frac{N_{\max,i} - N_{\min,i}}{N_{\max,i}}
\end{equation}

\begin{equation}
\text{Average Crop-Level NIR} = \frac{1}{K} \sum_{i=1}^{K} \text{NIR}_i
\end{equation}
}

{\footnotesize
\noindent where:
\begin{itemize}
\setlength\itemsep{0pt}
\setlength\parskip{0pt}
\setlength\parsep{0pt}
    \item $\text{NIR}_i$ represents the normalized imbalance ratio of the $i^{th}$ crop.\item $N_{\max,i}$ : maximum number of images among all disease classes of crop $i$
    \item $N_{\min,i}$ : minimum number of images among all disease classes of crop $i$
    \item $K$ : total number of crops in the multi-plant dataset
\end{itemize}
}
For multi-plant datasets, calculating imbalance at the global class level may produce distorted interpretations due to species-level dominance or missing classes. Therefore, we first compute the NIR separately for each crop within a dataset and then average these values to obtain a dataset-level structural imbalance estimate. This approach ensures that each crop contributes proportionally to the imbalance metric and prevents skew caused by crops with exceptionally large sample volumes.

The results, summarized in Table~\ref{tab:cross_dataset_imbalance}, reveal several notable patterns. First, structural imbalance is prevalent across nearly all representative multi-plant datasets. Even widely adopted benchmarks such as PlantVillage exhibit high average crop-level NIR values, indicating substantial intra-crop skewness. Similarly, datasets such as AI Challenger dataset demonstrate pronounced imbalance despite their large scale, suggesting that increasing dataset size alone does not guarantee balanced class distributions.

Second, inter-crop variance highlights another important structural issue: unequal species representation. In several datasets, a small number of crops dominate the overall image volume, while others are represented by comparatively fewer samples. This phenomenon is particularly evident in datasets with high inter-crop variance values, where crop-level dominance may bias model training toward specific species.

Moreover, environmental context does not inherently resolve imbalance. Laboratory-acquired datasets typically exhibit structured and controlled distributions but still demonstrate high imbalance levels. Conversely, field-acquired datasets introduce natural variability but often retain moderate to high structural skewness. This finding suggests that dataset realism and dataset balance are distinct properties, both of which influence model generalization in different ways.

\begin{table*}[ht]
\centering
\footnotesize
\caption{Cross-dataset structural imbalance comparison across representative multi-plant plant disease datasets.}
\vspace*{-3mm}
\label{tab:cross_dataset_imbalance}
\begin{tabular}{p{3.2cm}p{1.0cm}p{1.2cm}p{2.0cm}p{2.0cm}p{1.8cm}p{2.0cm}}
\hline
\textbf{Dataset} & \textbf{Total Crops} & \textbf{Total Images} & \textbf{Avg. Crop-Level NIR} & \textbf{Inter-Crop Variance} & \textbf{Variance Level} & \textbf{Acquisition} \\
\hline

PlantVillage & 14 & 54,306 & 0.77 & 0.97 & High & Lab \\ \hline
PlantDoc & 13 & 2,569 & 0.68 & 0.92 & High & Mixed \\ \hline
AI Challenger & 10 & 26,747 & 0.77 & 0.92 & High & Lab \\ \hline
NZDL Fruit Dataset & 5 & 3,545 & 0.53 & 0.63 & Moderate & Field \\ \hline
Turkey Plant Dataset & 7 & 4,447 & 0.79 & 0.95 & High & Field \\ \hline
FieldPlant Dataset & 3 & 8,629 & 0.99 & 0.18 & Moderate & Field \\ \hline
Multi\_Plant Datset & 11 & 4,079 & 0.37 & 0.81 & Moderate & Field \\ \hline
UCI Rice \& Wheat & 2 & 200 & 0.00 & 0.33 & Low & Lab \\ \hline
Rice \& Betel Dataset & 2 & 3,710 & 0.02 & 0.63 & Moderate & Mixed \\ \hline
\end{tabular}
\end{table*}

\subsection{Comparative Analysis of Structural Imbalance in Plant Disease Datasets}
The preceding sections examined the structural properties of single-plant and multi-plant disease datasets independently. In this subsection, we synthesize these observations to provide a unified perspective on structural imbalance across plant disease data resources.

A key distinction emerges between single-plant and multi-plant datasets. Single-plant datasets typically exhibit imbalance primarily at the intra-class level, where certain disease categories dominate others within the same crop. In several cases, imbalance is amplified by limited sample sizes or the presence of dominant “healthy” classes. However, structural complexity remains relatively contained because the dataset focuses on a single species. As a result, while class skewness is common, cross-species dominance is absent.

In contrast, multi-plant datasets introduce a layered imbalance structure. First, intra-crop imbalance persists within each species, often mirroring the skew observed in single-plant datasets. Second, an additional dimension of imbalance emerges at the inter-crop level, where certain species contribute disproportionately to the total dataset volume. This dual level imbalance—within crops and across crops creates hierarchical skewness that may significantly influence model learning behavior.

The cross-dataset comparison further reveals that structural imbalance is not strongly correlated with dataset scale. Large benchmark datasets, despite containing tens of thousands of images, frequently demonstrate high average crop-level NIR values. Conversely, some smaller datasets exhibit relatively moderate imbalance. This indicates that increasing dataset size alone does not inherently resolve skewness; balanced sampling strategies remain essential.

Environmental context also does not eliminate imbalance. Laboratory-acquired datasets often display structured but uneven class distributions, while field-acquired datasets introduce natural variability yet still retain moderate to high imbalance levels. Thus, dataset realism and dataset balance represent distinct structural properties that must be addressed independently in model design.

Overall, the comparative analysis highlights three persistent structural challenges in plant disease datasets:
\begin{itemize}
\setlength{\itemsep}{2pt}
    \setlength{\parskip}{0pt}
    \setlength{\parsep}{0pt}
    \item Intra-class imbalance within crops

    \item Inter-crop dominance in multi-plant datasets

    \item Inconsistent representation of healthy versus diseased samples
\end{itemize}
\par
These findings reinforce the central argument of this review: dataset structure plays a decisive role in shaping model performance, robustness, and real-world deployment readiness. Addressing architectural improvements without systematically considering dataset imbalance may yield limited practical benefit.

\section{Class Imbalance and Distribution Bias}
\label{sec:Class Imbalance and Distribution Bias}
Although the structural analysis in Section \ref{sec:Critical Analysis of Plant Disease Datasets} revealed varying degrees of imbalance across plant disease datasets, understanding the numerical distribution alone is insufficient. Class imbalance directly influences model training dynamics, feature learning, optimization behavior, and predictive reliability. In deep learning systems, uneven data distributions may introduce systematic bias toward dominant classes, thereby compromising minority disease detection and limiting real-world applicability. This section examines the quantitative severity of imbalance and analyzes its implications for model learning and generalization.

\subsection{Quantitative Imbalance Analysis}
The structural analysis presented in Section \ref{sec:Critical Analysis of Plant Disease Datasets} revealed that imbalance is a recurring characteristic across both single-plant and multi-plant disease datasets. However, beyond reporting numerical measures such as the normalized imbalance ratio (NIR), it is essential to interpret what these values signify in terms of dataset quality and representational fairness.

Across single-plant datasets, imbalance typically manifests as unequal distribution of images among disease categories within the same crop. In several cases, one disease class contains several times more samples than others, while certain datasets exhibit dominance of either healthy or diseased classes. Although some datasets demonstrate relatively low skewness, many exhibit moderate to high imbalance levels, indicating that minority disease categories are underrepresented during training. This uneven representation may introduce bias even before model optimization begins.

In multi-plant datasets, imbalance becomes structurally more complex. The average crop-level NIR analysis shows that many crops within the same dataset exhibit substantial intra-crop skewness. Moreover, cross-dataset comparisons indicate that inter-crop variance is frequently moderate to high, meaning that certain plant species dominate the dataset in terms of image volume. This layered imbalance within crops and across crops suggests that representation disparities are not isolated phenomena but are embedded within dataset construction practices.

A notable pattern observed across multiple datasets is the inconsistent inclusion of healthy classes. In some datasets, only diseased samples are collected, while in others healthy classes dominate the distribution. Such inconsistencies create structural distortions that may affect the stability of imbalance metrics and complicate fair comparison across datasets. The normalized imbalance formulation adopted in this study helps mitigate these issues by bounding skewness between 0 and 1; nevertheless, high NIR values indicate significant dominance of certain categories.

\begin{figure}[ht]
\centering
\includegraphics[width=0.4\textwidth]{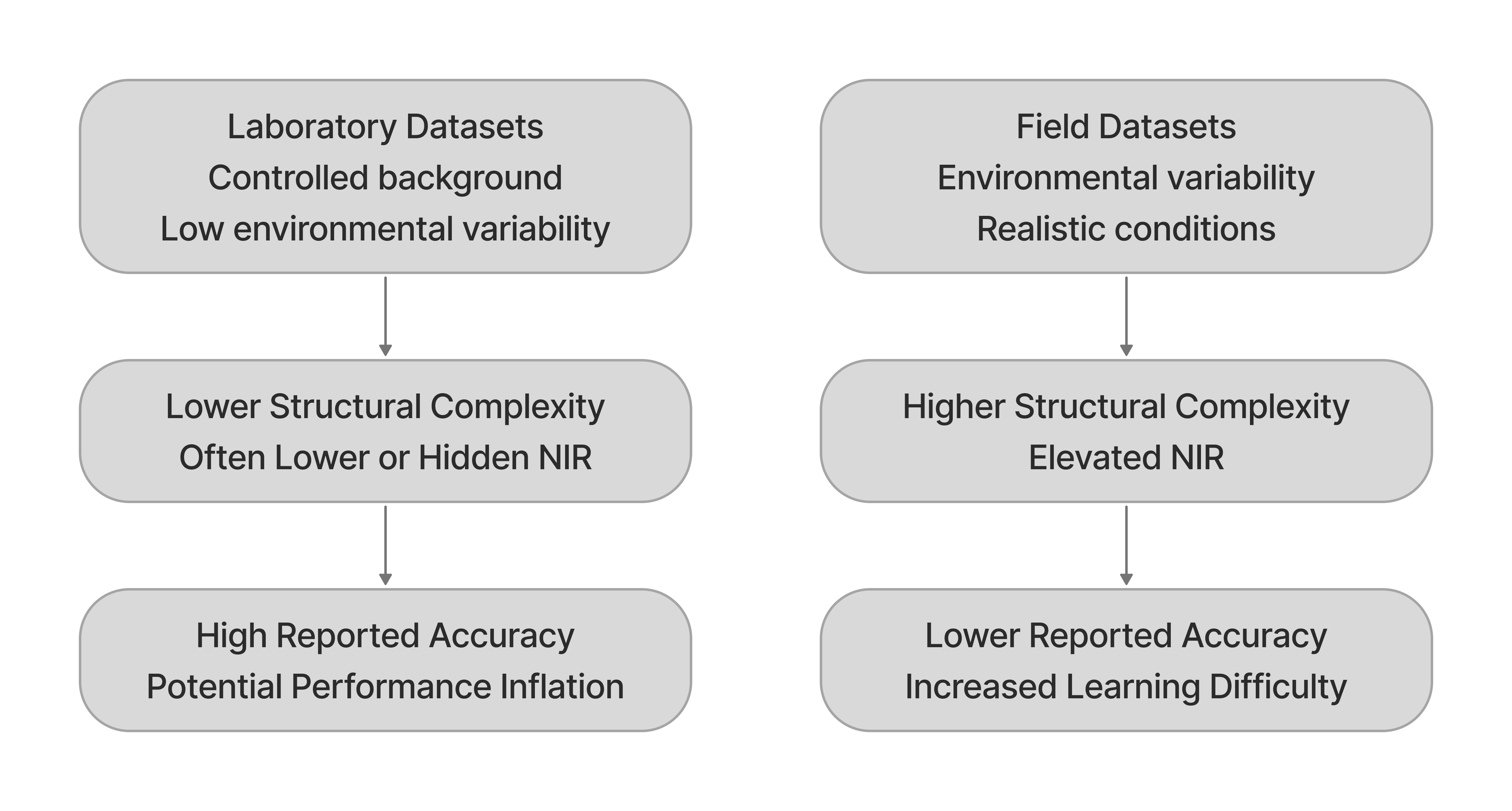}
\vspace*{-6mm}
\caption{Structural contrast between laboratory and field datasets and their influence on imbalance severity (NIR), learning complexity, and reported performance inflation.}
\label{fig:lab_field_nir}
\end{figure}

Importantly, dataset scale does not guarantee structural balance. Large benchmark datasets often exhibit imbalance levels comparable to, or even greater than, smaller datasets. This observation suggests that imbalance is not merely a consequence of limited data availability but is frequently tied to data collection priorities, disease prevalence assumptions, and annotation practices. The structural differences between laboratory and field datasets further clarify why reported model performance must be interpreted in context. Figure~\ref{fig:lab_field_nir} illustrates how acquisition setting influences dataset imbalance severity and perceived accuracy.
\vspace*{-3mm}

\begin{figure}[H]
\centering
    \includegraphics[width=0.2\textwidth]{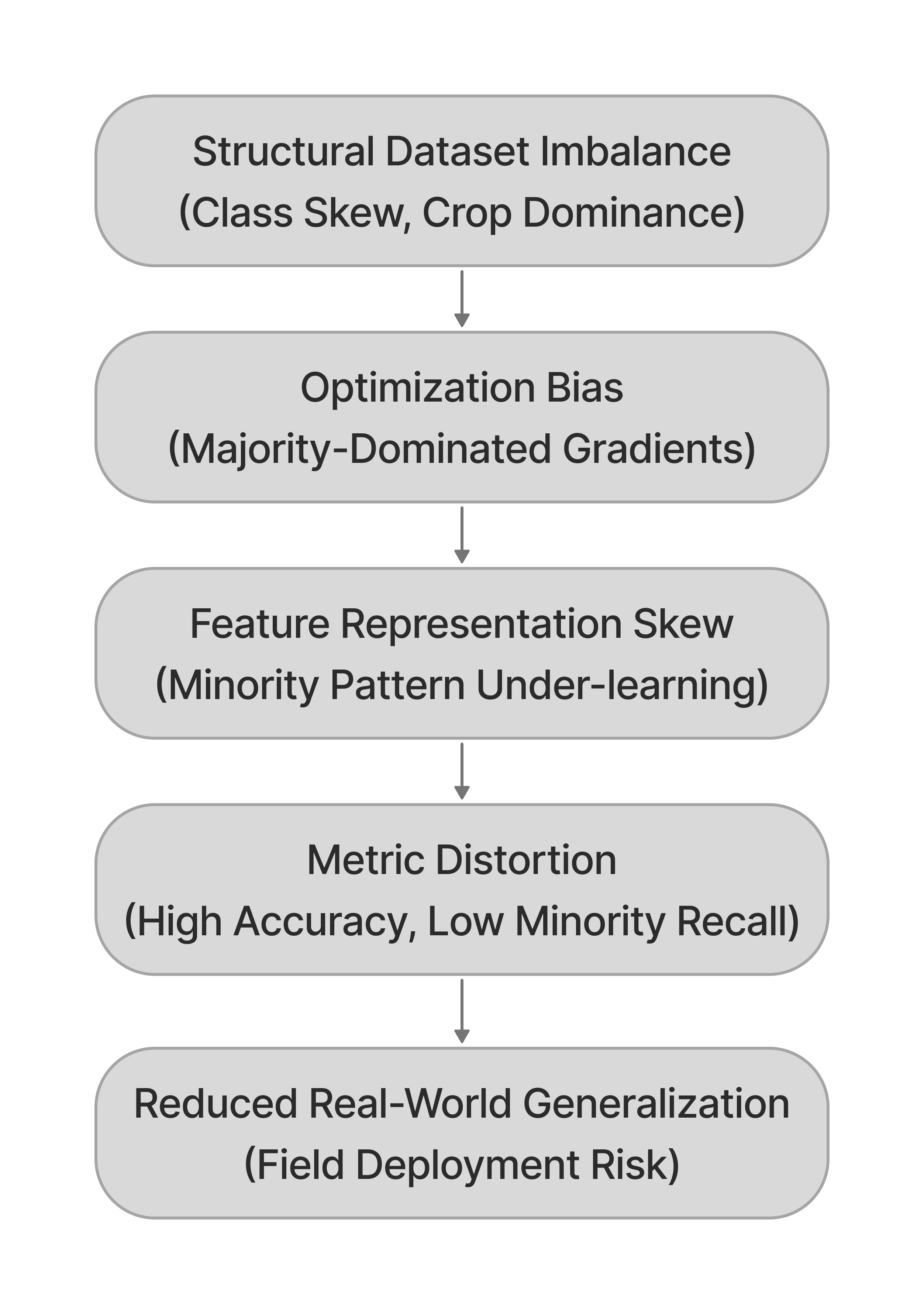}
    \vspace*{-6mm}
    \caption{Conceptual pathway illustrating how structural dataset imbalance propagates into model optimization bias, representation skewness, evaluation distortion, and reduced generalization performance.}
\label{fig:imbalance_pipeline}
\end{figure}

Overall, the quantitative evidence indicates that imbalance is pervasive across contemporary plant disease datasets. While some datasets demonstrate relatively moderate skewness, many exhibit pronounced distribution asymmetry. Such structural disparities form the foundation upon which deep learning models are trained, making imbalance not only a statistical property but a critical factor influencing downstream learning behavior. As illustrated in Figure~\ref{fig:imbalance_pipeline}, structural dataset imbalance can propagate through the learning process, influencing model optimization bias, feature representation, evaluation outcomes, and ultimately the generalization capability of disease detection models.
\vspace*{-3mm}

\par

\subsection{Implications of Structural Imbalance for Model Learning}
While quantitative imbalance analysis provides insight into dataset structure, its true significance lies in how it influences model behavior. Deep learning models are inherently data driven, and their optimization process is directly shaped by the statistical properties of the training distribution. Consequently, structural imbalance in plant disease datasets can systematically bias model learning, affecting feature representation, decision boundaries, and predictive reliability.

One of the primary consequences of class imbalance is optimization bias. Standard loss functions, such as categorical cross-entropy, implicitly assume balanced class distributions. During training, gradients are dominated by majority classes because they contribute more frequently to the loss. As a result, models tend to minimize overall error by prioritizing dominant categories, often at the expense of minority disease classes. This can lead to reduced recall for rare diseases, even when overall accuracy appears high.

Feature representation is also influenced by distribution skewness. In highly imbalanced datasets, dominant classes shape the learned feature space, causing the model to develop strong discriminative patterns for frequently occurring diseases while insufficiently encoding features associated with underrepresented classes. This imbalance in representation can result in unstable predictions when minority disease samples are encountered, particularly in real-field scenarios where disease prevalence may differ from the training distribution.

In multi-plant datasets, hierarchical imbalance introduces additional complexity. When certain crops dominate the dataset volume, models may implicitly prioritize species-specific features associated with those crops. This can reduce cross-species generalization and limit the robustness of the model when deployed in diverse agricultural environments. Thus, inter-crop imbalance may bias not only disease recognition but also plant-level representation learning.

Evaluation metrics further complicate the issue. Accuracy, which is frequently reported in plant disease detection studies, may not adequately reflect model performance under imbalance conditions. A model trained on a skewed dataset may achieve high overall accuracy while failing to correctly identify minority diseases. Without careful use of class-sensitive metrics such as precision, recall, F1-score, or balanced accuracy, imbalance-induced bias may remain undetected.

Another important implication concerns generalization under environmental variability. Field-acquired datasets already introduce visual complexity through background clutter, lighting variation, and natural occlusions. When structural imbalance is combined with environmental variability, minority classes become even more difficult to learn effectively. This compounded challenge may partly explain why models that perform well in controlled laboratory settings often struggle in real-world agricultural conditions.

Data augmentation is commonly employed to mitigate imbalance by artificially increasing the number of samples in minority classes \cite{seetharaman2022leaf,saleem2021mango,jiang2021recognition}. While augmentation can improve apparent model performance, it does not fundamentally alter the underlying data distribution. Synthetic transformations cannot fully replicate the diversity of real disease manifestations, particularly in field environments. Therefore, augmentation should be viewed as a complementary strategy rather than a complete solution to imbalance.  The major learning implications associated with different forms of structural imbalance are summarized in Table~\ref{tab:imbalance_implications}.

\begin{table}[ht]
\centering
\footnotesize
\caption{Model-level implications of structural imbalance in plant disease datasets.}
\vspace*{-3mm}
\label{tab:imbalance_implications}

\begin{tabular}{p{3.2cm}p{4.2cm}p{4.2cm}}
\hline
\textbf{Imbalance Type} & \textbf{Impact on Model Learning} & \textbf{Potential Risk} \\
\hline

Intra-Class Imbalance &
Majority classes dominate gradient updates and decision boundaries &
Low recall for minority diseases; biased classification \\ \hline

Inter-Crop Imbalance &
Species with higher sample volume shape learned representations &
Reduced cross-species generalization \\ \hline

Missing or Underrepresented Healthy Classes &
Distorted separation between healthy and diseased states &
Overconfident predictions in real-world deployment \\ \hline

Long-Tail Disease Distribution &
Rare diseases insufficiently encoded in feature space &
Poor detection of uncommon or emerging diseases \\ \hline

\end{tabular}

\end{table}

\par
While class imbalance is often acknowledged in plant disease datasets, its influence on model performance is rarely examined in a structured manner. Deep learning models trained on skewed datasets tend to optimize overall accuracy by prioritizing majority classes, thereby underperforming on minority disease categories. This phenomenon becomes particularly critical in agricultural contexts, where rare diseases may carry significant economic consequences despite limited representation in the dataset.

Datasets exhibiting high normalized imbalance ratio (NIR) values frequently require additional mitigation strategies, including data augmentation, class-weighted loss functions, oversampling, or ensemble learning approaches. In several reviewed studies, ensemble or hybrid architectures were employed on datasets with pronounced skewness, suggesting that architectural complexity often compensates for structural data imbalance rather than purely enhancing feature representation.

Moreover, laboratory-based datasets, which often exhibit lower intra-crop variability and simplified background conditions, tend to report higher classification accuracy. However, these performance gains may partially reflect reduced structural and environmental complexity rather than genuine model robustness. In contrast, field-acquired datasets with higher NIR values and greater environmental heterogeneity typically demonstrate lower reported accuracy, indicating increased learning difficulty.

This pattern suggests that reported model performance cannot be interpreted independently of dataset structure. High accuracy on structurally simplified datasets does not necessarily translate to generalizable performance in field conditions. Therefore, imbalance metrics such as NIR provide critical contextual information for interpreting model results and for distinguishing between architectural effectiveness and dataset-driven performance inflation. Although a direct statistical meta-analysis between NIR and reported accuracy is beyond the scope of this review, the consistent structural patterns observed across datasets suggest a strong contextual relationship between imbalance severity and reported model behavior. Table~\ref{tab:nir_behavior} summarizes the conceptual relationship between NIR levels, typical model learning behavior, and potential deployment risks in plant disease detection systems.

\begin{table}[H]
\centering
\footnotesize
\caption{Relationship between normalized imbalance ratio (NIR) and typical model learning behavior in plant disease detection datasets.}
\vspace*{-3mm}
\label{tab:nir_behavior}
\begin{tabular}{p{1.2cm}p{4.0cm}p{4.8cm}p{4.0cm}}
\hline
\textbf{NIR Range} & \textbf{Structural Interpretation} & \textbf{Typical Model Behavior} & \textbf{Deployment Risk} \\ \hline

0.0 -- 0.3 &
Relatively balanced class distribution &
Stable training; improved per-class recall; reduced need for heavy augmentation &
Lower risk of minority-class suppression; better generalization potential \\ \hline

0.3 -- 0.7 &
Moderate imbalance across disease categories &
Bias toward majority classes; reliance on augmentation or class-weighting; possible overfitting &
Reduced reliability for underrepresented diseases in field conditions \\ \hline

0.7 -- 1.0 &
Severe skewness; dominance of few classes or absence of certain categories &
Majority-class dominance; inflated overall accuracy; poor minority recall; architectural compensation (ensemble/hybrid models) &
High risk of failure in real-field deployment, especially for rare but critical diseases \\ \hline

\end{tabular}
\end{table}

\section{Deep Learning Models Across Plant Disease Datasets}
\label{sec:Deep Learning Models Across Plant Disease Datasets}
Deep learning architectures for plant disease detection have evolved rapidly over the past decade, ranging from single convolutional neural networks to ensemble and hybrid frameworks. However, model selection and reported performance are closely tied to the datasets on which these architectures are trained and evaluated. Rather than examining models in isolation, it is essential to analyze how architectural strategies interact with dataset structure, imbalance, and environmental variability. This section provides a dataset-aware examination of deep learning model usage across plant disease benchmarks.

\subsection{Model Usage Patterns Across Dataset Type}
Deep learning architectures for plant disease detection have expanded considerably, yet their adoption is not uniform across dataset categories. Instead, model selection often reflects the structural properties, scale, and acquisition conditions of the datasets on which they are evaluated. Understanding these usage patterns is essential for assessing whether architectural advancements are genuinely driven by methodological innovation or by dataset characteristics.

In single-plant datasets, model usage tends to favor relatively straightforward architectures. Transfer learning with pre-trained convolutional neural networks (such as VGG, ResNet, DenseNet, or MobileNet variants) is particularly common. Because single-plant datasets typically focus on a limited number of disease classes, researchers often prioritize fine-tuning existing architectures rather than developing complex hybrid or ensemble models. This approach is computationally efficient and yields strong performance when datasets are moderately sized and collected under controlled conditions. However, the reliance on pre-trained models also suggests that architectural complexity is not always necessary to achieve high reported accuracy on structurally constrained datasets.
\par There are plenty of single CNN deep learning models employed for plant leaf disease detection, such as AlexNet \cite{krizhevsky2017imagenet}, VGGNet \cite{simonyan2014very}, ResNet \cite{he2016deep}, DenseNet \cite{huang2017densely}, MobileNet, and others. Various methods for detecting plant diseases have been proposed using single CNN (Convolutional Neural Network) models. One notable example is the application of AlexNet, achieving an accuracy of 86.5\% in identifying diseases in grape leaves \cite{agarwal2019grape} and with tomato leaf images, where it reached an accuracy of 95.75\% \cite{ashok2020tomato}. Numerous studies have utilized VGGNet to detect plant diseases. For instance, VGG-16 was employed in one survey to identify grape leaf diseases, achieving an accuracy of 98.4\% \cite{thet2020grape}. DenseNet121, a different single CNN model architecture, has a 93.5\% accuracy rate in predicting apple leaf disease \cite{zhong2020research}. Furthermore, researchers fine-tune the parameters of these models for optimal performance, as seen with the VGG model used for identifying diseases in rice leaf images \cite{jiang2021recognition}. Table~\ref{tab:single_dl_frameworks} below presents various research studies on single CNN model architectures for plant diseases.

{\footnotesize
\setlength{\LTleft}{0pt}
\setlength{\LTright}{0pt}

\begin{longtable}{>{\bfseries}p{0.6cm} p{4.3cm} p{5.3cm} p{2.7cm} p{2.5cm}}
\caption{Single deep learning frameworks applied to plant disease datasets.}
\vspace*{-3mm}
\label{tab:single_dl_frameworks} \\

\hline
\textbf{Ref} & \textbf{Dataset(s) Name} & \textbf{Learning Model} & \textbf{Data Augmentation Used} & \textbf{Model Accuracy (\%)} \\
\hline
\endfirsthead

\hline
\textbf{Ref} & \textbf{Dataset Name} & \textbf{Learning Model} & \textbf{Data Augmentation Used} & \textbf{Model Accuracy (\%)} \\
\hline
\endhead

\hline
\endfoot

\small

\cite{agarwal2019grape} & PlantVillage Dataset & AlexNet & Yes & 86.5\% \\ \hline
\cite{srivastava2024plant} & PlantVillage Dataset & Comparison between VGG16, MobileNetV2, InceptionV3, Xception and DenseNet121 (Modified Transfer Learning) & Yes & DenseNet121 achieved the highest accuracy (99.9\%) \\ \hline
\cite{varma2025automatic} & Proprietary Dataset (Mango Dataset collected from different fileds in Bangladesh) \cite{ahmed2023mangoleafbd} & Comparison between VGG19, InceptionV3, ResNet152V2, DenseNet121, InceptionResNetV2, MobileNetV2 and Xception & Yes & InceptionV3 achieved the highest accuracy (99.87\%) \\ \hline
\cite{bhagat2024optimized} & PlantVillage Dataset & CNN Models - VGG16 and Xception & Yes &
VGG with 98.18\% \\ \hline

\cite{ashok2020tomato} & PlantVillage Dataset & AlexNet & NA & 95.75\% \\ \hline
\cite{thet2020grape} & Proprietary Dataset (Images collected from Myanmar Grapevine Yard) & VGGNet 16 & Yes & 98.4 \% \\ \hline
\cite{zhong2020research} & AI-Challenger dataset & DenseNet 121 & No & 93.5\% \\ \hline
\cite{too2019comparative} & PlantVillage Dataset & Comparison between VGG 16, Inception V4, ResNet with 50, ResNet with 101, ResNet with 152, and DenseNet with 121 layers & No & 98.08\%, 81.83\%, 99.59\%, 99.66\%, 99.59\%, 99.75\% \\ \hline
\cite{elfatimi2022beans} & Proprietary Dataset (Images collected from fields by Makerere AI lab in collaboration with the National Crops Resources Research Institute, Uganda) & MobileNet & No & 97\% \\ \hline
\cite{seetharaman2022leaf} & Proprietary Dataset (Images collected from farms) & RCNN & No & 98\% \\ \hline
\cite{shehu2025early} &
PlantVillage + Proprietary Dataset (TomatoEbola) &
Vision Transformer (ViT-ImageNet, ViT-Base, ViT-Small) & Yes & ViT-Base with 99.17\% \\ \hline

\cite{singh2022deep} & PlantVillage Dataset & AlexNet & Yes & 99.16\% \\ \hline
\cite{chen2022mobile} & AI Challenger and PlantVillage & MobileNetV2 & Yes & 97.33\% \\ \hline
\cite{ashwinkumar2022automated} & PlantDoc Dataset & Single architecture model OMNCNN (Optimal mobile network-based convolutional neural network) & NA & 98.80\% \\ \hline
\cite{khatoon2021image} & Proprietary Dataset (Images collected from fields by King Faisal University) & ShallowNet8, VGGNet-16, ResNet-50, ResNet 152, and DenseNet 121 are compared for performance. & No & 78.05\%, 80.32\%, 95.31\%, 92.01\%, 90.85\% \\ \hline
\cite{bevers2022pictures} & Proprietary Dataset (ASDID - Images obtained from three Agricultural Research Stations in Alabama, US) & DenseNet121, ResNet50, VGG16, and Xception architectures are used & Yes & DenseNet 121 achieved an accuracy of 96.8\% \\ \hline
\cite{khan2022deep} & Proprietary Dataset (Images collected from real fields of orchids) & EfcientNet, FR-CNN, and YOLOv4 & Yes & 87.9\% \\ \hline
\cite{liu2022ds} & Proprietary Dataset (Images collected from real fields in Changsha City, China) & Modified DenseNet121 & No & 95.02\% \\ \hline
\cite{rangarajan2022detection} & Proprietary Dataset (Images collected from the wheat canopy using hyperspectral imaging and scanned in the laboratory) & Eight different lightweight CNN models were used that are pre-trained and require limited computing resources. & Yes & DarkNet 19 model with accuracy of 100 \% utilizing spectral line graph images \\ \hline
\cite{saleem2021mango} & Proprietary Dataset (Images were collected from fields of Multan, Lahore, and Faisalabad, Pakistan) & RCNN & Yes & 93.8\% \\ \hline
\cite{jiang2021recognition} & 3 Database Sources (UCI machine learning database, Internet, and Proprietary) & VGG-16 & Yes & 97.22\% \\ \hline
\cite{tassis2021deep} & Proprietary Dataset (BRACOT - The dataset was collected from the mountain regions of Marechal Floriano, state of Espirito Santo, Brazil) & ResNet & Yes & 94.25\% \\ \hline
\cite{moupojou2023fieldplant} & Proprietary Dataset (FieldPlant dataset – images collected from the Cameroon plantations fields) & Performance comparisons are made for four CNN models: MobileNet, VGG16, InceptionV3, and InceptionResNetV2. & No & 82.9\%, 88.54\%, 81.81\%, 82.54\% \\ \hline
\cite{kaya2023novel} & PlantVillage Dataset & Dual-input model fusion based DenseNet121 & Yes & 98.17\% \\ \hline
\cite{haridasan2023deep} & Proprietary Dataset (Images of diseased leaves collected from a laboratory) & SVM, CNN & Yes & 91.45\% \\ \hline
\cite{kumar2023systematic} & PlantVillage Dataset & 10 single architecture deep learning models such as DenseNet201, DenseNet121, NasNetLarge, Xception, ResNet152V2, EfficientNetB5, EfficientNetB7, VGG19, MobileNetV2 and a hybrid model (EfficientNetB7 and ResNet152V2) are compared. & Yes & DenseNet achieves the best accuracy with 98.67\% \\ \hline
\cite{tembhurne2023plant} & Unified 4 datasets– PlantVillage, Cassava leaf, Plant leaf, and Flower dataset & MobileNetV1 & Yes & 96\% \\ \hline
\cite{nawaz2024coffeenet} & Arabica coffee Dataset \cite{jepkoech2021arabica} & Improved version of CenterNet with ResNet-50 as its backbone and a spatial-channel attention mechanism. & Yes & 98.54\% \\ \hline
\cite{jepkoech2021arabica} & PlantVillage Dataset & Three single architecture CNN models, AlexNet, VGG16, and VGG19, were used for plant disease diagnosis. & Yes & AlexNet: 96.63\% VGG16: 95.05\% VGG19: 95.22\% \\ \hline

\end{longtable}
}

\par
In contrast, multi-plant datasets demonstrate greater architectural diversity. The presence of multiple species and larger class counts often motivates the use of ensemble methods, deeper residual networks, or hybrid frameworks combining feature extractors with attention mechanisms or auxiliary classifiers. In several studies, ensemble models are employed to improve robustness across species, while hybrid architectures are introduced to handle increased intra-class variability. Nevertheless, such architectural enhancements are frequently evaluated on benchmark datasets where class imbalance and environmental complexity are not systematically controlled. For instance, an ensemble of AlexNet, GoogleNet, ResNet18, ResNet50, ResNet10 and DenseNet201 CNNs combines predictions from multiple base models to predict diseases from multiple crops \cite{tugrul2022convolutional}. Table~\ref{tab:ensemble_dl_frameworks} below presents various research studies on ensemble CNN model architectures for detecting different plant diseases.Hybrid CNN models combine various model strengths, often using pre-trained CNNs as feature extractors with additional layers tailored for plant disease detection. Variants of CNNs, such as AlexNet, VGGNet, GoogleNet, ResNet etc, are often combined to enhance accuracy and efficiency in detecting plant diseases. For instance, VGG-19 with EfficientNetB0 and MobileNetV2 combined were used for detecting diseases in tomato leaves, reaching an accuracy of 96.86\% \cite{bir2020transfer}. Additionally, Jiang et al. \cite{jiang2019real} modified VGGNet by integrating it with the Inception network, which resulted in a 78.80\% accuracy for apple leaf disease detection. ResNet has also been employed to detect plant diseases. Wu et al. \cite{wu2019identification} used variant ResNet-50 AlexNet and GoogLeNet and achieved an accuracy of 94.29\% for soybean leaf disease detection. Another variant, a customized ResNet model reached an accuracy of 82.78\% on a lab-based dataset \cite{guan2021novel}. A hybrid version of the ResNet model is used by Hu et al. \cite{hu2020mdfc}, containing two ResNet-50 models and one ResNet-34 model to recognize different crop diseases. Table~\ref{tab:hybrid_dl_frameworks} below presents various research studies on hybrid CNN model architectures for detecting different plant diseases. Transfer learning has gained traction, allowing researchers to fine-tune pre-trained models like GoogLeNet and InceptionV3 on specific datasets, which enhances performance while reducing the need for extensive labeled data. Various related works in the field of disease detection include Ji et al. \cite{ji2020automatic} used a combination of two architectures, GoogLeNet and ResNet, for grape leaf disease detection and achieved an accuracy of 99.17\%. InceptionV3 which is the third version of GoogleNet, has been applied for plant disease detection and achieved an accuracy of 98.42\% \cite{hassan2021identification}. More recently, Generative Adversarial Networks (GANs) \cite{jin2022grapegan,chen2023apple} have emerged, offering innovative solutions for data augmentation by generating synthetic images to enhance model training, especially when dealing with limited datasets. Furthermore, Long Short-Term Memory (LSTM) \cite{liu2021plant} networks are designed to recognize and understand patterns over time. This unique structure makes them useful for analyzing complex sequences, such as the cycles of disease outbreaks. Transformer models, known for their attention mechanisms, are beginning to be applied for their efficiency in handling large datasets and complex patterns \cite{brown30plant}.

\begin{table}[H]
\centering
\footnotesize
\caption{Ensemble deep learning frameworks applied to plant disease datasets.}
\vspace*{-3mm}
\label{tab:ensemble_dl_frameworks}

\begin{tabular}{>{\bfseries}p{0.6cm} p{4.3cm} p{5.3cm} p{2.5cm} p{2.5cm}}
\hline
\textbf{Ref} & \textbf{Dataset(s) Name} & \textbf{Learning Model} & \textbf{Data Augmentation Used} & \textbf{Model Accuracy (\%)} \\ 
\hline

\cite{pacal2025utilizing} & PlantVillage Dataset & Ensemble of CNN (VGG-16, ResNet-50, DenseNet-121, Xception) and ViT (MobileViT, MViTv2, DeiT3, MaxViT) & Yes & Ensemble of DeiT3 and MaxViT achieved accuracy of 99.83\% \\ \hline

\cite{yousafzai2025multi} & Kaggle Wheat Disease Datasets &
DenseNet201, MobileNetV2, Custom CNN (Ensemble - Bagging based Multi stage CNN) & No & 99.16\% \\ \hline

\cite{oad2024plant} &
PlantVillage Dataset & VGG16, VGG19, InceptionV3, ResNet101V2 (Ensemble Voting) & Yes & 92.3\% \\ \hline

\cite{ji2020automatic} & PlantVillage Dataset & United Ensemble Model - GoogLeNet and ResNet & Yes & 98.57\% \\ \hline

\cite{amin2022end} & PlantVillage Dataset & EfcientNetB0 and DenseNet 121 in ensemble mode. & No & 98.56\% \\ \hline

\cite{turkoglu2022plantdiseasenet} & Proprietary Dataset (The Turkey-Plant Dataset – images collected from the agricultural fields of the Bingol and Inonu Universities in Turkey) & AlexNet, GoogleNet, ResNet18, ResNet50, ResNet10 and DenseNet201. & No & 97.56\% \\ \hline

\cite{shovon2023plantdet} & Proprietary Dataset (Rice Leaf dataset and a custom Betel Leaf Dataset from a village of district: Rajshahi, Bangladesh) & An Ensemble model known as PlantDet - based on Inception, ResNetV2, EfficientNetV2L, and Xception. & Yes & Rice: 98.53\% Betel: 97.50\% \\ \hline

\cite{jha2024deep} & PlantVillage Dataset & Three CNN models, MobileNet, Inception, and ResNet combined into a deep stacking ensemble model for potato disease prediction. & Yes & 98.86\% \\ \hline

\end{tabular}
\end{table}

{\footnotesize
\setlength{\LTleft}{0pt}
\setlength{\LTright}{0pt}

\begin{longtable}{>{\bfseries}p{0.6cm} p{4.3cm} p{5.3cm} p{2.5cm} p{2.5cm}}
\caption{Hybrid deep learning frameworks applied to plant disease datasets.}
\vspace*{-3mm}
\label{tab:hybrid_dl_frameworks} \\

\hline
\textbf{Ref} & \textbf{Dataset(s) Name} & \textbf{Learning Model} & \textbf{Data Augmentation Used} & \textbf{Model Accuracy (\%)} \\
\hline
\endfirsthead

\hline
\textbf{Ref} & \textbf{Dataset Name} & \textbf{Learning Model} & \textbf{Data Augmentation Used} & \textbf{Model Accuracy (\%)} \\
\hline
\endhead

\hline
\endfoot

\small
\cite{aboelenin2025hybrid} & PlantVillage Dataset (Apple, Corn) & VGG16 + Inception-V3 + DenseNet201 + ViT (Hybrid (CNN Ensemble + Vision Transformer)) & Yes & 99.24\% (Apple), 98\% (Corn) \\ \hline

\cite{adem2023sugar} & Proprietary Dataset (Images collected from fields) & Four CNN models VGG16, SSD, Faster R-CNN, and Yolov4 were used directly, with Yolov4 employed in a hybrid approach. & Yes & 96.47\% \\ \hline

\cite{brown30plant} & Proprietary Dataset (Images captured using multispectral camera) & CNN (ResNet) with Vision Transformers & Yes & 88.88\% \\ \hline

\cite{thakur2024sunet} &
Proprietary Dataset (Coffee \cite{jepkoech2021arabica}) &
SUNet (SegNet encoder + U-Net decoder + VGG16 backbone + Mask R-CNN + PSPNet) & Yes &
98.45\% \\ \hline

\cite{reddy2021deep} & PlantVillage Dataset & AlexNet and GoogleNet & NA & 97.62 \% \\ \hline

\cite{bir2020transfer} & PlantVillage Dataset & VGGNet 19 with EfficientNetB0, MobileNetV2 & Yes & 96.86 \% \\ \hline

\cite{kaur2024novel} &
PlantVillage dataset & ResNet-50, ResNet-101 with CBAM attention and SVM classifier (Hybrid - CNN + Attention + SVM) & Yes & 98.72\% \\ \hline

\cite{jiang2019real} & Proprietary Dataset (ALDD - Apple Leaf Disease Dataset) & VGGNet with Inception Network and rainbow concatenation & Yes & 78.80\% \\ \hline

\cite{thai2025mobileh} &
PlantVillage datset and Corn Disease Dataset (Made using  PlantVillage and PlantDoc datasets) & MobileH-Transformer (CNN + Transformer encoder) & No & 97.20\% (Corn), 96.80\% (PlantVillage) \\ \hline

\cite{wu2019identification} & Proprietary Dataset (Images collected from Northeast Agricultural University in Heilongjiang Province) & AlexNet, GoogLeNet and ResNet & Yes & 94.29\% \\ \hline

\cite{ghosh2025advanced} &
Proprietary Dataset (Tomato leaf dataset) &
Swin Transformer + Vision Transformer + EfficientNetV2 (Hybrid model) & Yes &
98\%  \\ \hline

\cite{guan2021novel} & AI-Challenger dataset & Inception, Resnet, Inception Resnet, and DenseNet & Yes & 87\% \\ \hline

\cite{hu2020mdfc} & AI-Challenger dataset & MDFC–ResNet Multidimensional feature compensation residual neural network containing two ResNet-50 models and one ResNet-34 model & Yes & 85.22\% \\ \hline

\cite{hassan2021identification} & PlantVillage Dataset & InceptionV3, InceptionResNetV2, MobileNetV2, and EfficientNetB0 & Yes & 99.56\% \\ \hline

\cite{syed2022citrus} & Citrus Leaves dataset from Kaggle & R-CNN with ResNet101 as a feature extractor and a region proposal network & Yes & 94.37 \% \\ \hline

\cite{shantkumari2023grape} & PlantVillage Dataset & Convolution Neural Network based Classification (CNNC) model and improvised K-Nearest Neighbor (IKNN) model & No & 98\% \\ \hline

\cite{jin2022grapegan} & PlantVillage Dataset & CNN, VGG, and Inception v1 model with GAN & Yes & 96.13\% \\ \hline

\cite{singh2020plantdoc} & PlantDoc Dataset & Comparison between a hybrid CNN model of Faster R-CNN with InceptionResnetV2 model and a single architecture MobileNet model & Yes & Faster R-CNN - mAP of 38.9, MobileNet - mAP of 22 \\ \hline

\cite{dananjayan2022assessment} & Proprietary Dataset (CCL'20 - leaf images collected from Guangzhou, China) & Backbone networks using CenterNet2 include ResNet18 and Resnet101, Hourglass-104, DLA34, Res2Net, and ResNeXt. & No & 91.4\% \\ \hline

\cite{liu2021plant} & Proprietary Dataset (PDD271 - 271 plant disease categories) & SeNet154 ResNet152 LSTM & No & 90.01\% \\ \hline

\cite{gajjar2022real} & PlantVillage Dataset & Single shot detector with MobileNet for classification and a Custom CNN & Yes & 96.88\% \\ \hline

\cite{chen2023apple} & PlantVillage Dataset & Cycle GAN, ResNet & Yes & 97.78\% \\ \hline

\cite{chug2023novel} & Proprietary Dataset (IARI-TomEBD field) & 40 hybrid deep learning models were utilized, containing a combined eight variants of pre-trained EfficientNet architectures (B0–B7) as feature extractors with five machine learning classifiers: LR, kNN, RF, ADB, and SGB. & Yes & HDL models achieved 100\% accuracy \\ \hline

\cite{li2024sugarcanegan} & Proprietary Dataset & A hybrid architecture of ResBet-50 with SugarGAN based on the CycleGAN model. The model included a leaf extraction module and a generator module using U-RSwinT. & Yes & 85.78\% \\ \hline

\end{longtable}
}

\par 
Convolutional Neural Network (CNN) models were initially conceived as powerful tools for general image classification tasks, achieving strong results through rigorous evaluation of standard benchmark datasets. Despite their success in general image classification, many researchers using proprietary datasets have observed that the standard CNN architectures may not be the best solution for specialized leaf disease detection. So, many researchers have created CNN models from scratch or modified existing ones to achieve better performance on specific datasets. For instance, Joshi et al. \cite{joshi2022ricebios} created a custom Convolutional Neural Network (CNN) model to identify diseases in rice leaves. The model incorporates two shallow convolutional layers which facilitate feature extraction, batch normalization to enhance training efficiency and stability, and max-pooling layer to reduce dimensionality. Similarly, Agarwal et al. \cite{agarwal2020toled} developed a new CNN model utilizing tomato images from the Plant Village dataset to classify leaf diseases, achieving an impressive accuracy of 91.2\%, which outperformed other models such as VGG-16 and InceptionV3. Similarly, Sunil et al. \cite{sunil2020empirical} designed a Multi-Layered CNN model that classified leaves from three sub-datasets - Strawberry, Peach, and Pepper from the Plant Village dataset, with accuracy ranging from 87.47\% to 99.25\%. The other modified CNN models were - Hu et al. \cite{hu2020mdfc} used a finetuned version of the ResNet model by combining two ResNet-50 models and one ResNet-34 model to recognize different crop diseases. Jiang et al. \cite{jiang2019real} used a fine-tuned version of VGGNet, VGG with Inception network, and achieved 78.80\% for apple leaf disease detection. Several other CNN models, like R-CNN (Region-based CNN) \cite{seetharaman2022leaf,saleem2021mango}, YOLO \cite{khan2022deep,adem2023sugar}, and some lightweight CNNs like MobileNet \cite{gajjar2022real,moupojou2023fieldplant,kumar2023systematic,tembhurne2023plant} have been employed for disease prediction in plants. Table \ref{tab:custom_dl_frameworks} provides various research studies on custom CNN model architectures for detecting different plant diseases.

\begin{table}[H]
\centering
\footnotesize
\caption{Custom deep learning frameworks applied to plant disease datasets.}
\vspace*{-3mm}
\label{tab:custom_dl_frameworks}

\begin{tabular}{>{\bfseries}p{0.6cm} p{4.0cm} p{6.0cm} p{2.5cm} p{2.5cm}}
\hline
\textbf{Ref} & \textbf{Dataset Name} & \textbf{Learning Model} & \textbf{Data Augmentation Used} & \textbf{Model Accuracy (\%)} \\
\hline

\cite{perez2022exploring} & Proprietary Dataset (Images collected from a vineyard scanned with lab-scale HSI system in 900–1700 nm spectral range) 
& Partial least squares discriminant analysis (PLS-DA) 
& No 
& 97.17\% \\ \hline 

\cite{islam2025plantcarenet} & Combined Dataset (PlantVillage, VegNet, Bangladeshi Crop Disease, Eggplant Dataset \cite{matin2023dataset}) & PlantCareNet (5 convolution layers, 3 pooling layers, followed by fully connected layers before the final Softmax classifier.) & Yes & 97\% \\ \hline 

\cite{agarwal2020toled} & PlantVillage Dataset 
& Custom CNN with 3 convolution and 3 max-pooling layers followed by 2 fully connected layer 
& Yes 
& 91.2\% \\ \hline

\cite{sunil2020empirical} & PlantVillage Dataset 
& Custom CNN incorporates several convolutional layers, with a pooling, activation, and fully connected layer. 
& No 
& 87.47\% to 99.25\% \\ \hline

\cite{nagi2022deep} & PlantVillage Dataset 
& Custom lightweight 6-layer CNN 
& No 
& 98.4\% \\ \hline

\cite{joshi2022ricebios} & Proprietary Dataset (Images collected from a farmland) 
& Custom CNN model designed with two shallow convolutional layers followed by 3 Fully Connected layers, along with batch normalization and max-pooling layers. 
& Yes 
& 93.25\% \\ \hline

\cite{alam2024comparing} & PlantVillage Dataset & Comparison of DenseNet201, EfficientNetB3, EfficientNetB4, InceptionResNetV2, MobileNetV2, ResNet50, ResNet152, VGG16, Xception and a Custom CNN (LDDTA with 3 convolution layers, 2 max-pooling layers, 1 fully connected layer) & Yes & Xception	99.9\%
LDDTA (custom CNN)	97.9\% \\ \hline

\cite{rakib2024lightweight} & PlantVillage Dataset 
& Custom CNN model designed with 2 convolutional, 2 max-pooling layers followed by 3 dense layers with the Last layer acting as the output layer, and 1 dropout layer. 
& Yes 
& 98\% \\ \hline

\cite{sharma2024enhanced} & PlantVillage Dataset & Custom CNN (Depthwise Separable Conv + Dilated Conv + Attention) & Yes & 99.44\% \\ \hline

\cite{mustafa2023pepper} & PlantVillage Dataset 
& Custom CNN model designed with 5 convolutional layers, 2 max pool layers, a flatten layer, and a fully connected layer. 
& Yes 
& 99.99\% \\ \hline

\cite{singla2024detection} & PlantVillage Dataset 
& Custom CNN model with 4 Convolutional layers, 4 Max pooling layers, 1 flatten layer, 1 dense layer followed by a dropout layer, and a final dense layer 
& Yes 
& 99.39\% \\ \hline

\cite{joseph2024real} & Dataset collected from internet sources 
& Eight fine-tuned deep learning models (Xception, MobileNet, MobileNetV2, Inception V3, InceptionResNetV2, ResNet 50, ResNet 101, VGG16) and a newly proposed custom MRW-CNN model (Maize, Rice, Wheat-CNN Model) is used for plant disease detection 
& Yes 
& Proposed MRW-CNN model: Maize: 97.04\%, Rice: 97.06\%, Wheat: 98.08\% \\ \hline

\end{tabular}
\end{table}

\par 
Research in the field of plant disease detection has shown a significant evolution in the use of CNN model architectures. Initially, traditional single CNN models were predominantly used for disease detection tasks. These models were effective in extracting features from images and providing accurate predictions. However, recent trends have shown a noticeable shift towards more advanced architectures, including ensemble, hybrid, and custom CNN models. Ensemble CNN models combine the strengths of multiple individual CNNs, often trained on different subsets of data or using different architectures. This approach leverages the diversity of predictions to improve overall performance, enhancing both accuracy and robustness in disease detection \cite{amin2022end,tugrul2022convolutional,shovon2023plantdet}. Hybrid CNN models integrate components from various neural network architectures or incorporate additional domain-specific knowledge, optimizing the models for specific tasks and conditions. These hybrid approaches benefit from the combined advantages of different methodologies, leading to more precise disease identification \cite{chen2023apple,kumar2023systematic,chug2023novel,adem2023sugar,brown30plant}. Custom CNN architectures have also gained prominence, particularly for proprietary datasets that researchers collect locally. These datasets are tailored to specific regional and environmental conditions. Custom CNNs are designed to handle the unique characteristics of these datasets, incorporating specialized layers and mechanisms to extract subtle features indicative of diseases \cite{syed2022citrus,nagi2022deep,joshi2022ricebios,rakib2024lightweight,mustafa2023pepper,singla2024detection}. By focusing on locally collected data, researchers can develop highly relevant and effective models for their specific agricultural contexts. However, the choice of CNN model architecture depends significantly on the diversity and quality of the dataset. Diverse datasets that include a wide range of plant species, disease symptoms, and environmental conditions help create more robust models capable of generalizing well to new, unseen data \cite{seetharaman2022leaf,syed2022citrus}. High-quality datasets, with accurate and comprehensive annotations, further enhance the performance of CNN models. As a result, selecting the appropriate model architecture is not only about the complexity of the model but also about ensuring it is well-suited to the dataset's characteristics.

\par
Dataset acquisition conditions further influence model usage patterns. Laboratory acquired datasets, characterized by uniform backgrounds and controlled lighting, often serve as testbeds for benchmarking high capacity architectures. These settings reduce environmental noise, allowing models to achieve high accuracy with limited structural adaptation. Conversely, field acquired datasets, which include background clutter, occlusions, and illumination variability, tend to exhibit more modest reported performance. In these cases, researchers frequently rely on data augmentation strategies to compensate for variability rather than fundamentally redesigning model architectures.

An additional observation concerns the relationship between dataset scale and model complexity. While larger datasets theoretically support deeper and more expressive architectures, many studies apply complex models even to relatively small datasets, often relying heavily on transfer learning. This practice may inflate reported performance while masking potential overfitting risks. At the same time, some large-scale datasets continue to be evaluated primarily with standard CNN backbones, suggesting that architectural innovation is not consistently aligned with dataset magnitude.

Importantly, the connection between structural imbalance and model choice remains limited in the existing literature. Although class imbalance is pervasive across plant disease datasets, relatively few studies explicitly adopt imbalance aware training strategies such as class-weighted loss functions or focal loss. Instead, performance improvements are frequently attributed to architectural refinement or augmentation techniques. This indicates a gap between dataset structural analysis and model adaptation practices.

Overall, model usage patterns across plant disease datasets reflect a strong dependence on dataset type, acquisition setting, and scale. Single-plant datasets tend to encourage fine-tuned standard architectures, while multi-plant datasets promote architectural diversification. Laboratory datasets frequently enable high reported performance under controlled conditions, whereas field datasets expose generalization limitations. These observations reinforce the central premise of this review: deep learning model performance cannot be evaluated independently of the datasets on which they are trained and tested.

\subsection{Model Complexity vs Dataset Characteristics}
The connection between model complexity and dataset characteristics in plant disease detection research is not always straightforward. While deeper and more sophisticated architectures are often assumed to deliver better performance, their actual effectiveness depends heavily on the nature of the dataset on which they are trained.

A noticeable trend in the literature is the frequent use of high capacity architectures such as AlexNet, VGGNet, ResNet, DenseNet, and ensemble models even when the underlying datasets are relatively small. In many cases, transfer learning is employed to compensate for limited data availability. While this approach can stabilize training and improve convergence, it may also mask deeper issues such as overfitting, especially when datasets exhibit strong class imbalance. Applying highly complex models to structurally limited datasets does not necessarily guarantee meaningful generalization.

Interestingly, the opposite pattern is also observed. Some large and widely used benchmark datasets are evaluated using relatively standard convolutional architectures, yet they report extremely high accuracy. This is particularly common in laboratory-based datasets where backgrounds are uniform and environmental noise is minimal. In such controlled settings, the classification task becomes comparatively simpler, allowing even conventional CNN models to achieve near-saturated performance. In these cases, the dataset characteristics rather than architectural innovation often drive the reported results.
\par
\begin{figure}[ht]
\centering

\begin{tikzpicture}

\definecolor{high}{RGB}{33,113,181}
\definecolor{medium}{RGB}{107,174,214}
\definecolor{low}{RGB}{198,219,239}

\def\cell{0.8}

\node[rotate=45] at (1*\cell,5.5*\cell) {Lab-Based};
\node[rotate=45] at (2*\cell,5.5*\cell) {Field-Based};
\node[rotate=45] at (3*\cell,5.5*\cell) {Small Dataset};
\node[rotate=45] at (4*\cell,5.5*\cell) {Large Dataset};
\node[rotate=45] at (5*\cell,5.5*\cell) {Imbalanced};

\node[left] at (0,4*\cell) {Single CNN};
\node[left] at (0,3*\cell) {Ensemble};
\node[left] at (0,2*\cell) {Hybrid};
\node[left] at (0,1*\cell) {Custom CNN};

\fill[high] (0.5*\cell,3.5*\cell) rectangle (1.5*\cell,4.5*\cell);
\fill[medium] (1.5*\cell,3.5*\cell) rectangle (2.5*\cell,4.5*\cell);
\fill[high] (2.5*\cell,3.5*\cell) rectangle (3.5*\cell,4.5*\cell);
\fill[medium] (3.5*\cell,3.5*\cell) rectangle (4.5*\cell,4.5*\cell);
\fill[medium] (4.5*\cell,3.5*\cell) rectangle (5.5*\cell,4.5*\cell);

\fill[medium] (0.5*\cell,2.5*\cell) rectangle (1.5*\cell,3.5*\cell);
\fill[medium] (1.5*\cell,2.5*\cell) rectangle (2.5*\cell,3.5*\cell);
\fill[low] (2.5*\cell,2.5*\cell) rectangle (3.5*\cell,3.5*\cell);
\fill[high] (3.5*\cell,2.5*\cell) rectangle (4.5*\cell,3.5*\cell);
\fill[low] (4.5*\cell,2.5*\cell) rectangle (5.5*\cell,3.5*\cell);

\fill[low] (0.5*\cell,1.5*\cell) rectangle (1.5*\cell,2.5*\cell);
\fill[medium] (1.5*\cell,1.5*\cell) rectangle (2.5*\cell,2.5*\cell);
\fill[low] (2.5*\cell,1.5*\cell) rectangle (3.5*\cell,2.5*\cell);
\fill[medium] (3.5*\cell,1.5*\cell) rectangle (4.5*\cell,2.5*\cell);
\fill[low] (4.5*\cell,1.5*\cell) rectangle (5.5*\cell,2.5*\cell);

\fill[medium] (0.5*\cell,0.5*\cell) rectangle (1.5*\cell,1.5*\cell);
\fill[medium] (1.5*\cell,0.5*\cell) rectangle (2.5*\cell,1.5*\cell);
\fill[high] (2.5*\cell,0.5*\cell) rectangle (3.5*\cell,1.5*\cell);
\fill[low] (3.5*\cell,0.5*\cell) rectangle (4.5*\cell,1.5*\cell);
\fill[medium] (4.5*\cell,0.5*\cell) rectangle (5.5*\cell,1.5*\cell);

\end{tikzpicture}
\vspace*{-3mm}
\caption{Heatmap-style summary of model usage across different dataset characteristics in plant disease detection research. Darker shades indicate higher frequency of usage.}
\label{fig:model_dataset_heatmap}
\end{figure}
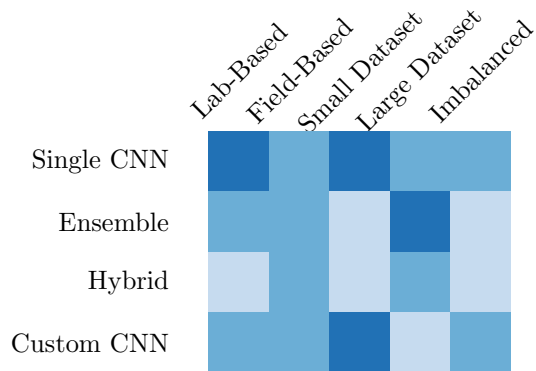

 
An additional challenge arises from the lack of standardized benchmarks for model comparison. Architectural diversity across studies ranging from shallow custom CNNs to deep ensembles and hybrid frameworks combined with differences in dataset splits, preprocessing strategies, augmentation pipelines, and evaluation metrics, makes direct comparison difficult. The absence of unified benchmarking protocols limits the ability to assess whether performance gains stem from architectural innovation, dataset characteristics, or experimental configuration. Consequently, reported improvements often reflect dataset-specific optimization rather than generalizable advances. Figure~\ref{fig:model_dataset_heatmap} summarizes the general patterns observed in the literature by illustrating the relative frequency of different model types across varying dataset characteristics.

Environmental variability further complicates the relationship between dataset structure and model complexity. Field-collected datasets introduce natural backgrounds, occlusions, lighting variation, and heterogeneous symptom expression. These factors increase task difficulty and more closely resemble real agricultural conditions. Although deeper or ensemble architectures may improve robustness in such settings, performance gains are generally more modest compared to laboratory benchmarks. This suggests that environmental realism imposes constraints that architectural depth alone cannot fully overcome.

Class imbalance is another structural factor that influences how model complexity translates into performance. Complex networks have the capacity to fit dominant classes extremely well, but this can come at the expense of minority disease categories. Without explicit imbalance aware training strategies such as class weighting or resampling, greater architectural depth may unintentionally reinforce distribution bias rather than mitigate it. In this context, dataset composition often defines the practical limits of model effectiveness.

Data augmentation is commonly used to address limited dataset size. Techniques such as rotation, flipping, and color transformations increase apparent diversity and can improve generalization. However, augmentation does not fundamentally correct skewed class distributions or inter-crop dominance in multi-plant datasets. As a result, high reported accuracy may reflect enhanced pattern recognition within existing structural constraints rather than true robustness.

Overall, these observations suggest that model complexity in plant disease detection research is often influenced more by dataset convenience than by dataset suitability. Architectural sophistication must be interpreted in light of dataset scale, structural balance, and environmental realism. Without careful alignment between model design and dataset characteristics and without standardized evaluation frameworks, advances in architecture may remain confined to experimental settings rather than translating into reliable, field-ready disease detection systems.
\vspace*{-3mm}

\par
These findings reinforce the dataset-centric premise of this review: advances in plant disease detection cannot rely solely on deeper architectures or ensemble strategies. Instead, future progress depends on the development of structurally balanced, environmentally realistic datasets and standardized benchmarking frameworks that enable fair and reproducible comparison. Importantly, dataset structure is not limited to class distribution and scale. Environmental variability such as temperature, humidity, and leaf wetness plays a fundamental role in disease progression and visual manifestation. The integration of such contextual information represents a necessary step toward more reliable and field-ready intelligent agricultural systems. Accordingly, the next section examines the role of environmental parameters in plant disease detection and explores how multimodal integration can bridge the gap between laboratory performance and real-world deployment.

\section{Environmental Variability and Multimodal Consideration}
\label{sec:Environmental Variability and Multimodal Consideration}
\par
A critical connection between dataset acquisition settings and model learning behavior emerges when structural imbalance and environmental variability are examined jointly. Field-acquired datasets, while more representative of real agricultural environments, typically exhibit higher normalized imbalance ratios (NIR) due to uneven crop representation, inconsistent disease incidence rates, and naturally occurring variability in symptom expression. This increased structural skewness introduces greater learning complexity, as models must differentiate between heterogeneous backgrounds, fluctuating lighting conditions, and minority disease classes under realistic noise. In contrast, laboratory-generated datasets often demonstrate lower variability and more controlled class distributions, resulting in reduced structural complexity and comparatively lower NIR values. While such conditions facilitate optimization and yield high reported classification accuracies, they may inadvertently inflate performance metrics that do not translate to field conditions. Moreover, pronounced dataset skewness particularly in multi-plant collections can suppress minority disease learning, causing models to bias predictions toward dominant classes. Consequently, environmental variability, structural imbalance, and reported model performance are not independent factors but are intrinsically interconnected in shaping real-world generalization capability.
\par
Plant diseases do not emerge in isolation. Their onset, severity, and progression are strongly influenced by environmental conditions such as temperature, humidity, rainfall, soil moisture, wind patterns, and leaf wetness duration \cite{stevens1960cultural}. These factors directly affect pathogen growth cycles, host susceptibility, and the rate of disease transmission across crops. For example, prolonged leaf wetness combined with high humidity creates favorable conditions for fungal infections, while temperature fluctuations can accelerate or suppress pathogen development. Similarly, excessive soil moisture may predispose crops to root-based diseases. As climate variability increases across agricultural regions, these environmental dependencies become even more pronounced.
\par
\begin{figure}[H]
    \centering
    \includegraphics[width=0.4\textwidth]{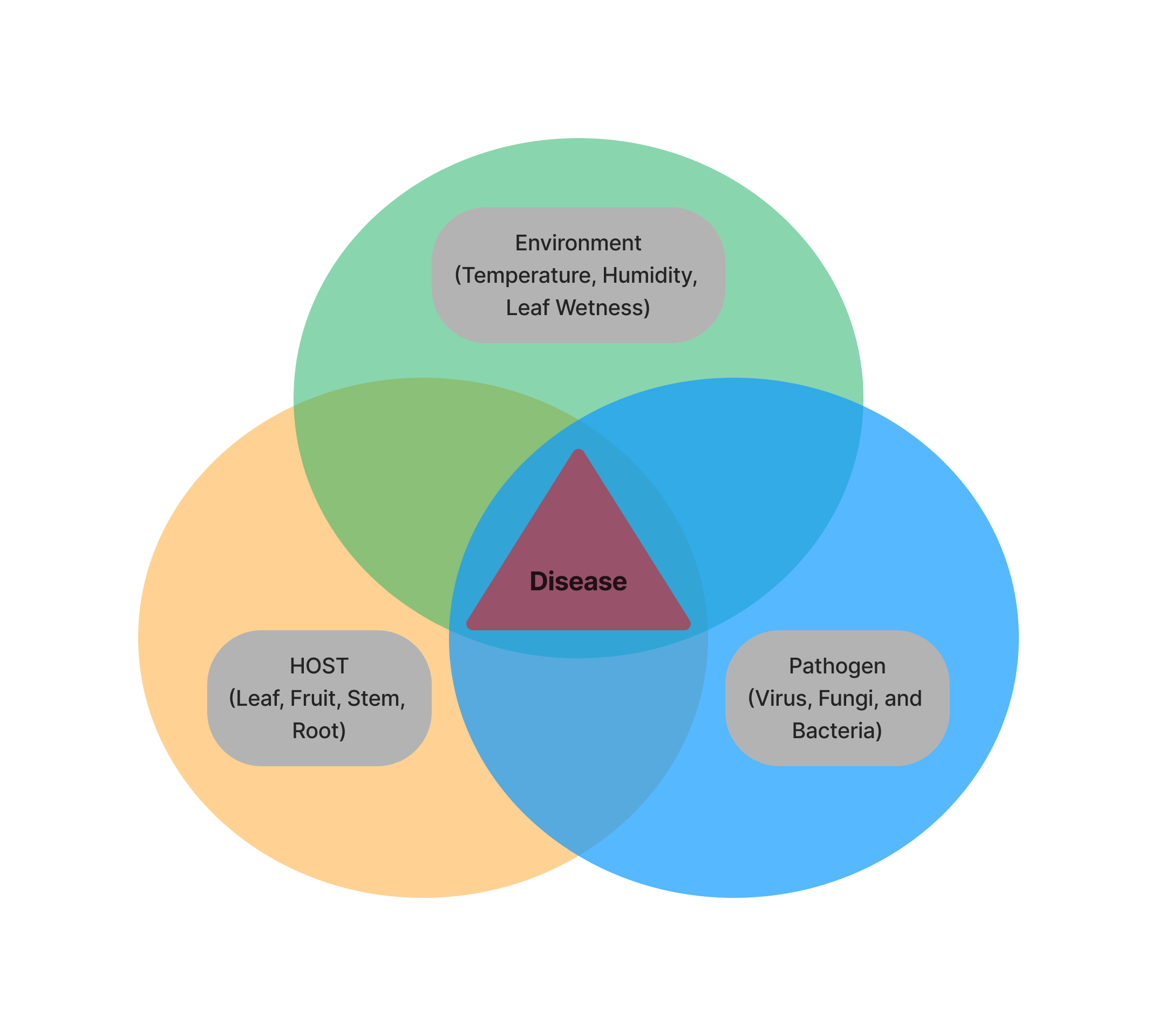}
    \vspace*{-10mm}
    \caption{Disease Triangle.}
    \label{fig: Disease Triangle.}
\end{figure}

This section will explore the significant impact of environmental variability on plant diseases, detailing how factors such as temperature changes, leaf wetness duration, humidity level, and soil moisture influence the occurrence and intensity of these diseases. By examining how changing environmental conditions can create favorable conditions for pathogens, we can better understand their prevalence and severity in various plant species. Environmental conditions play an important role in the development of a crop \cite{roussin2024plant}. Factors like temperature changes, soil moisture level, relative humidity, precipitation rate, wind speed, sun radiation, and leaf wetness duration (LWD) have been demonstrated to have significant effects on the development of plant diseases in several academic studies \cite{colhoun1973effects,lee2023deep}. Understanding these environmental conditions helps predict disease outbreaks and implement timely interventions. The primary cause of illnesses in plants is pathogens (Disease-Causing Agents). A pathogen is a parasite that causes disease in crops. The main types of pathogens are viruses, fungi, and bacteria. The Disease Triangle or Illness triangle, as shown in Figure~\ref{fig: Disease Triangle.}, is a conceptual model that the literature regards as a core tenet of the variables involved in disease causation. According to Stevens \cite{stevens1960cultural}, when an appropriate host organism and a disease-causing agent, or pathogen, come together in an environment that is conducive to the development of a disease, a disease is caused. A disease (shaded region) will develop if a vulnerable host plant is intimately associated with a virulent plant pathogen under favorable environmental circumstances when all three of these factors are present.
\vspace*{-5mm}

\par
The disease triangle illustrates three critical conditions necessary for the development of diseases in plants. First, there must be a vulnerable host plant. The host plant provides the necessary nutrients for the pathogen to thrive. The second condition requires the presence of an active pathogen. Without a pathogen, the process of disease cannot begin. Varieties of plants are often targeted by specific bacteria, fungi, or viruses, emphasizing the importance of understanding each pathogen's host range. Just a few viruses can infect each variety of plant. The third and the important condition is the existence of a favorable environment that supports the pathogen’s survival and proliferation, which enables the plant to become ill. Effective disease management relies on early detection of environment changes that create favorable conditions for pathogens. This is the main idea behind disease forecasting using real-time plant data analysis. Sensing real-time data regarding crops is necessary for the detection of conditions like crop health, weather conditions, disease, the severity of infection, area of infection, etc. The combination of environmental data and image-based approaches holds immense promise for early plant disease detection. By integrating these environmental parameters, we can improve the accuracy and reliability of predictive models for plant health. For example, temperature affects pathogen growth, host susceptibility, and disease development, like Apple scab requires specific temperature ranges for spore germination and infection. Similarly, high humidity favors fungal spore production, dispersal, and infection. For example, Powdery mildew thrives in humid conditions. Leaf wetness is critical for spore germination and infection. For example, Apple scab ascospores require specific minimum durations of leaf wetness for successful infection. Shorter or longer wetness periods affect disease severity. These environmental factors play a crucial role in the onset and progression of diseases in plants, affecting conditions that either encourage or hinder pathogen development. Overall, environmental factors such as temperature, soil moisture, relative humidity, precipitation rate, wind speed, sunlight, and leaf wetness duration have proven to play a substantial role in the onset and progression of plant diseases \cite{lian2021effects,cao2024pathogenic,papp2020field}. The following tables~\ref{table:apple-temperature}--\ref{table:apple-leaf-wetness} show different types of environmental conditions responsible for different apple diseases.

\begin{table}[ht]
\centering
\footnotesize
\caption{Apple Diseases Due to Temperature \cite{lian2021effects}.}
\vspace*{-3mm}
\label{table:apple-temperature}
\begin{tabular}{p{4.0cm}p{3.0cm}p{6.0cm}}
\hline
\textbf{Disease Type} & \textbf{Affected Plant Part/Appearance} & \textbf{Favorable Environmental Conditions} \\ \hline
Apple Scab & Leaf and Fruit & Temperature on average: 20°C \\ \hline
Apple Powdery Mildew & Leaf and Fruit & Temperature ranging from 10°-25° C \\ \hline
Apple Marssonina Leaf Blotch & Leaf & High rainfall and temperatures between 20° and 22° C \\ \hline
Apple Alternaria Leaf Spot & Leaf & Temperature ranging from 25°-30° C \\ \hline
Apple Black Rot & Leaf and Fruit & 20°-22° C Temperature range and Moisture \\ \hline
Apple Sooty Blotch & Fruit & Temperature range 18°-27° C \\ \hline
Apple Seedling Blight & Root & Below Temperature of 24° C \\ \hline
\end{tabular}
\end{table}

\begin{table}[ht]
\centering
\footnotesize
\caption{Apple Diseases Due to Humidity \cite{cao2024pathogenic}.}
\vspace*{-3mm}
\label{table:apple-humidity}
\begin{tabular}{p{4.0cm}p{3.0cm}p{6.0cm}}
\hline
\textbf{Disease Type} & \textbf{Affected Plant Part/Appearance} & \textbf{Favorable Environmental Conditions} \\ \hline
Apple Powdery Mildew & Leaf and Fruit & Humidity greater than 70\% \\ \hline
Apple Collar Rot & Root & Over-watering and Moisture \\ \hline
Apple Seedling Blight & Root & High Humidity \\ \hline
Apple Core Rot & Fruit & High humidity and warm temperature \\ \hline
Apple Brown Rot & Fruit & Wet humid conditions \\ \hline
\end{tabular}
\end{table}

\begin{table}[ht]
\centering
\footnotesize
\caption{Apple Diseases Due to Leaf Wetness Duration \cite{papp2020field}.}
\vspace*{-3mm}
\label{table:apple-leaf-wetness}
\begin{tabular}{p{4.0cm}p{3.0cm}p{6.0cm}}
\hline
\textbf{Disease Type} & \textbf{Affected Plant Part/Appearance} & \textbf{Favorable Environmental Conditions} \\ \hline
Apple Scab & Leaf & Light infection: 9 hours at 18°C \\ \hline
Apple Alternaria Leaf Spot & Leaf & More than 6 hours of LWD \\ \hline
\end{tabular}
\end{table}
\par

\par
Despite this well established biological foundation, most deep learning based plant disease detection frameworks rely exclusively on visual information extracted from leaf images. While image-based models capture symptom appearance, they often overlook the contextual signals that influence disease manifestation. As a result, predictions may remain reactive identifying visible symptoms rather than proactive in anticipating disease risk under evolving environmental conditions. Image-based models trained solely on lab data do not capture the variability and complexity of real-world agricultural settings. Integrating environmental data (such as temperature, humidity, and leaf wetness) with image data can enhance disease prediction accuracy. In the agricultural field, plants are subjected to a dynamic range of environmental conditions that can alter disease expression. Models trained solely on static image data from controlled environments may fail to generalize to these varying conditions, leading to decreased performance in practical applications. Combining environmental data with image data represents a transformative approach to plant disease detection. While image data provides visual symptoms of disease, such as spots, discoloration, or wilting, it often lacks the context of underlying environmental conditions that may have led to these symptoms. By integrating environmental parameters, deep learning models can be trained to recognize not only the visual cues of disease but also the environmental triggers that precede visible symptoms. This dual approach enhances the precision of disease prediction, enabling earlier and more accurate interventions.

A review of publicly available plant disease datasets indicates that only a very small fraction provide synchronized environmental metadata alongside image samples. Most benchmark datasets are purely image based, with no associated temperature, humidity, or leaf wetness information. This structural absence limits the ability to evaluate context-aware disease prediction systems under standardized conditions.

\subsection{Limitations of Image-Only Detection Systems}
Although image based deep learning models have demonstrated high classification accuracy particularly on laboratory datasets, their reliance on visual features introduces several limitations:
\begin{itemize}
    \setlength{\itemsep}{2pt}
    \setlength{\parskip}{0pt}
    \setlength{\parsep}{0pt}
    \item Visual symptoms often appear after disease progression has begun.

    \item Environmental stress may mimic disease symptoms, leading to misclassification.

    \item Field variability (lighting, occlusion, background clutter) reduces reliability.

    \item Disease risk cannot be inferred solely from current leaf appearance.
\end{itemize}

In addition, image-only systems may struggle in early-stage disease detection, where visible symptoms are subtle or indistinguishable from natural variation.

These limitations suggest that visual classification alone may be insufficient for robust, field-ready plant disease management systems.

\subsection{Multimodal Learning for Context-Aware Disease Prediction}
Multimodal learning integrates heterogeneous data sources into a unified predictive framework. In the context of plant disease detection, this typically involves combining:
\begin{itemize}
    \setlength{\itemsep}{2pt}
    \setlength{\parskip}{0pt}
    \setlength{\parsep}{0pt}
    \item Leaf image features (CNN-based visual representations)

    \item Environmental sensor data (temperature, humidity, soil moisture, leaf wetness)

    \item Temporal trends (seasonal or time-series information)
\end{itemize}

The integration of environmental data enables models to incorporate causal context alongside visual symptoms. For instance, a model detecting early discoloration in leaves may weigh the prediction differently depending on whether recent humidity levels support fungal growth.

\subsection{Challenges in Environmental Integration}
While multimodal systems offer strong potential, several practical challenges remain:
\begin{itemize}
    \setlength{\itemsep}{2pt}
    \setlength{\parskip}{0pt}
    \setlength{\parsep}{0pt}
    \item Lack of synchronized environmental and image datasets

    \item Inconsistent sensor quality across regions

    \item Missing or incomplete environmental records

    \item Increased model complexity and deployment cost

    \item Limited standardized multimodal benchmarks
\end{itemize}

\par
The challenges discussed above are not isolated technical issues but reflect broader structural limitations in current plant disease datasets. To clarify these limitations from a data-centric perspective, Table~\ref{tab:env_challenges} summarizes the key barriers that hinder effective environmental integration in multimodal plant disease detection systems.

\begin{table}[H]
\centering
\footnotesize
\caption{Structural challenges limiting environmental integration in plant disease datasets.}
\vspace*{-3mm}
\label{tab:env_challenges}
\begin{tabular}{p{5cm}p{7cm}}
\hline
\textbf{Challenge} & \textbf{Implication for Model Development and Deployment} \\
\hline
Lack of synchronized environmental metadata &
Prevents context-aware learning and limits multimodal benchmarking consistency. \\ \hline

Sensor heterogeneity across studies &
Reduces reproducibility and complicates cross-dataset comparison. \\ \hline

Environmental variability across regions &
Introduces domain shift and weakens model generalization in field deployment. \\ \hline

Absence of standardized multimodal benchmarks &
Hinders fair architectural comparison and systematic performance evaluation. \\ \hline

Limited public availability of multimodal datasets &
Restricts collaborative research and slows development of field-ready systems. \\ \hline
\end{tabular}
\end{table}

\par
The analysis presented throughout this review underscores that architectural sophistication alone cannot overcome structural dataset limitations or environmental variability. While deep learning has advanced automated plant disease detection considerably, future progress depends on the development of structurally balanced, environmentally enriched datasets and standardized multimodal benchmarking frameworks. Integrating contextual environmental parameters with visual analysis represents a critical step toward reliable, scalable, and real-world-ready plant disease detection systems capable of supporting sustainable precision agriculture.

\section{Challenges and Open issues}
\label{sec:Challenges and Open issues}
Plant disease detection research has witnessed substantial progress over the past decade, particularly with the adoption of deep learning and computer vision techniques. However, despite high reported accuracies under controlled experimental conditions, several fundamental challenges continue to hinder reliable real-world deployment. These challenges extend beyond model architecture design and are deeply rooted in dataset structure, environmental variability, evaluation practices, and integration limitations. Addressing these open issues is essential for transitioning from laboratory success to scalable, field-ready intelligent agricultural systems, as discussed in Table~\ref{tab:open_challenges_long}.

{\small
\begin{longtable}{p{4.5cm}p{12.0cm}}
\caption{Open challenges and emerging issues in real-world deep learning--based plant disease detection.}
\vspace*{-3mm}
\label{tab:open_challenges_long} \\

\hline
\textbf{Open Challenge} & \textbf{Detailed Description and Implications} \\
\hline
\endfirsthead

\hline
\textbf{Open Challenge} & \textbf{Detailed Description and Implications} \\
\hline
\endhead

\hline
\endfoot

\hline
\endlastfoot

Scarcity of Large-Scale Real-Field Datasets &
Although numerous plant disease datasets have been introduced, a substantial proportion remains laboratory-generated, characterized by uniform backgrounds, controlled illumination, and limited environmental noise. Real-field datasets are often smaller in size, geographically constrained, crop-specific, and frequently proprietary. This structural dependency on controlled datasets inflates experimental performance while limiting ecological validity and real-world generalization. The development of large-scale, geographically diverse, publicly accessible field datasets remains a pressing research priority. \\ \hline

Multi-Level Class Imbalance and Data Skewness &
Class imbalance persists at multiple levels, including intra-class imbalance within disease categories, inter-crop imbalance across plant species, and cross-dataset imbalance due to inconsistent dataset design. Severe skewness biases model learning toward dominant classes, suppresses minority disease detection, and produces misleading overall accuracy metrics. While augmentation and resampling strategies are commonly applied, they do not fundamentally resolve structural imbalance inherent in dataset construction. More principled imbalance-aware learning and evaluation strategies remain underexplored in agricultural AI research. \\ \hline

Lack of Standardized Benchmarking Protocols &
The absence of unified benchmarking frameworks complicates fair comparison across studies. Variability in train–test splits, preprocessing pipelines, augmentation policies, and performance metrics limits reproducibility. Additionally, architectural diversity across single, ensemble, hybrid, and custom models further hinders standardized evaluation. Without common validation protocols and cross-dataset testing strategies, reported improvements may reflect experimental configuration rather than genuine methodological advancement. \\ \hline

Limited Cross-Dataset Generalization Studies &
Most existing research evaluates models on the same dataset used for training. Cross-dataset validation—training on one dataset and testing on another is rarely performed. This limits understanding of domain shift effects, acquisition variability, crop morphological differences, and environmental transferability. Robust real-world deployment requires models that generalize across datasets, regions, and acquisition conditions, yet systematic cross-dataset evaluation remains largely unexplored. \\ \hline

Environmental Context Integration Constraints &
Environmental factors such as temperature, humidity, leaf wetness, and soil conditions significantly influence disease progression. However, most detection frameworks rely exclusively on visual inputs. Multimodal datasets combining image data with synchronized environmental measurements remain scarce. Key challenges include temporal alignment between sensor data and image capture, sensor calibration variability, missing environmental readings, and increased fusion complexity. Establishing standardized multimodal benchmarks is critical for advancing environmentally aware disease prediction systems. \\ \hline

Annotation Quality and Label Noise &
Accurate annotation of plant disease images requires domain expertise. Field images may contain mixed symptoms, early-stage infections with subtle visual cues, or co-existing diseases on the same leaf. Inconsistent labeling protocols introduce annotation noise that negatively impacts model learning, generalization, and reproducibility. Establishing standardized annotation guidelines and expert-verified labeling pipelines is essential for improving dataset reliability. \\ \hline

Over-Reliance on Overall Accuracy Metrics &
Many studies emphasize overall accuracy as the primary evaluation metric. In imbalanced datasets, high accuracy may mask poor minority-class detection performance. Critical metrics such as class-wise recall, F1-score, balanced accuracy, precision--recall curves, and confusion matrices are often underreported. In agricultural deployment scenarios, failure to detect rare but severe diseases may lead to significant economic consequences, necessitating more robust and imbalance-aware evaluation practices. \\ \hline

Computational and Deployment Constraints &
High-performing ensemble and hybrid models often require substantial computational resources, limiting deployment on edge devices commonly used in agricultural settings. Real-world farms may experience limited connectivity, power constraints, and hardware limitations. Designing lightweight, energy efficient, and latencyaware models suitable for field deployment remains a major engineering and research challenge. \\ \hline

Geographic and Climatic Bias &
Many plant disease datasets are region-specific, reflecting localized agro-climatic conditions. Disease manifestation patterns vary across geographic regions due to differences in climate, soil composition, and cultivation practices. Models trained on data from one region may fail to generalize to another due to climatic and phenological variability. Creating globally representative and geographically diverse datasets remains logistically and scientifically demanding. \\ \hline

Toward Next-Generation Multimodal, Balanced, Field-Ready Systems &
Addressing the above challenges requires a paradigm shift from accuracy driven experimentation toward deployment oriented system design. Future research must prioritize large-scale real-field dataset construction, balanced multi/single crop disease representation, cross-dataset validation protocols, standardized benchmarking frameworks, multimodal integration of environmental and visual data, lightweight deployable architectures, and transparent reporting standards. Only through coordinated dataset engineering and methodological rigor can plant disease detection systems transition from experimental prototypes to scalable, field-ready precision agriculture solutions. \\

\end{longtable}
}

\par In summary, the future of plant disease detection lies not solely in developing deeper architectures, but in addressing foundational data-centric challenges. Structural imbalance, environmental variability, benchmarking inconsistency, and limited multimodal integration collectively constrain progress toward field-ready systems. Bridging these gaps requires a paradigm shift from accuracy-driven experimentation to deployment-oriented design, grounded in realistic datasets and robust evaluation practices.

\section{Discussion}
\label{sec:Discussion}
This review was motivated by a fundamental question: Why do plant disease detection models that achieve high accuracy in controlled settings often struggle under real-world agricultural conditions? The comprehensive analysis conducted throughout this study suggests that the answer lies less in architectural sophistication and more in dataset structure, imbalance, and environmental realism.

A central observation emerging from this review is the structural mismatch between laboratory datasets and real-field agricultural environments. Laboratory datasets while carefully curated and often publicly accessible typically exhibit uniform backgrounds, controlled lighting, and simplified class distributions. Such structural consistency reduces variability and frequently results in lower normalized imbalance ratios (NIR), enabling models to learn discriminative features more easily. However, this apparent stability can lead to inflated accuracy and over-optimistic generalization claims. In contrast, field-acquired datasets capture environmental complexity, occlusions, lighting variability, and natural class prevalence patterns. These datasets often exhibit higher skewness, minority-class suppression, and inter-crop imbalance, which significantly increases learning difficulty and exposes limitations in model robustness.

The structural analysis of single-plant and multi-plant datasets further reveals that imbalance is not merely a numerical artifact but a defining characteristic of agricultural data. In single-plant datasets, skewness frequently arises from disproportionate representation of dominant disease categories. In multi-plant datasets, imbalance operates at multiple levels: within individual crops (intra-crop imbalance) and across crops (inter-crop imbalance). This layered imbalance structure directly affects feature learning, biasing models toward majority disease classes while under-representing rare but agriculturally critical conditions.

The dataset model mapping analysis demonstrates that researchers often compensate for imbalance and limited data diversity through architectural complexity, ensemble strategies, hybrid models, and aggressive data augmentation. While such approaches may improve performance metrics, they do not fundamentally resolve structural data limitations. Moreover, the lack of standardized benchmarks combined with architectural diversity and inconsistent dataset splits hampers fair model comparison and obscures the true impact of dataset characteristics on performance.

Environmental variability adds a further dimension to this challenge. Disease manifestation is inherently contextual, influenced by temperature, humidity, leaf wetness duration, and seasonal patterns. Image-only systems, even when trained on balanced datasets, remain reactive; they detect visible symptoms but rarely anticipate disease risk. Multimodal learning frameworks that integrate environmental signals represent a promising direction; however, the scarcity of synchronized image environment datasets remains a major barrier. Without structurally balanced and environmentally enriched benchmarks, multimodal architectures cannot be rigorously evaluated.

Taken together, the findings of this review suggest that future progress in plant disease detection depends on a paradigm shift. Rather than prioritizing incremental architectural improvements, research must focus on:

\begin{itemize}
    \setlength{\itemsep}{2pt}
    \setlength{\parskip}{0pt}
    \setlength{\parsep}{0pt}
    \item Developing structurally balanced, large-scale field datasets

    \item Reporting standardized imbalance metrics (e.g., NIR)

    \item Providing transparent dataset splits and benchmarking protocols

    \item Integrating environmental metadata into disease detection pipelines
\end{itemize}
The transition from laboratory prototypes to field-ready intelligent systems requires harmonizing dataset realism, structural balance, and multimodal integration. Architectural innovation alone is insufficient without parallel advances in data quality and representativeness.

\section{Conclusions}
\label{sec:Conclusion}
Computer Vision and Deep Learning has improved the plant disease prediction and detection in many crops which leads to improved production and quality of the produce. Yet, despite remarkable performance gains in controlled experiments, reliable deployment in real-world agricultural environments remains limited. This review demonstrates that the primary bottleneck is not the absence of sophisticated models but the structural limitations of the datasets on which they are trained.

By adopting a dataset-centric perspective, this study systematically organized plant disease datasets according to acquisition setting, plant diversity, accessibility, class composition, and imbalance severity. Through quantitative imbalance analysis using normalized imbalance ratio (NIR), we revealed pervasive skewness across both single-plant and multi-plant datasets, highlighting structural biases that directly influence model learning behavior. The comparative evaluation of laboratory and field datasets exposed critical gaps in environmental realism, dataset diversity, and public accessibility.

Furthermore, the dataset model mapping analysis showed that while single, ensemble, hybrid, and custom architectures are widely explored, their performance remains intrinsically tied to dataset characteristics. The absence of standardized benchmarks and the widespread reliance on augmentation strategies often obscure the true generalization capability of these systems.

Environmental variability introduces an additional layer of complexity. Disease emergence is governed by the interaction between host, pathogen, and environment, yet most existing detection systems rely exclusively on visual cues. Multimodal approaches integrating environmental parameters offer a promising pathway toward proactive and context-aware disease prediction. However, the limited availability of aligned multimodal datasets represents a major research gap.

In summary, advancing plant disease detection toward scalable, field-deployable precision agriculture systems requires:

\begin{itemize}
    \setlength{\itemsep}{2pt}
    \setlength{\parskip}{0pt}
    \setlength{\parsep}{0pt}
    \item Structurally balanced and environmentally representative datasets

    \item Transparent reporting of class distributions and imbalance metrics

    \item Standardized evaluation benchmarks across architectures

    \item Integration of multimodal sensing frameworks

    \item Publicly accessible, synchronized image–environment repositories
\end {itemize}

The future of intelligent plant disease detection lies not solely in deeper networks, but in richer, more representative data ecosystems. By aligning dataset design, model development, and environmental integration, next-generation systems can move beyond controlled experimental success toward reliable, real-world agricultural impact.

\section*{Funding}
The authors received no financial support for the research, authorship, and/or publication of this article.

\bibliographystyle{unsrtnat}
\bibliography{mybib}

\begin{thebibliography}{135}
\providecommand{\natexlab}[1]{#1}
\providecommand{\url}[1]{\texttt{#1}}
\expandafter\ifx\csname urlstyle\endcsname\relax
  \providecommand{\doi}[1]{doi: #1}\else
  \providecommand{\doi}{doi: \begingroup \urlstyle{rm}\Url}\fi

\bibitem[Food and of~the United~Nations(n.d.)]{fao_plant_production_protection}
Food and Agriculture~Organization of~the United~Nations.
\newblock Plant production and protection.
\newblock \url{https://www.fao.org/plant-production-protection/en}, n.d.
\newblock Accessed: 2026-01-27.

\bibitem[Nations(2022{\natexlab{a}})]{un2022worldpopulation}
United Nations.
\newblock World population to reach 8 billion on 15 november 2022.
\newblock \url{https://www.un.org/en/desa/world-population-reach-8-billion-15-november-2022}, 2022{\natexlab{a}}.
\newblock Accessed: 2026-01-27.

\bibitem[Nations(2022{\natexlab{b}})]{un2022wpp}
United Nations.
\newblock {UN Department of Economic and Social Affairs Population Division - World Population Prospects 2022}.
\newblock \url{https://population.un.org/wpp/}, 2022{\natexlab{b}}.
\newblock Accessed: 2026-01-27.

\bibitem[{United Nations}(2021)]{un_population_2021}
{United Nations}.
\newblock Un news - population, 2021.
\newblock URL \url{https://www.un.org/en/global-issues/population}.
\newblock Accessed: 2026-01-27.

\bibitem[{United Nations Population Fund}(2024)]{unfpa2024}
{United Nations Population Fund}.
\newblock {World Population Dashboard India}.
\newblock \url{https://www.unfpa.org/data/world-population/IN}, 2024.
\newblock Accessed: 2026-01-27.

\bibitem[Chouhan et~al.(2020)Chouhan, Singh, and Jain]{chouhan2020applications}
Siddharth~Singh Chouhan, Uday~Pratap Singh, and Sanjeev Jain.
\newblock Applications of computer vision in plant pathology: a survey.
\newblock \emph{Archives of computational methods in engineering}, 27\penalty0 (2):\penalty0 611--632, 2020.

\bibitem[Gajjar et~al.(2022)Gajjar, Gajjar, Thakor, Patel, and Ruparelia]{gajjar2022real}
Ruchi Gajjar, Nagendra Gajjar, Vaibhavkumar~Jigneshkumar Thakor, Nikhilkumar~Pareshbhai Patel, and Stavan Ruparelia.
\newblock Real-time detection and identification of plant leaf diseases using convolutional neural networks on an embedded platform.
\newblock \emph{The Visual Computer}, pages 1--16, 2022.

\bibitem[Mure{\c{s}}an et~al.(2020)Mure{\c{s}}an, Coroiu, and C{\u{a}}lin]{murecsan2020detecting}
HB~Mure{\c{s}}an, Adriana~Mihaela Coroiu, and AD~C{\u{a}}lin.
\newblock Detecting leaf plant diseases using deep learning: A review.
\newblock In \emph{2020 International Conference on Software, Telecommunications and Computer Networks (SoftCOM)}, pages 1--6. IEEE, 2020.

\bibitem[Yao et~al.(2024)Yao, Tran, Garg, and Sawyer]{yao2024deep}
Jianping Yao, Son~N Tran, Saurabh Garg, and Samantha Sawyer.
\newblock Deep learning for plant identification and disease classification from leaf images: multi-prediction approaches.
\newblock \emph{ACM Computing Surveys}, 56\penalty0 (6):\penalty0 1--37, 2024.

\bibitem[Ashok et~al.(2020)Ashok, Kishore, Rajesh, Suchitra, Sophia, and Pavithra]{ashok2020tomato}
Surampalli Ashok, Gemini Kishore, Velpula Rajesh, S~Suchitra, SG~Gino Sophia, and B~Pavithra.
\newblock Tomato leaf disease detection using deep learning techniques.
\newblock In \emph{2020 5th International Conference on Communication and Electronics Systems (ICCES)}, pages 979--983. IEEE, 2020.

\bibitem[Azlah et~al.(2019)Azlah, Chua, Rahmad, Abdullah, and Wan~Alwi]{azlah2019review}
Muhammad Azfar~Firdaus Azlah, Lee~Suan Chua, Fakhrul~Razan Rahmad, Farah~Izana Abdullah, and Sharifah~Rafidah Wan~Alwi.
\newblock Review on techniques for plant leaf classification and recognition.
\newblock \emph{Computers}, 8\penalty0 (4):\penalty0 77, 2019.

\bibitem[Li et~al.(2021)Li, Zhang, and Wang]{li2021plant}
Lili Li, Shujuan Zhang, and Bin Wang.
\newblock Plant disease detection and classification by deep learning—a review.
\newblock \emph{IEEE Access}, 9:\penalty0 56683--56698, 2021.

\bibitem[Kanda et~al.(2021)Kanda, Xia, and Sanusi]{kanda2021deep}
Paul~Shekonya Kanda, Kewen Xia, and Olanrewaju~Hazzan Sanusi.
\newblock A deep learning-based recognition technique for plant leaf classification.
\newblock \emph{IEEE Access}, 9:\penalty0 162590--162613, 2021.

\bibitem[Lee et~al.(2021)Lee, Go{\"e}au, Bonnet, and Joly]{lee2021conditional}
Sue~Han Lee, Herv{\'e} Go{\"e}au, Pierre Bonnet, and Alexis Joly.
\newblock Conditional multi-task learning for plant disease identification.
\newblock In \emph{2020 25th international conference on pattern recognition (ICPR)}, pages 3320--3327. IEEE, 2021.

\bibitem[Othman et~al.(2022)Othman, Damanhuri, Ali, Meng, and Abd~Samat]{othman2022plant}
Nor~Azlan Othman, Nor~Salwa Damanhuri, Nabilah~Md Ali, Belinda Chong~Chiew Meng, and Ahmad~Asri Abd~Samat.
\newblock Plant leaf classification using convolutional neural network.
\newblock In \emph{2022 8th International Conference on Control, Decision and Information Technologies (CoDIT)}, volume~1, pages 1043--1048. IEEE, 2022.

\bibitem[Yao et~al.(2023)Yao, Tran, Sawyer, and Garg]{yao2023machine}
Jianping Yao, Son~N Tran, Samantha Sawyer, and Saurabh Garg.
\newblock Machine learning for leaf disease classification: data, techniques and applications.
\newblock \emph{Artificial Intelligence Review}, 56\penalty0 (Suppl 3):\penalty0 3571--3616, 2023.

\bibitem[Kumar et~al.(2022)Kumar, Chug, Singh, and Singh]{kumar2022retracted}
Raj Kumar, Anuradha Chug, Amit~Prakash Singh, and Dinesh Singh.
\newblock [retracted] a systematic analysis of machine learning and deep learning based approaches for plant leaf disease classification: A review.
\newblock \emph{Journal of Sensors}, 2022\penalty0 (1):\penalty0 3287561, 2022.

\bibitem[Agarwal et~al.(2021)Agarwal, Chawla, and Tiwari]{agarwal2021plant}
Deeksha Agarwal, Meenu Chawla, and Namita Tiwari.
\newblock Plant leaf disease classification using deep learning: A survey.
\newblock In \emph{2021 Third International Conference on Inventive Research in Computing Applications (ICIRCA)}, pages 643--650. IEEE, 2021.

\bibitem[Seetharaman and Mahendran(2022)]{seetharaman2022leaf}
K~Seetharaman and T~Mahendran.
\newblock Leaf disease detection in banana plant using gabor extraction and region-based convolution neural network (rcnn).
\newblock \emph{Journal of The Institution of Engineers (India): Series A}, 103\penalty0 (2):\penalty0 501--507, 2022.

\bibitem[Syed-Ab-Rahman et~al.(2022)Syed-Ab-Rahman, Hesamian, and Prasad]{syed2022citrus}
Sharifah~Farhana Syed-Ab-Rahman, Mohammad~Hesam Hesamian, and Mukesh Prasad.
\newblock Citrus disease detection and classification using end-to-end anchor-based deep learning model.
\newblock \emph{Applied Intelligence}, 52\penalty0 (1):\penalty0 927--938, 2022.

\bibitem[Mohanty et~al.(2016)Mohanty, Hughes, and Salath{\'e}]{mohanty2016using}
Sharada~P Mohanty, David~P Hughes, and Marcel Salath{\'e}.
\newblock Using deep learning for image-based plant disease detection.
\newblock \emph{Frontiers in plant science}, 7:\penalty0 1419, 2016.

\bibitem[Fenu and Malloci(2022)]{fenu2022evaluating}
Gianni Fenu and Francesca~Maridina Malloci.
\newblock Evaluating impacts between laboratory and field-collected datasets for plant disease classification.
\newblock \emph{Agronomy}, 12\penalty0 (10):\penalty0 2359, 2022.

\bibitem[Hughes et~al.(2015)Hughes, Salath{\'e}, et~al.]{hughes2015open}
David Hughes, Marcel Salath{\'e}, et~al.
\newblock An open access repository of images on plant health to enable the development of mobile disease diagnostics.
\newblock \emph{arXiv preprint arXiv:1511.08060}, 2015.

\bibitem[Liu et~al.(2021)Liu, Min, Mei, Wang, and Jiang]{liu2021plant}
Xinda Liu, Weiqing Min, Shuhuan Mei, Lili Wang, and Shuqiang Jiang.
\newblock Plant disease recognition: A large-scale benchmark dataset and a visual region and loss reweighting approach.
\newblock \emph{IEEE Transactions on Image Processing}, 30:\penalty0 2003--2015, 2021.

\bibitem[Xu et~al.(2024)Xu, Park, Lee, Yang, and Yoon]{xu2024plant}
Mingle Xu, Ji-Eun Park, Jaehwan Lee, Jucheng Yang, and Sook Yoon.
\newblock Plant disease recognition datasets in the age of deep learning: challenges and opportunities.
\newblock \emph{Frontiers in Plant Science}, 15:\penalty0 1452551, 2024.

\bibitem[Stevens(1960)]{stevens1960cultural}
Russell~B Stevens.
\newblock Cultural practices in disease control.
\newblock \emph{Plant pathology}, pages 357--429, 1960.

\bibitem[Colhoun(1973)]{colhoun1973effects}
JC~Colhoun.
\newblock Effects of environmental factors on plant diseases.
\newblock \emph{Annu. Rev. Phytopathol.;(United States)}, 11, 1973.

\bibitem[Hunjan and Lore(2020)]{hunjan2020climate}
Mandeep~Singh Hunjan and Jagjeet~Singh Lore.
\newblock Climate change: Impact on plant pathogens, diseases, and their management.
\newblock \emph{Crop protection under changing climate}, pages 85--100, 2020.

\bibitem[Sunil et~al.(2023)Sunil, Jaidhar, and Patil]{sunil2023systematic}
CK~Sunil, CD~Jaidhar, and Nagamma Patil.
\newblock Systematic study on deep learning-based plant disease detection or classification.
\newblock \emph{Artificial Intelligence Review}, 56\penalty0 (12):\penalty0 14955--15052, 2023.

\bibitem[Lee and Yun(2023)]{lee2023deep}
Sangyeon Lee and Choa~Mun Yun.
\newblock A deep learning model for predicting risks of crop pests and diseases from sequential environmental data.
\newblock \emph{Plant Methods}, 19\penalty0 (1):\penalty0 145, 2023.

\bibitem[Barbedo(2016)]{barbedo2016review}
Jayme Garcia~Arnal Barbedo.
\newblock A review on the main challenges in automatic plant disease identification based on visible range images.
\newblock \emph{Biosystems engineering}, 144:\penalty0 52--60, 2016.

\bibitem[Patr{\'\i}cio and Rieder(2018)]{patricio2018computer}
Diego~In{\'a}cio Patr{\'\i}cio and Rafael Rieder.
\newblock Computer vision and artificial intelligence in precision agriculture for grain crops: A systematic review.
\newblock \emph{Computers and electronics in agriculture}, 153:\penalty0 69--81, 2018.

\bibitem[Samajpati and Degadwala(2016)]{samajpati2016hybrid}
Bhavini~J Samajpati and Sheshang~D Degadwala.
\newblock Hybrid approach for apple fruit diseases detection and classification using random forest classifier.
\newblock In \emph{2016 International conference on communication and signal processing (ICCSP)}, pages 1015--1019. IEEE, 2016.

\bibitem[Rozario et~al.(2016)Rozario, Rahman, and Uddin]{rozario2016segmentation}
Liton~Jude Rozario, Tanzila Rahman, and Mohammad~Shorif Uddin.
\newblock Segmentation of the region of defects in fruits and vegetables.
\newblock \emph{International Journal of Computer Science and Information Security}, 14\penalty0 (5):\penalty0 399, 2016.

\bibitem[Kumar et~al.(2020)Kumar, Prasad, Srilekha, Suman, Rao, and Krishna]{kumar2020leaf}
Sandeep Kumar, KMVV Prasad, A~Srilekha, T~Suman, B~Pranav Rao, and J~Naga~Vamshi Krishna.
\newblock Leaf disease detection and classification based on machine learning.
\newblock In \emph{2020 International Conference on Smart Technologies in Computing, Electrical and Electronics (ICSTCEE)}, pages 361--365. IEEE, 2020.

\bibitem[Bharate and Shirdhonkar(2020)]{bharate2020classification}
Anil~A Bharate and MS~Shirdhonkar.
\newblock Classification of grape leaves using knn and svm classifiers.
\newblock In \emph{2020 Fourth International Conference on Computing Methodologies and Communication (ICCMC)}, pages 745--749. IEEE, 2020.

\bibitem[Dang-Ngoc et~al.(2021)Dang-Ngoc, Cao, and Dang-Nguyen]{dang2021citrus}
Hanh Dang-Ngoc, Trang~NM Cao, and Chau Dang-Nguyen.
\newblock Citrus leaf disease detection and classification using hierarchical support vector machine.
\newblock In \emph{2021 international symposium on electrical and electronics engineering (ISEE)}, pages 69--74. IEEE, 2021.

\bibitem[Tulshan and Raul(2019)]{tulshan2019plant}
Amrita~S Tulshan and Nataasha Raul.
\newblock Plant leaf disease detection using machine learning.
\newblock In \emph{2019 10th International Conference on Computing, Communication and Networking Technologies (ICCCNT)}, pages 1--6. IEEE, 2019.

\bibitem[Rumy et~al.(2021)Rumy, Hossain, Jahan, and Tanvin]{rumy2021iot}
SM~Shahidur~Harun Rumy, Md~Ishan~Arefin Hossain, Forji Jahan, and Tanjina Tanvin.
\newblock An iot based system with edge intelligence for rice leaf disease detection using machine learning.
\newblock In \emph{2021 IEEE International IOT, Electronics and Mechatronics Conference (IEMTRONICS)}, pages 1--6. IEEE, 2021.

\bibitem[Rajesh et~al.(2020)Rajesh, Vardhan, and Sujihelen]{rajesh2020leaf}
B~Rajesh, M~Vishnu~Sai Vardhan, and L~Sujihelen.
\newblock Leaf disease detection and classification by decision tree.
\newblock In \emph{2020 4th international conference on trends in electronics and informatics (ICOEI)(48184)}, pages 705--708. IEEE, 2020.

\bibitem[Jiang et~al.(2021)Jiang, Dong, Jiang, and Yang]{jiang2021recognition}
Zhencun Jiang, Zhengxin Dong, Wenping Jiang, and Yuze Yang.
\newblock Recognition of rice leaf diseases and wheat leaf diseases based on multi-task deep transfer learning.
\newblock \emph{Computers and Electronics in Agriculture}, 186:\penalty0 106184, 2021.

\bibitem[Turkoglu et~al.(2022)Turkoglu, Yaniko{\u{g}}lu, and Hanbay]{turkoglu2022plantdiseasenet}
Muammer Turkoglu, Berrin Yaniko{\u{g}}lu, and Davut Hanbay.
\newblock Plantdiseasenet: Convolutional neural network ensemble for plant disease and pest detection.
\newblock \emph{Signal, Image and Video Processing}, 16\penalty0 (2):\penalty0 301--309, 2022.

\bibitem[Chen et~al.(2023)Chen, Pan, and Wu]{chen2023apple}
Yiping Chen, Jinchao Pan, and Qiufeng Wu.
\newblock Apple leaf disease identification via improved cyclegan and convolutional neural network.
\newblock \emph{Soft Computing}, 27\penalty0 (14):\penalty0 9773--9786, 2023.

\bibitem[Rakib et~al.(2024)Rakib, Rahman, Razi, and Hasan]{rakib2024lightweight}
Abdul~Fattah Rakib, Rashik Rahman, Alim~Al Razi, and ASM~Touhidul Hasan.
\newblock A lightweight quantized cnn model for plant disease recognition.
\newblock \emph{Arabian Journal for Science and Engineering}, 49\penalty0 (3):\penalty0 4097--4108, 2024.

\bibitem[Elfatimi et~al.(2022)Elfatimi, Eryigit, and Elfatimi]{elfatimi2022beans}
Elhoucine Elfatimi, Recep Eryigit, and Lahcen Elfatimi.
\newblock Beans leaf diseases classification using mobilenet models.
\newblock \emph{IEEE Access}, 10:\penalty0 9471--9482, 2022.

\bibitem[Saleem et~al.(2021)Saleem, Shah, Sharif, Yasmin, Yong, and Cha]{saleem2021mango}
Rabia Saleem, Jamal~Hussain Shah, Muhammad Sharif, Mussarat Yasmin, Hwan-Seung Yong, and Jaehyuk Cha.
\newblock Mango leaf disease recognition and classification using novel segmentation and vein pattern technique.
\newblock \emph{Applied Sciences}, 11\penalty0 (24):\penalty0 11901, 2021.

\bibitem[Jin et~al.(2022)Jin, Li, Qi, Feng, Tian, and Mu]{jin2022grapegan}
Haibin Jin, Yue Li, Jianfang Qi, Jianying Feng, Dong Tian, and Weisong Mu.
\newblock Grapegan: Unsupervised image enhancement for improved grape leaf disease recognition.
\newblock \emph{Computers and Electronics in Agriculture}, 198:\penalty0 107055, 2022.

\bibitem[Metre and Sawarkar(2022)]{metre2022reviewing}
Vishakha~A Metre and Sudhir~D Sawarkar.
\newblock Reviewing important aspects of plant leaf disease detection and classification.
\newblock In \emph{2022 International Conference for Advancement in Technology (ICONAT)}, pages 1--8. IEEE, 2022.

\bibitem[Thakur et~al.(2022)Thakur, Khanna, Sheorey, and Ojha]{thakur2022trends}
Poornima~Singh Thakur, Pritee Khanna, Tanuja Sheorey, and Aparajita Ojha.
\newblock Trends in vision-based machine learning techniques for plant disease identification: A systematic review.
\newblock \emph{Expert Systems with Applications}, 208:\penalty0 118117, 2022.

\bibitem[Upadhyay et~al.(2025)Upadhyay, Chandel, Singh, Chakraborty, Nandede, Kumar, Subeesh, Upendar, Salem, and Elbeltagi]{upadhyay2025deep}
Abhishek Upadhyay, Narendra~Singh Chandel, Krishna~Pratap Singh, Subir~Kumar Chakraborty, Balaji~M Nandede, Mohit Kumar, A~Subeesh, Konga Upendar, Ali Salem, and Ahmed Elbeltagi.
\newblock Deep learning and computer vision in plant disease detection: a comprehensive review of techniques, models, and trends in precision agriculture.
\newblock \emph{Artificial Intelligence Review}, 58\penalty0 (3):\penalty0 92, 2025.

\bibitem[Yue et~al.(2025)Yue, Qi, Na, Liu, Yang, and Wang]{yue2025deep}
Xiang Yue, Kai Qi, Xinyi Na, Yanhua Liu, Fuhao Yang, and Wei Wang.
\newblock Deep learning for recognition and detection of plant diseases and pests.
\newblock \emph{Neural Computing and Applications}, 37\penalty0 (17):\penalty0 11265--11310, 2025.

\bibitem[Pacal et~al.(2024)Pacal, Kunduracioglu, Alma, Deveci, Kadry, Nedoma, Slany, and Martinek]{pacal2024systematic}
Ishak Pacal, Ismail Kunduracioglu, Mehmet~Hakki Alma, Muhammet Deveci, Seifedine Kadry, Jan Nedoma, Vlastimil Slany, and Radek Martinek.
\newblock A systematic review of deep learning techniques for plant diseases.
\newblock \emph{Artificial Intelligence Review}, 57\penalty0 (11):\penalty0 304, 2024.

\bibitem[Yilmaz et~al.(2025)Yilmaz, Bocekci, Safak, and Yildiz]{yilmaz2025advancements}
Esra Yilmaz, Sevim~Ceylan Bocekci, Cengiz Safak, and Kazim Yildiz.
\newblock Advancements in smart agriculture: A systematic literature review on state-of-the-art plant disease detection with computer vision.
\newblock \emph{IET Computer Vision}, 19\penalty0 (1):\penalty0 e70004, 2025.

\bibitem[Bhargava et~al.(2024)Bhargava, Shukla, Goswami, Alsharif, Uthansakul, and Uthansakul]{bhargava2024plant}
Anuja Bhargava, Aasheesh Shukla, Om~Prakash Goswami, Mohammed~H Alsharif, Peerapong Uthansakul, and Monthippa Uthansakul.
\newblock Plant leaf disease detection, classification, and diagnosis using computer vision and artificial intelligence: A review.
\newblock \emph{IEEE access}, 12:\penalty0 37443--37469, 2024.

\bibitem[Ngugi et~al.(2024)Ngugi, Ezugwu, Akinyelu, and Abualigah]{ngugi2024revolutionizing}
Habiba~N Ngugi, Absalom~E Ezugwu, Andronicus~A Akinyelu, and Laith Abualigah.
\newblock Revolutionizing crop disease detection with computational deep learning: a comprehensive review.
\newblock \emph{Environmental Monitoring and Assessment}, 196\penalty0 (3):\penalty0 302, 2024.

\bibitem[Shafik et~al.(2023)Shafik, Tufail, Namoun, De~Silva, and Apong]{shafik2023systematic}
Wasswa Shafik, Ali Tufail, Abdallah Namoun, Liyanage~Chandratilak De~Silva, and Rosyzie Anna Awg Haji~Mohd Apong.
\newblock A systematic literature review on plant disease detection: Motivations, classification techniques, datasets, challenges, and future trends.
\newblock \emph{Ieee Access}, 11:\penalty0 59174--59203, 2023.

\bibitem[Bhagat and Kumar(2022)]{bhagat2022comprehensive}
Monu Bhagat and Dilip Kumar.
\newblock A comprehensive survey on leaf disease identification \& classification.
\newblock \emph{Multimedia Tools and Applications}, 81\penalty0 (23):\penalty0 33897--33925, 2022.

\bibitem[Bangari et~al.(2022)Bangari, Rachana, Gupta, Sudi, and Baniya]{bangari2022survey}
Sindhuja Bangari, P~Rachana, Nihit Gupta, Pappu~Sah Sudi, and Kamlesh~Kumar Baniya.
\newblock A survey on disease detection of a potato leaf using cnn.
\newblock In \emph{2022 Second International Conference on Artificial Intelligence and Smart Energy (ICAIS)}, pages 144--149. IEEE, 2022.

\bibitem[Tugrul et~al.(2022)Tugrul, Elfatimi, and Eryigit]{tugrul2022convolutional}
Bulent Tugrul, Elhoucine Elfatimi, and Recep Eryigit.
\newblock Convolutional neural networks in detection of plant leaf diseases: A review.
\newblock \emph{Agriculture}, 12\penalty0 (8):\penalty0 1192, 2022.

\bibitem[Applalanaidu and Kumaravelan(2021)]{applalanaidu2021review}
Majji~V Applalanaidu and G~Kumaravelan.
\newblock A review of machine learning approaches in plant leaf disease detection and classification.
\newblock In \emph{2021 Third International Conference on Intelligent Communication Technologies and Virtual Mobile Networks (ICICV)}, pages 716--724. IEEE, 2021.

\bibitem[Jiang et~al.(2019)Jiang, Chen, Liu, He, and Liang]{jiang2019real}
Peng Jiang, Yuehan Chen, Bin Liu, Dongjian He, and Chunquan Liang.
\newblock Real-time detection of apple leaf diseases using deep learning approach based on improved convolutional neural networks.
\newblock \emph{Ieee Access}, 7:\penalty0 59069--59080, 2019.

\bibitem[Wu et~al.(2019)Wu, Zhang, and Meng]{wu2019identification}
Qiufeng Wu, Keke Zhang, and Jun Meng.
\newblock Identification of soybean leaf diseases via deep learning.
\newblock \emph{Journal of The Institution of Engineers (India): Series A}, 100:\penalty0 659--666, 2019.

\bibitem[Dananjayan et~al.(2022)Dananjayan, Tang, Zhuang, Hou, and Luo]{dananjayan2022assessment}
Sathian Dananjayan, Yu~Tang, Jiajun Zhuang, Chaojun Hou, and Shaoming Luo.
\newblock Assessment of state-of-the-art deep learning based citrus disease detection techniques using annotated optical leaf images.
\newblock \emph{computers and Electronics in Agriculture}, 193:\penalty0 106658, 2022.

\bibitem[Joshi et~al.(2022)Joshi, Das, Udutalapally, Pradhan, and Misra]{joshi2022ricebios}
Pranjal Joshi, Debanjan Das, Venkanna Udutalapally, Monoj~K Pradhan, and Sudip Misra.
\newblock Ricebios: Identification of biotic stress in rice crops using edge-as-a-service.
\newblock \emph{IEEE Sensors Journal}, 22\penalty0 (5):\penalty0 4616--4624, 2022.

\bibitem[Khan et~al.(2022)Khan, Quadri, Banday, and Shah]{khan2022deep}
Asif~Iqbal Khan, SMK Quadri, Saba Banday, and Junaid~Latief Shah.
\newblock Deep diagnosis: A real-time apple leaf disease detection system based on deep learning.
\newblock \emph{computers and Electronics in Agriculture}, 198:\penalty0 107093, 2022.

\bibitem[Chug et~al.(2023)Chug, Bhatia, Singh, and Singh]{chug2023novel}
Anuradha Chug, Anshul Bhatia, Amit~Prakash Singh, and Dinesh Singh.
\newblock A novel framework for image-based plant disease detection using hybrid deep learning approach.
\newblock \emph{Soft Computing}, 27\penalty0 (18):\penalty0 13613--13638, 2023.

\bibitem[Khatoon et~al.(2021)Khatoon, Hasan, Asif, Alshmari, and Yap]{khatoon2021image}
Shaheen Khatoon, Md~Maruf Hasan, Amna Asif, Majed Alshmari, and Yun-Kiam Yap.
\newblock Image-based automatic diagnostic system for tomato plants using deep learning.
\newblock \emph{Computers, Materials \& Continua}, 67\penalty0 (1), 2021.

\bibitem[Ashwinkumar et~al.(2022)Ashwinkumar, Rajagopal, Manimaran, and Jegajothi]{ashwinkumar2022automated}
S~Ashwinkumar, S~Rajagopal, V~Manimaran, and B~Jegajothi.
\newblock Automated plant leaf disease detection and classification using optimal mobilenet based convolutional neural networks.
\newblock \emph{Materials Today: Proceedings}, 51:\penalty0 480--487, 2022.

\bibitem[Prajapati et~al.(2017)Prajapati, Shah, and Dabhi]{prajapati2017detection}
Harshadkumar~B Prajapati, Jitesh~P Shah, and Vipul~K Dabhi.
\newblock Detection and classification of rice plant diseases.
\newblock \emph{Intelligent Decision Technologies}, 11\penalty0 (3):\penalty0 357--373, 2017.

\bibitem[Bevers et~al.(2022)Bevers, Sikora, and Hardy]{bevers2022pictures}
Noah Bevers, Edward~J Sikora, and Nate~B Hardy.
\newblock Pictures of diseased soybean leaves by category captured in field and with controlled backgrounds: Auburn soybean disease image dataset (asdid). dryad.
\newblock \emph{Dryad}, 2022.

\bibitem[Tassis et~al.(2021)Tassis, de~Souza, and Krohling]{tassis2021deep}
Lucas~M Tassis, Joao E~Tozzi de~Souza, and Renato~A Krohling.
\newblock A deep learning approach combining instance and semantic segmentation to identify diseases and pests of coffee leaves from in-field images.
\newblock \emph{Computers and Electronics in Agriculture}, 186:\penalty0 106191, 2021.

\bibitem[Adem et~al.(2023)Adem, Ozguven, and Altas]{adem2023sugar}
Kemal Adem, Mehmet~Metin Ozguven, and Ziya Altas.
\newblock A sugar beet leaf disease classification method based on image processing and deep learning.
\newblock \emph{Multimedia Tools and Applications}, 82\penalty0 (8):\penalty0 12577--12594, 2023.

\bibitem[Liu et~al.(2022)Liu, Hu, Zhou, Cai, He, Zhan, Hu, and Li]{liu2022ds}
Xuyao Liu, Yaowen Hu, Guoxiong Zhou, Weiwei Cai, Mingfang He, Jialei Zhan, Yahui Hu, and Liujun Li.
\newblock Ds-menet for the classification of citrus disease.
\newblock \emph{Frontiers in Plant Science}, 13:\penalty0 884464, 2022.

\bibitem[Thet et~al.(2020)Thet, Htwe, and Thein]{thet2020grape}
Khaing~Zin Thet, Khine~Khine Htwe, and Myint~Myint Thein.
\newblock Grape leaf diseases classification using convolutional neural network.
\newblock In \emph{2020 international conference on advanced information technologies (ICAIT)}, pages 147--152. IEEE, 2020.

\bibitem[Rangarajan et~al.(2022)Rangarajan, Whetton, and Mouazen]{rangarajan2022detection}
Aravind~Krishnaswamy Rangarajan, Rebecca~Louise Whetton, and Abdul~Mounem Mouazen.
\newblock Detection of fusarium head blight in wheat using hyperspectral data and deep learning.
\newblock \emph{Expert Systems with Applications}, 208:\penalty0 118240, 2022.

\bibitem[P{\'e}rez-Roncal et~al.(2022)P{\'e}rez-Roncal, Arazuri, Lopez-Molina, Jar{\'e}n, Santesteban, and L{\'o}pez-Maestresalas]{perez2022exploring}
Claudia P{\'e}rez-Roncal, Silvia Arazuri, Carlos Lopez-Molina, Carmen Jar{\'e}n, Luis~G Santesteban, and Ainara L{\'o}pez-Maestresalas.
\newblock Exploring the potential of hyperspectral imaging to detect esca disease complex in asymptomatic grapevine leaves.
\newblock \emph{Computers and Electronics in Agriculture}, 196:\penalty0 106863, 2022.

\bibitem[Singh et~al.(2020)Singh, Jain, Jain, Kayal, Kumawat, and Batra]{singh2020plantdoc}
Davinder Singh, Naman Jain, Pranjali Jain, Pratik Kayal, Sudhakar Kumawat, and Nipun Batra.
\newblock Plantdoc: A dataset for visual plant disease detection.
\newblock In \emph{Proceedings of the 7th ACM IKDD CoDS and 25th COMAD}, pages 249--253. 2020.

\bibitem[Saleem et~al.(2022)Saleem, Potgieter, and Arif]{saleem2022performance}
Muhammad~Hammad Saleem, Johan Potgieter, and Khalid~Mahmood Arif.
\newblock A performance-optimized deep learning-based plant disease detection approach for horticultural crops of new zealand.
\newblock \emph{IEEE Access}, 10:\penalty0 89798--89822, 2022.

\bibitem[Liang et~al.(2019)Liang, Xiang, Hu, Coppola, Zhang, and Sun]{liang2019pd2se}
Qiaokang Liang, Shao Xiang, Yucheng Hu, Gianmarc Coppola, Dan Zhang, and Wei Sun.
\newblock Pd2se-net: Computer-assisted plant disease diagnosis and severity estimation network.
\newblock \emph{Computers and electronics in agriculture}, 157:\penalty0 518--529, 2019.

\bibitem[Chouhan et~al.(2019)Chouhan, Singh, Kaul, and Jain]{chouhan2019data}
Siddharth~Singh Chouhan, Uday~Pratap Singh, Ajay Kaul, and Sanjeev Jain.
\newblock A data repository of leaf images: Practice towards plant conservation with plant pathology.
\newblock In \emph{2019 4th International Conference on Information Systems and Computer Networks (ISCON)}, pages 700--707. IEEE, 2019.

\bibitem[Moupojou et~al.(2023)Moupojou, Tagne, Retraint, Tadonkemwa, Wilfried, Tapamo, and Nkenlifack]{moupojou2023fieldplant}
Emmanuel Moupojou, Appolinaire Tagne, Florent Retraint, Anicet Tadonkemwa, Dongmo Wilfried, Hyppolite Tapamo, and Marcellin Nkenlifack.
\newblock Fieldplant: A dataset of field plant images for plant disease detection and classification with deep learning.
\newblock \emph{IEEE Access}, 11:\penalty0 35398--35410, 2023.

\bibitem[Shovon et~al.(2023)Shovon, Mozumder, Pal, Mridha, Asai, and Shin]{shovon2023plantdet}
Md~Sakib~Hossain Shovon, Shakrin~Jahan Mozumder, Osim~Kumar Pal, Muhammad~Firoz Mridha, Nobuyoshi Asai, and Jungpil Shin.
\newblock Plantdet: A robust multi-model ensemble method based on deep learning for plant disease detection.
\newblock \emph{IEEE Access}, 11:\penalty0 34846--34859, 2023.

\bibitem[Brown and De~Silva()]{brown30plant}
Dane Brown and Malithi De~Silva.
\newblock Plant disease detection on multispectral images using vision transformers.
\newblock In \emph{Proceedings of the 25th Irish Machine Vision and Image Processing Conference (IMVIP), Galway, Ireland}, volume~30.

\bibitem[Krizhevsky et~al.(2017)Krizhevsky, Sutskever, and Hinton]{krizhevsky2017imagenet}
Alex Krizhevsky, Ilya Sutskever, and Geoffrey~E Hinton.
\newblock Imagenet classification with deep convolutional neural networks.
\newblock \emph{Communications of the ACM}, 60\penalty0 (6):\penalty0 84--90, 2017.

\bibitem[Simonyan and Zisserman(2014)]{simonyan2014very}
Karen Simonyan and Andrew Zisserman.
\newblock Very deep convolutional networks for large-scale image recognition.
\newblock \emph{arXiv preprint arXiv:1409.1556}, 2014.

\bibitem[He et~al.(2016)He, Zhang, Ren, and Sun]{he2016deep}
Kaiming He, Xiangyu Zhang, Shaoqing Ren, and Jian Sun.
\newblock Deep residual learning for image recognition.
\newblock In \emph{Proceedings of the IEEE conference on computer vision and pattern recognition}, pages 770--778, 2016.

\bibitem[Huang et~al.(2017)Huang, Liu, Van Der~Maaten, and Weinberger]{huang2017densely}
Gao Huang, Zhuang Liu, Laurens Van Der~Maaten, and Kilian~Q Weinberger.
\newblock Densely connected convolutional networks.
\newblock In \emph{Proceedings of the IEEE conference on computer vision and pattern recognition}, pages 4700--4708, 2017.

\bibitem[Agarwal et~al.(2019)Agarwal, Gupta, and Biswas]{agarwal2019grape}
Mohit Agarwal, Suneet~Kr Gupta, and Kanad~Kishore Biswas.
\newblock Grape disease identification using convolution neural network.
\newblock In \emph{2019 23rd International Computer Science and Engineering Conference (ICSEC)}, pages 224--229. IEEE, 2019.

\bibitem[Zhong and Zhao(2020)]{zhong2020research}
Yong Zhong and Ming Zhao.
\newblock Research on deep learning in apple leaf disease recognition.
\newblock \emph{Computers and electronics in agriculture}, 168:\penalty0 105146, 2020.

\bibitem[Srivastava and Meena(2024)]{srivastava2024plant}
Meenakshi Srivastava and Jasraj Meena.
\newblock Plant leaf disease detection and classification using modified transfer learning models.
\newblock \emph{Multimedia Tools and Applications}, 83\penalty0 (13):\penalty0 38411--38441, 2024.

\bibitem[Varma et~al.(2025)Varma, Mate, Azeem, Sharma, and Singh]{varma2025automatic}
Teena Varma, Prajwal Mate, Noamaan~Abdul Azeem, Sanjeev Sharma, and Bhupendra Singh.
\newblock Automatic mango leaf disease detection using different transfer learning models.
\newblock \emph{Multimedia Tools and Applications}, 84\penalty0 (11):\penalty0 9185--9218, 2025.

\bibitem[Ahmed et~al.(2023)Ahmed, Ibrahim, Nadim, Rahman, Shejunti, Jabid, and Ali]{ahmed2023mangoleafbd}
Sarder~Iftekhar Ahmed, Muhammad Ibrahim, Md~Nadim, Md~Mizanur Rahman, Maria~Mehjabin Shejunti, Taskeed Jabid, and Md~Sawkat Ali.
\newblock Mangoleafbd: A comprehensive image dataset to classify diseased and healthy mango leaves.
\newblock \emph{Data in Brief}, 47:\penalty0 108941, 2023.

\bibitem[Bhagat et~al.(2024)Bhagat, Kumar, and Kumar]{bhagat2024optimized}
Monu Bhagat, Dilip Kumar, and Sunil Kumar.
\newblock Optimized transfer learning approach for leaf disease classification in smart agriculture.
\newblock \emph{Multimedia Tools and Applications}, 83\penalty0 (20):\penalty0 58103--58123, 2024.

\bibitem[Too et~al.(2019)Too, Yujian, Njuki, and Yingchun]{too2019comparative}
Edna~Chebet Too, Li~Yujian, Sam Njuki, and Liu Yingchun.
\newblock A comparative study of fine-tuning deep learning models for plant disease identification.
\newblock \emph{Computers and Electronics in Agriculture}, 161:\penalty0 272--279, 2019.

\bibitem[Shehu et~al.(2025)Shehu, Ackley, Marvellous, and Eteng]{shehu2025early}
Harisu~Abdullahi Shehu, Aniebietabasi Ackley, Mark Marvellous, and Ofem~Ebriba Eteng.
\newblock Early detection of tomato leaf diseases using transformers and transfer learning.
\newblock \emph{European Journal of Agronomy}, 168:\penalty0 127625, 2025.

\bibitem[Singh et~al.(2022)Singh, Tiwari, and Gupta]{singh2022deep}
Rajeev~Kumar Singh, Akhilesh Tiwari, and Rajendra~Kumar Gupta.
\newblock Deep transfer modeling for classification of maize plant leaf disease.
\newblock \emph{Multimedia tools and applications}, 81\penalty0 (5):\penalty0 6051--6067, 2022.

\bibitem[Chen et~al.(2022)Chen, Chen, Zeb, Yang, and Zhang]{chen2022mobile}
Weirong Chen, Junde Chen, Adnan Zeb, Shuangyuan Yang, and Defu Zhang.
\newblock Mobile convolution neural network for the recognition of potato leaf disease images.
\newblock \emph{Multimedia Tools and Applications}, 81\penalty0 (15):\penalty0 20797--20816, 2022.

\bibitem[Kaya and G{\"u}rsoy(2023)]{kaya2023novel}
Yasin Kaya and Ercan G{\"u}rsoy.
\newblock A novel multi-head cnn design to identify plant diseases using the fusion of rgb images.
\newblock \emph{Ecological Informatics}, 75:\penalty0 101998, 2023.

\bibitem[Haridasan et~al.(2023)Haridasan, Thomas, and Raj]{haridasan2023deep}
Amritha Haridasan, Jeena Thomas, and Ebin~Deni Raj.
\newblock Deep learning system for paddy plant disease detection and classification.
\newblock \emph{Environmental monitoring and assessment}, 195\penalty0 (1):\penalty0 120, 2023.

\bibitem[Kumar et~al.(2023)Kumar, Singh, Moudgil, and Kamini]{kumar2023systematic}
Yogesh Kumar, Rupinder Singh, Manu~Raj Moudgil, and Kamini.
\newblock A systematic review of different categories of plant disease detection using deep learning-based approaches.
\newblock \emph{Archives of Computational Methods in Engineering}, 30\penalty0 (8):\penalty0 4757--4779, 2023.

\bibitem[Tembhurne et~al.(2023)Tembhurne, Gajbhiye, Gannarpwar, Khandait, Goydani, and Diwan]{tembhurne2023plant}
Jitendra~V Tembhurne, Saurav~M Gajbhiye, Vedant~R Gannarpwar, Harshal~R Khandait, Purva~R Goydani, and Tausif Diwan.
\newblock Plant disease detection using deep learning based mobile application.
\newblock \emph{Multimedia Tools and Applications}, 82\penalty0 (18):\penalty0 27365--27390, 2023.

\bibitem[Nawaz et~al.(2024)Nawaz, Nazir, Javed, Amin, Jeribi, and Tahir]{nawaz2024coffeenet}
Marriam Nawaz, Tahira Nazir, Ali Javed, Sherif~Tawfik Amin, Fathe Jeribi, and Ali Tahir.
\newblock Coffeenet: A deep learning approach for coffee plant leaves diseases recognition.
\newblock \emph{Expert Systems with Applications}, 237:\penalty0 121481, 2024.

\bibitem[Jepkoech et~al.(2021)Jepkoech, Mugo, Kenduiywo, and Too]{jepkoech2021arabica}
Jennifer Jepkoech, David~Muchangi Mugo, Benson~K Kenduiywo, and Edna~Chebet Too.
\newblock Arabica coffee leaf images dataset for coffee leaf disease detection and classification.
\newblock \emph{Data in brief}, 36:\penalty0 107142, 2021.

\bibitem[Bir et~al.(2020)Bir, Kumar, and Singh]{bir2020transfer}
Paarth Bir, Rajesh Kumar, and Ghanshyam Singh.
\newblock Transfer learning based tomato leaf disease detection for mobile applications.
\newblock In \emph{2020 IEEE International Conference on Computing, Power and Communication Technologies (GUCON)}, pages 34--39. IEEE, 2020.

\bibitem[Guan(2021)]{guan2021novel}
Xulang Guan.
\newblock A novel method of plant leaf disease detection based on deep learning and convolutional neural network.
\newblock In \emph{2021 6th International conference on intelligent computing and signal processing (ICSP)}, pages 816--819. IEEE, 2021.

\bibitem[Hu et~al.(2020)Hu, Fan, Du, Li, Xiong, and Bekkering]{hu2020mdfc}
Wei-Jian Hu, Jie Fan, Yong-Xing Du, Bao-Shan Li, Naixue Xiong, and Ernst Bekkering.
\newblock Mdfc--resnet: An agricultural iot system to accurately recognize crop diseases.
\newblock \emph{IEEE Access}, 8:\penalty0 115287--115298, 2020.

\bibitem[Ji et~al.(2020)Ji, Zhang, and Wu]{ji2020automatic}
Miaomiao Ji, Lei Zhang, and Qiufeng Wu.
\newblock Automatic grape leaf diseases identification via unitedmodel based on multiple convolutional neural networks.
\newblock \emph{Information Processing in Agriculture}, 7\penalty0 (3):\penalty0 418--426, 2020.

\bibitem[Hassan et~al.(2021)Hassan, Maji, Jasi{\'n}ski, Leonowicz, and Jasi{\'n}ska]{hassan2021identification}
Sk~Mahmudul Hassan, Arnab~Kumar Maji, Micha{\l} Jasi{\'n}ski, Zbigniew Leonowicz, and El{\.z}bieta Jasi{\'n}ska.
\newblock Identification of plant-leaf diseases using cnn and transfer-learning approach.
\newblock \emph{Electronics}, 10\penalty0 (12):\penalty0 1388, 2021.

\bibitem[Pacal and I{\c{s}}{\i}k(2025)]{pacal2025utilizing}
Ishak Pacal and G{\"u}ltekin I{\c{s}}{\i}k.
\newblock Utilizing convolutional neural networks and vision transformers for precise corn leaf disease identification.
\newblock \emph{Neural Computing and Applications}, 37\penalty0 (4):\penalty0 2479--2496, 2025.

\bibitem[Yousafzai et~al.(2025)Yousafzai, Nasir, Tehsin, Malik, Keshta, Fitriyani, Gu, and Syafrudin]{yousafzai2025multi}
Samia~Nawaz Yousafzai, Inzamam~Mashood Nasir, Sara Tehsin, Dania~Saleem Malik, Ismail Keshta, Norma~Latif Fitriyani, Yeonghyeon Gu, and Muhammad Syafrudin.
\newblock Multi-stage neural network-based ensemble learning approach for wheat leaf disease classification.
\newblock \emph{IEEE Access}, 2025.

\bibitem[Oad et~al.(2024)Oad, Abbas, Zafar, Akram, Dong, Talpur, and Uddin]{oad2024plant}
Ammar Oad, Syed~Shoaib Abbas, Amna Zafar, Beenish~Ayesha Akram, Feng Dong, Mir Sajjad~Hussain Talpur, and Mueen Uddin.
\newblock Plant leaf disease detection using ensemble learning and explainable ai.
\newblock \emph{IEEE Access}, 12:\penalty0 156038--156049, 2024.

\bibitem[Amin et~al.(2022)Amin, Darwish, Hassanien, and Soliman]{amin2022end}
Hassan Amin, Ashraf Darwish, Aboul~Ella Hassanien, and Mona Soliman.
\newblock End-to-end deep learning model for corn leaf disease classification.
\newblock \emph{IEEE Access}, 10:\penalty0 31103--31115, 2022.

\bibitem[Jha et~al.(2024)Jha, Dembla, and Dubey]{jha2024deep}
Pradeep Jha, Deepak Dembla, and Widhi Dubey.
\newblock Deep learning models for enhancing potato leaf disease prediction: Implementation of transfer learning based stacking ensemble model.
\newblock \emph{Multimedia Tools and Applications}, 83\penalty0 (13):\penalty0 37839--37858, 2024.

\bibitem[Aboelenin et~al.(2025)Aboelenin, Elbasheer, Eltoukhy, El-Hady, and Hosny]{aboelenin2025hybrid}
Sherihan Aboelenin, Foriaa~Ahmed Elbasheer, Mohamed~Meselhy Eltoukhy, Walaa~M El-Hady, and Khalid~M Hosny.
\newblock A hybrid framework for plant leaf disease detection and classification using convolutional neural networks and vision transformer.
\newblock \emph{Complex \& Intelligent Systems}, 11\penalty0 (2):\penalty0 142, 2025.

\bibitem[Thakur et~al.(2024)Thakur, Gera, Aggarwal, Verma, Kaur, Singh, and Amoon]{thakur2024sunet}
Deepak Thakur, Tanya Gera, Ambika Aggarwal, Madhushi Verma, Manjit Kaur, Dilbag Singh, and Mohammed Amoon.
\newblock Sunet: Coffee leaf disease detection using hybrid deep learning model.
\newblock \emph{IEEE Access}, 12:\penalty0 149173--149191, 2024.

\bibitem[Reddy and Rekha(2021)]{reddy2021deep}
T~Vijaykanth Reddy and K~Sashi Rekha.
\newblock Deep leaf disease prediction framework (dldpf) with transfer learning for automatic leaf disease detection.
\newblock In \emph{2021 5th International Conference on Computing Methodologies and Communication (ICCMC)}, pages 1408--1415. IEEE, 2021.

\bibitem[Kaur et~al.(2024)Kaur, Mishra, Goyal, Gupta, Shankar, and Viriyasitavat]{kaur2024novel}
Prabhjot Kaur, Anand~Muni Mishra, Nitin Goyal, Sachin~Kumar Gupta, Achyut Shankar, and Wattana Viriyasitavat.
\newblock A novel hybrid cnn methodology for automated leaf disease detection and classification.
\newblock \emph{Expert Systems}, 41\penalty0 (7):\penalty0 e13543, 2024.

\bibitem[Thai and Le(2025)]{thai2025mobileh}
Huy-Tan Thai and Kim-Hung Le.
\newblock Mobileh-transformer: Enabling real-time leaf disease detection using hybrid deep learning approach for smart agriculture.
\newblock \emph{Crop Protection}, 189:\penalty0 107002, 2025.

\bibitem[Ghosh et~al.(2025)Ghosh, Rahat, Emon, Mashrafi, Al~Arafat~Tanzin, Mohanty, and Kant]{ghosh2025advanced}
Hritwik Ghosh, Irfan~Sadiq Rahat, Md~Mintajur~Rahman Emon, Md~Jisan Mashrafi, Mohammed~Abdul Al~Arafat~Tanzin, Sachi~Nandan Mohanty, and Shashi Kant.
\newblock Advanced neural network architectures for tomato leaf disease diagnosis in precision agriculture.
\newblock \emph{Discover Sustainability}, 6\penalty0 (1):\penalty0 312, 2025.

\bibitem[Shantkumari and Uma(2023)]{shantkumari2023grape}
M~Shantkumari and SV~Uma.
\newblock Grape leaf image classification based on machine learning technique for accurate leaf disease detection.
\newblock \emph{Multimedia tools and applications}, 82\penalty0 (1):\penalty0 1477--1487, 2023.

\bibitem[Li et~al.(2024)Li, Li, Zhang, Dong, Zhang, Wang, and Wei]{li2024sugarcanegan}
Xuechen Li, Xiuhua Li, Muqing Zhang, Qinghan Dong, Guiying Zhang, Zeping Wang, and Peng Wei.
\newblock Sugarcanegan: A novel dataset generating approach for sugarcane leaf diseases based on lightweight hybrid cnn-transformer network.
\newblock \emph{Computers and Electronics in Agriculture}, 219:\penalty0 108762, 2024.

\bibitem[Agarwal et~al.(2020)Agarwal, Singh, Arjaria, Sinha, and Gupta]{agarwal2020toled}
Mohit Agarwal, Abhishek Singh, Siddhartha Arjaria, Amit Sinha, and Suneet Gupta.
\newblock Toled: Tomato leaf disease detection using convolution neural network.
\newblock \emph{Procedia Computer Science}, 167:\penalty0 293--301, 2020.

\bibitem[Sunil et~al.(2020)Sunil, Jaidhar, and Patil]{sunil2020empirical}
CK~Sunil, CD~Jaidhar, and Nagamma Patil.
\newblock Empirical study on multi convolutional layer-based convolutional neural network classifier for plant leaf disease detection.
\newblock In \emph{2020 IEEE 15th International Conference on Industrial and Information Systems (ICIIS)}, pages 460--465. IEEE, 2020.

\bibitem[Islam et~al.(2025)Islam, Azad, Arman, Alyami, and Hasan]{islam2025plantcarenet}
Muhaiminul Islam, AKM Azad, Shifat~E Arman, Salem~A Alyami, and Md~Mehedi Hasan.
\newblock Plantcarenet: an advanced system to recognize plant diseases with dual-mode recommendations for prevention.
\newblock \emph{Plant Methods}, 21\penalty0 (1):\penalty0 52, 2025.

\bibitem[Matin et~al.(2023)Matin, Ava, Uddin, et~al.]{matin2023dataset}
Md~Mafiul~Hasan Matin, Mst Ava, Mohammad~Shorif Uddin, et~al.
\newblock A dataset for accurately identifying eggplant diseases.
\newblock \emph{Golam Moazzam and Uddin, Mohammad Shorif, A Dataset for Accurately Identifying Eggplant Diseases}, 2023.

\bibitem[Nagi and Tripathy(2022)]{nagi2022deep}
Reva Nagi and Sanjaya~Shankar Tripathy.
\newblock Deep convolutional neural network based disease identification in grapevine leaf images.
\newblock \emph{Multimedia Tools and Applications}, 81\penalty0 (18):\penalty0 24995--25006, 2022.

\bibitem[Alam et~al.(2024)Alam, Jowthi, and Pathak]{alam2024comparing}
Touhidul~Seyam Alam, Chandni~Barua Jowthi, and Abhijit Pathak.
\newblock Comparing pre-trained models for efficient leaf disease detection: a study on custom cnn.
\newblock \emph{Journal of Electrical Systems and Information Technology}, 11\penalty0 (1):\penalty0 12, 2024.

\bibitem[Sharma and Vardhan(2024)]{sharma2024enhanced}
Shubham Sharma and Manu Vardhan.
\newblock Enhanced plant disease detection using a custom cnn with advanced feature extraction techniques.
\newblock In \emph{2024 15th International Conference on Computing Communication and Networking Technologies (ICCCNT)}, pages 1--7. IEEE, 2024.

\bibitem[Mustafa et~al.(2023)Mustafa, Umer, Hafeez, Hameed, Sohaib, Ullah, and Madni]{mustafa2023pepper}
Hassan Mustafa, Muhammad Umer, Umair Hafeez, Ahmad Hameed, Ahmed Sohaib, Saleem Ullah, and Hamza~Ahmad Madni.
\newblock Pepper bell leaf disease detection and classification using optimized convolutional neural network.
\newblock \emph{Multimedia Tools and Applications}, 82\penalty0 (8):\penalty0 12065--12080, 2023.

\bibitem[Singla et~al.(2024)Singla, Kalavakonda, and Senthil]{singla2024detection}
Puja Singla, Vijaya Kalavakonda, and Ramalingam Senthil.
\newblock Detection of plant leaf diseases using deep convolutional neural network models.
\newblock \emph{Multimedia Tools and Applications}, pages 1--17, 2024.

\bibitem[Joseph et~al.(2024)Joseph, Pawar, and Chakradeo]{joseph2024real}
Diana~Susan Joseph, Pranav~M Pawar, and Kaustubh Chakradeo.
\newblock Real-time plant disease dataset development and detection of plant disease using deep learning.
\newblock \emph{IEEE Access}, 2024.

\bibitem[Roussin-L{\'e}veill{\'e}e et~al.(2024)Roussin-L{\'e}veill{\'e}e, Rossi, Castroverde, and Moffett]{roussin2024plant}
Charles Roussin-L{\'e}veill{\'e}e, Christina~AM Rossi, Christian Danve~Marco Castroverde, and Peter Moffett.
\newblock The plant disease triangle facing climate change: A molecular perspective.
\newblock \emph{Trends in Plant Science}, 2024.

\bibitem[Lian et~al.(2021)Lian, Dong, Li, Wang, Zhou, and Li]{lian2021effects}
Sen Lian, Xiang-li Dong, Ping-liang Li, Cai-xia Wang, Shan-yue Zhou, and Bao-hua Li.
\newblock Effects of temperature and moisture on conidia germination, infection, and acervulus formation of the apple marssonina leaf blotch pathogen (diplocarpon mali) in china.
\newblock \emph{Plant Disease}, 105\penalty0 (4):\penalty0 1057--1064, 2021.

\bibitem[Cao et~al.(2024)Cao, Gong, Li, Tang, Li, and Zhang]{cao2024pathogenic}
Chen Cao, Shun Gong, Yan Li, Jinqi Tang, Tianzhong Li, and Qiulei Zhang.
\newblock Pathogenic factors and mechanisms of the alternaria leaf spot pathogen in apple.
\newblock \emph{Horticulturae}, 10\penalty0 (3):\penalty0 212, 2024.

\bibitem[Papp et~al.(2020)Papp, Gao, Thapa, Olmstead, and Khan]{papp2020field}
David Papp, Liqiang Gao, Ranjita Thapa, Dan Olmstead, and Awais Khan.
\newblock Field apple scab susceptibility of a diverse malus germplasm collection identifies potential sources of resistance for apple breeding.
\newblock \emph{CABI Agriculture and Bioscience}, 1:\penalty0 1--14, 2020.

\end{thebibliography}

\end{document}